\documentclass{article}

\usepackage[preprint, nonatbib]{neurips_2024}

\usepackage[pagebackref,breaklinks,colorlinks,citecolor=citeblue]{hyperref}  
\usepackage[utf8]{inputenc} 
\usepackage[T1]{fontenc}    
\usepackage{url}            
\usepackage{booktabs}       
\usepackage{amsfonts}       
\usepackage{nicefrac}       
\usepackage{microtype}      
\usepackage{xcolor}         
\usepackage{enumitem}
\usepackage{overpic}
\usepackage{float}
\usepackage{marginnote}

\usepackage{CJKutf8}

\definecolor{citeblue}{rgb}{0.21,0.49,0.74}
\newcommand{\eg}{\emph{e.g.}}

\title{The Dawn of Video Generation: Preliminary Explorations with SORA-like Models}

\author{%
  Ailing Zeng$^{1}$\thanks{Equal Contribution}, Yuhang Yang$^{2*}$, Weidong Chen$^{1}$, Wei Liu$^{1}$  \\
  $^{1}$~Tencent, AI Lab \quad $^{2}$~USTC\\
  \href{https://ailab-cvc.github.io/VideoGen-Eval/}{https://ailab-cvc.github.io/VideoGen-Eval/}
}

\begin{document}

\maketitle

\begin{abstract}

High-quality video generation, encompassing text-to-video (T2V), image-to-video (I2V), and video-to-video (V2V) generation, holds considerable significance in content creation to benefit anyone express their inherent creativity in new ways and world simulation to modeling and understanding the world. Models like SORA have advanced generating videos with higher resolution, more natural motion, better vision-language alignment, and increased controllability, particularly for long video sequences. These improvements have been driven by the evolution of model architectures, shifting from UNet to more scalable and parameter-rich DiT models, along with large-scale data expansion and refined training strategies. However, despite the emergence of DiT-based closed-source and open-source models, a comprehensive investigation into their capabilities and limitations remains lacking. Furthermore, the rapid development has made it challenging for recent benchmarks to fully cover SORA-like models and recognize their significant advancements. Additionally, evaluation metrics often fail to align with human preferences.

This report studies a series of SORA-like models to bridge the gap between academic research and industry practice, providing a more profound analysis of recent video generation advancements. We design over 700 prompts with systematic perspectives to thoroughly evaluate existing T2V, I2V, and V2V models. Then, we compare \textbf{10} closed-source and \textbf{3} open-source models, demonstrating over 8,000 generated video cases. Since automated assessments still struggle to reflect real performance, we encourage readers to view the generated video results on our website. Seeing is believing. This study examines: i) impacts on vertical-domain application models, such as human-centric animation and robotics; ii) key objective capabilities, such as text alignment, visual and motion quality, composition, stability, creativity, etc.; iii) applying to ten real-life applications; iv) potential usage scenarios and tasks. Finally, we provide in-depth discussions on challenges and future research. All the results are publicly accessible as a new video generation benchmark, and we will continuously update the results.

\end{abstract}

{
  \hypersetup{linkcolor=black}
  \tableofcontents
  \label{sec:toc}
}

\section{Introduction}

The emergence of SORA \cite{openai2024sora} enables the creation of highly realistic and imaginative videos from text instructions with one-minute sequences. It also demonstrates that scaling up video generation models is a promising path towards building general-purpose simulators of the physical world. Notably, several closed-source models have launched websites and products directly without open-sourcing their corresponding models. Meanwhile, a significant performance gap persists between open-source and closed-source models, especially in model training with large-scale computational resources, as well as the collection and annotation of extensive datasets. This disparity has resulted in a divergence between academic research and industrial development. Furthermore, numerous fundamental research questions in video generation remain inadequately addressed from a research perspective \cite{cho2024sora,liu2024sora,zhou2024sora,wang2024does_sora,kustudic2024hero_sora,sun2024sora_t2vsurvey,zhangopenai_medicine_sora,selva2023videotransformer_survey,zhang2024sora,mogavi2024sora,gupta2023photorealistic,ma2024latte_videogen,nan2024openvid}. 

In the past year, several benchmarks have sought to establish fair comparison methods by assembling and comparing numerous video generation models while introducing comprehensive evaluation metrics with detailed quantitative comparisons \cite{ji2024t2vbench,feng2024tc_bench,liu2024fetv_bench,hu2023t2v_benchmark,li2024videoeval_bench,liu2024evalcrafter_bench,miao2024t2vsafetybench,yuan2024magictime,yuan2024chronomagic,xing2023survey}. However, their evaluation subjects are often outdated and fail to represent the state-of-the-art model performance to date~\cite{guo2024comparative}. This limitation undermines the effort to keep problem understanding at the field's cutting edge. Moreover, these works tend to focus more on multi-dimensional quantitative evaluations. Yet, developing metrics that fully align with human perception remains challenging, and scores from user studies also exhibit biases. Lastly, most prompts are designed via GPTs instead of considering various expert domain knowledge and meaningful prompts. 

In contrast, the qualitative results of video generation more directly reflect the prevailing issues that current models suffer from. Simultaneously, much of video application work relies on fine-tuned and highly controlled video generation explorations built upon open-source foundation models (e.g., stable diffusion (SD) \cite{rombach2022sd}, stable video diffusion (SVD) \cite{blattmann2023svd}, and Open-Sora \cite{opensora,yuan_opensora}). The capabilities of these foundation models significantly impact these efforts, encompassing model architecture, data construction, training strategies, and final performance. Additionally, it remains unclear whether improvements in foundation models will resolve many existing research challenges or whether new challenges will emerge as these models continue to scale up.

This report adopts a different perspective from previous evaluation studies to more clearly investigate the capabilities and limitations of current SORA-like video generation models, with comprehensive comparisons of generated videos and qualitative results, inspired by ~\cite{yang2023dawn} covering a broad range of domains and tasks. Instead of providing quantitative metrics, we demonstrate over 8,000 non-cherry picked generated videos via over 700 designed inputs to systematically showcase, analyze, and compare the output videos of these recent models across four core aspects: i) vertical-domain video generation in Section~\ref{sec:2}; ii) subjective abilities (\emph{e.g.}, consistency, composition, and identity preservation) in Section~\ref{sec:3}; iii) practical video applications in conjunction with the needs of hundreds of surveyed users in Section~\ref{sec:4}; iv) explorations on potential features and usage cases in Section~\ref{sec:5}, more detailed comparisons with open-source models in Section~\ref{sec:opensource}, current challenges and future directions in Section~\ref{sec:challenge}. Seeing is believing. Unlike text and image generation, video generation needs to put more effort into observing the generated video to gain a deeper insight into the problem.

Specifically, we summarize our observations through a series of preliminary explorations below.

\begin{itemize}

\item \textbf{Superiority of Closed-Source Models:} From various perspectives, closed-source models consistently exhibit significantly higher visual and motion quality than open-source models and surpass previous UNet-based models, especially in generating natural and dynamic motions in videos, even with rich multi-shot scenes and emotional expressions. These models excel in simulating the cinematic quality and texture of scenes.

\item \textbf{Advantages in Text-to-Video (T2V) Generation:} Among closed-source models, Gen-3, Kling v1.5, and Minimax exhibit superior overall performance on T2V tasks. In specific, Minimax excels in textual control, particularly in depicting human expressions, camera motion, multi-shot generation, and subject dynamics. Gen-3, on the other hand, stands out in controlling lighting, textures, and cinematographic techniques. Kling v1.5 shows good trade-offs among visual, controllability, and motion ability. Interestingly, all models have some good features. Luma emphasizes broader camera movements while keeping the subject's movement more restrained. In contrast, Vidu exhibits larger subject movements with a fast speed, and Qingying shows moderate proficiency in text-aligned generation. Each model has distinct motion representation characteristics due to different data distributions, model sizes, and training strategies. 

\item \textbf{Advantages in Image-to-Video (I2V) Generation:} Thanks to the better local-to-global motion modeling of foundation T2V models, closed-source models can animate the given image with more reasonable and temporal-consistent motion compared with UNet-based models. Specifically, novelty views, poses, natural lights, and textures will be generated from the given image. For Kling and Gen-3, the character animation turns out to be high-quality character preservation and vivid motion generation. Interestingly, they can also conduct image-to-video inpainting, outpainting, interpolation, super-resolution, and general enhancement tasks. Vidu and Luma tend to show highly dynamic subject and camera movements, respectively. 

\item \textbf{Remaining Limitations in T2V:} 
Although closed-source models have significantly improved overall quality (\emph{i.e.}, from 10\% to 40\% overall performance), they still fall short of perfection in many aspects. There are shared deficiencies in closed-source models regarding T2V generation, particularly in aspects such as poor text-aligned generation along spatial and temporal dimensions, low-resolution region generation (\emph{e.g.}, small faces), dynamic motions, reasoning ability, ID consistency along long sequences (\emph{e.g.}, 10s duration or more) and multi-shot scenarios, compositional spatio-temporal relations, complex physical interactions and adherence to physical rules (\emph{e.g.}, breaking glass, inflating balloons, and playing with balls), multilingual text generation, and stability.  

\item \textbf{Remaining Limitations in I2V:} 
For I2V tasks, there are instances where certain capabilities of T2V models are not fully realized. These include difficulties understanding the detailed and semantic information of the input image and a tendency to introduce new objects rather than accurately animate the existing elements or objects. Besides, maintaining object and human consistency (appearance, posture, texture, structure, etc.) is especially hard when the motion is highly dynamic. 

\item \textbf{Comparisons in Vertical-Domain Tasks:} 
Current models lack spatio-temporal fine-grained caption and domain-specific knowledge, such as facial expressions, speech, actions, and specialized autonomous driving descriptions. This makes precise video generation control challenging through generic I2V with input text. However, general-purpose models enhance core capabilities like generalization, composition, and diversity, offering effective scene modeling and a better understanding of human-object/environment interactions. Unlike explicit keypoint-driven video generation, which suffers from accuracy condition issues, general-purpose models more robustly handle human action modeling, mitigating such challenges.

\item \textbf{Performance in Ten Application Scenarios:} These I2V models (\emph{e.g.}, Gen-3 and Kling v1.5) perform well in landscape scenes, single object and animal motion, relighting, creative scenarios, and subtle animation. However, they still face significant challenges in applications involving human character animation, complex physical movements, niche motion scenes, count or logic variations in game-related contexts, and maintaining consistency and natural transitions in film-based multi-shot changes. These issues are particularly evident in the imperfections of many local and fine-grained details or regions.

\item \textbf{Evaluation of Diverse Objective Capabilities:} There are varying strengths and weaknesses across different dimensions. In specific, current video generation models have shown good performance in basic semantic alignment, such as appearance, lighting, and style. Building on this capability, they have also demonstrated a certain degree of proficiency in combining multiple instances and motions, as well as creativity in merging unrelated concepts. The key point is to adjust the distribution of the scenarios to make them work well, such as eating shows and hugging actions from Kling, special effects from Pika, and expressive emotions from MiniMax. However, their ability to handle highly diverse generalization, fine-grained motions, and multiple scene transitions remains limited. In particular, the stability (or the trade-offs between stability and diversity) requires significant improvement, which means generating multiple results with the same input varies widely.

\item \textbf{Exploration of Usage Scenarios and Tasks:} This encompasses several advanced techniques, such as video outpainting, super-resolution, texture generation, and transforming other styles into photo-realistic styles. This showcases the potential of video generation models to adapt to complex scenarios and expand beyond conventional tasks.

\item \textbf{Future Directions:} 
Video generation is still at its dawn, with a vast amount of research topics yet to be explored. We summarize them as follows: modeling multimodal generation to simulate the world effectively, unifying video perception, understanding, and generation tasks, designing novel architectures for interactive and real-time video generation, incorporating few-shot learning techniques for rare scenario adaptation and test-time domain adaptation, and long sequence modeling with multi-shot and ID consistency for film-making. Additionally, efforts should be made to incorporate user feedback to continuously improve the quality, controllability, and customization of generated videos, enhance model robustness and stability while keeping diversity, and address ethical considerations, safety, and explainability to ensure the responsible use of video generation technologies. 
\end{itemize}

Notably, this report aims to showcase recent advanced general-purpose video generation models with various prompts and comparisons to learn more from generated demonstrations and inspire future works on emerging problems. 
Due to the misalignment and inaccuracy of existing evaluation metrics, we do not evaluate these models quantitatively at this stage. 
Instead, we encourage readers to watch these videos directly. 

\subsection{Task Definition and Input Modalities}
\paragraph{Text-to-Video Generation.} This task transforms natural language prompts from users or creators into text-aligned, natural, dynamic, and realistic videos. The input consists of a textual description containing key components. According to official instructions\footnote{https://help.runwayml.com/hc/en-us/articles/30586818553107-Gen-3-Alpha-Prompting-Guide}\footnote{ https://docs.qingque.cn/d/home/eZQDKi7uTmtUr3iXnALzw6vxp}\footnote{https://pkocx4o26p.feishu.cn/docx/UCc6dHBE3ohwqxxCgDPcSEMinMc}\footnote{https://lumaai.notion.site/FAQ-and-Prompt-Guide-Luma-Dream-Machine-9e4ec319320a49bc832b6708e4ae7c46}, the more detailed and structured the description, the richer, more precise, and of higher quality the resulting video will be. Challenges arise in achieving precise text alignment for spatio-temporal information, complex and consistent motion modeling, diverse and large-scale training data to cover the physical world, and creativity abilities, among other aspects.

The general structure of the textual input should be `` character (with its detailed appearance and motion descriptions, \emph{e.g.}, clothing, texture, number, expression, and action) + object (related to human-object interaction, object shape, function, and motion) + scene/environment (\emph{e.g.}, material, color, atmosphere, shadow, light, and dynamics) + style (\emph{e.g.}, artistic style, and film genre) + camera (\emph{e.g.}, FPV, static, and wide angle) with corresponding movement types, directions, speeds and strength.'' For instance, one of our prompts is ``A static, medium shot of three small toy robots on a table that make various expressions with their digital faces. The robots are stylized with an animation aesthetic.'' Existing models will improve poor prompts through prompt enhancement.

\paragraph{Image-to-Video Generation.} This task aims to generate a dynamic video from a static image, often supplemented by additional inputs such as text or motion prompts to guide the movement. I2V is a step further than T2V in controllability, enabling users to have fine-grained control over the content. Unlike text-to-video generation, this approach focuses more on motion and dynamics. The main challenge is preserving the appearance and semantic content while maintaining dynamic and text-aligned motions. Due to the creative and diverse motions, the model would show the ability to generate novel views and generate invisible appearance frames with temporal coherency and realism.

Some models support arbitrary input resolutions (\emph{e.g.}, Kling, and Vidu), while others (\emph{e.g.}, Gen-3, and Qingying) should be in a fixed size. Thus, we pad these input images into the fixed size.

\paragraph{Video-to-Video Generation.} This task covers many practical sub-tasks when generating a video for another video by altering or enhancing its visual or temporal characteristics while maintaining consistency with the original. This process could include video style transfer, enhancement, editing, novel scene synthesis, and even video perception tasks (\emph{e.g.}, video depth and motion estimation) etc. Applications extend from converting low-resolution videos to high-resolution, altering the appearance of objects or backgrounds, generating new scenes based on given movements, and creating artistic video transformations. This task encounters several challenges: i) preserving input video structure, motion, and identity contents; ii) ensuring temporal consistency (\emph{e.g.}, comparing Gen1 with Gen-3 due to different video foundation models); iii) aligning text for various precise styles or editing instructions.
In this report, since only Gen-3 provides video-to-video generation via textual descriptions, we follow the official guidance to evaluate the effectiveness of the stylization and editing abilities within a 10-second video.

\subsection{SORA-like Model Objectives}

For the SORA-like DiT-based (Transformer-based Diffusion Model) video generation models discussed below, compared to previous UNet-based models \cite{blattmann2023svd,guo2023animatediff,zhang2023show1,chen2024videocrafter2}, they exhibit improvements in several key aspects and demonstrates distinct advantages: i) \emph{Enhanced expressive capability}: The DiT architecture has a stronger long sequential modeling capacity, enabling it to better capture the complex spatio-temporal relationships across different frames and patches in a video. This results in more coherent and natural motion generation, especially for maintaining temporal consistency over longer video sequences; ii) Higher generation visual and motion quality: Due to the powerful global modeling ability of the Transformer-based architecture, DiT-based models effectively capture intricate details and overall structures more accurately as the scaling up data and model size, resulting in improved visual quality and clarity; iii) Superior multi-modal information integration, and flexible scale adaptability: DiT-based models excel in handling information from different modalities and captures local to global semantic information from text or images, handling complex contextual relationships and long-sequence dense descriptions effectively. 

\begin{figure}[t]
	\centering
        \begin{overpic}[width=1.\linewidth]{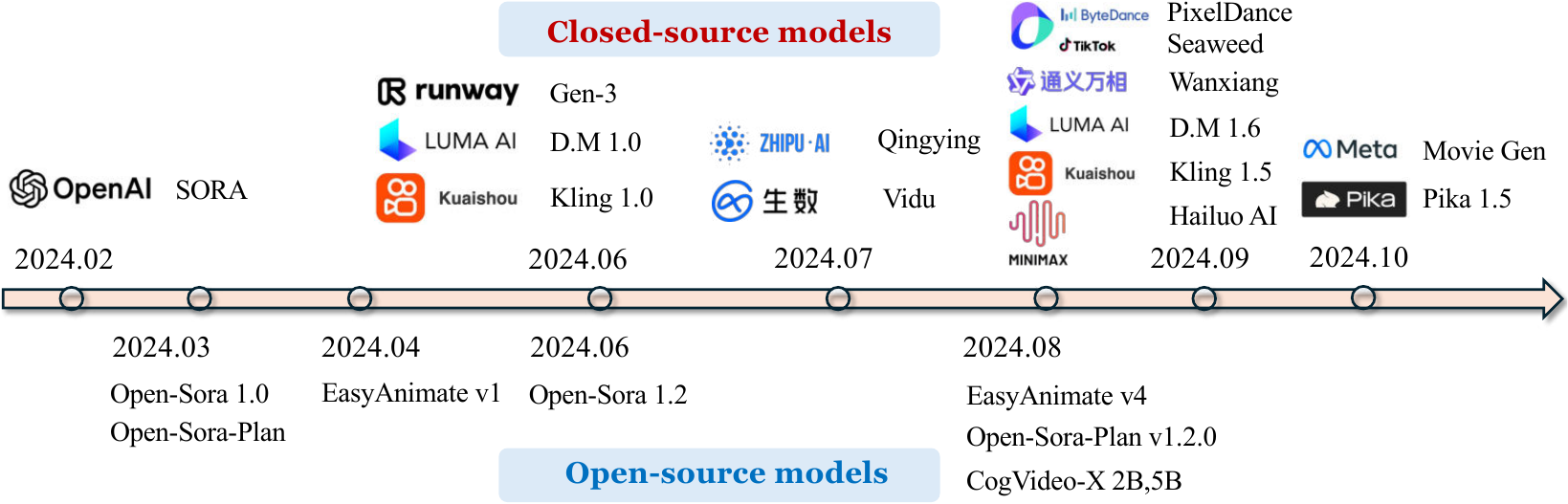}
	\end{overpic}
	\caption{The timeline of recent SORA-like models, including closed-source models (the upper) and open-source models (the lower). We summarize and introduce these models in this report.}
 \label{Fig:timeline}
\end{figure}

The advancements in SORA-like video generation can be attributed to improvements in the quantity and quality of data (videos and the corresponding captions), the capacity of models, and pre- and post-training and optimization strategies. We present a timeline of the rapidly emerging models in Fig.~\ref{Fig:timeline}, illustrating the swift development of this field and the necessity to stay updated on the progress of these large foundation video models.

\subsubsection{Closed-source Models}

\paragraph{SORA (OpenAI)~\cite{opensora}} was published in February 2024. Impressively, SORA generated intricate scenes featuring multiple characters, specific motion types, and precise details of both subject and background through large-scale training of generative models on video data. This indicates that scaling up such models is promising for building general-purpose simulators of the physical world \cite{openai2024sora,weng2024video}. Notably, unlike previous UNet-based models \cite{zhang2023show1,luo2023videofusion,blattmann2023align_ldm_video,blattmann2023svd,wang2023lavie,zeng2024pixel_Dance}, SORA employs a scalable transformer architecture operating on video and image latent patches. The text-conditioned diffusion model is trained on a diverse dataset of videos with varying durations, resolutions, and aspect ratios alongside images. The largest model developed demonstrates the capability of generating high-quality videos up to a minute in length. However, since SORA is not yet accessible to the public (we can only access the released videos), our evaluation will be postponed until the model becomes available.

\paragraph{Kling (Kuaishou)~\cite{kuaishou2024kling}} was released in June 2024, offering text-to-video, image-to-video (with start and end keyframe images provided), and video extension functionalities across its apps and website. Utilizing 3D spatio-temporal attention modules to capture intricate local-to-global relationships and a DiT-based architecture to enhance spatio-temporal capabilities, Kling generates lifelike, large-motion sequences. With efficient training and inference, Kling can produce 5-second or 10-second videos at 30 fps, ensuring natural motion and cinematic 1080p quality. Key features include authentic physics simulations for realistic scenes, imaginative concept fusion to translate creative text into vivid visuals, and flexible aspect ratios to accommodate various formats. Kling’s image-to-video function animates static images into captivating 5-second videos, while its video extension feature enables seamless video lengthening up to 3 minutes with creative text control. In September 2024, Kling released version 1.5 with a better text-to-video ability.

\paragraph{Dream Machine (LumaLabs)~\cite{luma2024dm}} was launched in June 2024 with text-to-video and image-to-video (with start and end keyframe images provided) functionalities on its website. All generated videos are 5 seconds long at 24 fps. The initial version of Dream Machine focuses on generating high-quality, realistic videos from text and images with remarkable speed and efficiency (\emph{e.g.}, capable of generating 120 frames in 120 seconds) through its built-in scalable and multimodal transformer architecture. Key features include generating high-quality videos from either text or images, fast inference speeds, action-packed realistic cinematography shots, character and physics consistency, and cinematic camera moves. Recently, Dream Machine has released two updates: i) version 1.5, with enhanced text-to-video capabilities, improved prompt understanding, custom text rendering, and higher-quality image-to-video; ii) version 1.6, which offers more controllable camera motion with simple textual commands (\emph{e.g.}, move left and up, push in and out, pan left and right, and orbit left and right).

\paragraph{Gen-3 Alpha (Runway)~\cite{runway2024gen3}} was also released in June 2024, featuring text-to-video and image-to-video (given start or end keyframe images) functionalities on its website. The generated videos are either 5 or 10 seconds long at 24 fps. Gen-3 Alpha employs a scalable, efficient transformer architecture trained on large-scale, high-quality video and image data. This multimodal approach enables the model to capture rich spatio-temporal information, producing physically reasonable, dynamically consistent videos with high text-aligned precision. Key features include fine-grained temporal controls via leveraging temporally dense captions for precise keyframe and imaginative transitions, photo-realistic human generation with realistic actions and expressions, artistic-centric designs with diverse styles and cinematic terminology, and complex physics-based simulations for hyper-realistic rendering asset generation and generative visual effects (\emph{e.g.}, hair and fur simulation, landscape flythroughs, green screen, special effects, 3D model rotation, etc\footnote{https://runwayml.com/product/use-cases}).

Recently, Runway has introduced a video-to-video function in both the Gen-3 Alpha and the Gen-3 Alpha Turbo models\footnote{https://runway-ai.ai/gen-3-alpha-turbo/}, enabling 7$\times$ faster image-to-video generation performance across various use cases compared to the Gen-3 Alpha model.

\paragraph{Vidu~\cite{bao2024vidu} (Shengshu)~\cite{shengshu2024vidu}} was released in late July 2024, providing text-to-video, image-to-video (given the start image), and character-to-video (given the reference image) capabilities on its platform. The generated videos are either 4 or 8 seconds long at 16 fps. Built on a diffusion model with U-ViT \cite{bao2023all_udit,bao2023one_udit} as its backbone, Vidu leverages scalability and long-sequence modeling, enabling the generation of coherent and dynamic videos. It supports realistic and imaginative video generation, showcasing strong performance in professional photography techniques such as scene transitions, smooth cuts with subject consistency, various camera movements, 3D consistency, subject-driven video generation, and lighting effects. In addition, Vidu performs T2V at a fast 30-second speed.

\begin{table}[!ht]
    \centering
    \small
    \caption{Demonstration of the released date, resolution of text-to-video generation, the FPS, generated length, and available functions. II2V means the task of inputting the first and last images to generate the video. The upper table lists closed-source models, and the lower table shows open-source models.}
    \scalebox{1}{
    \renewcommand\arraystretch{1}
    \setlength{\tabcolsep}{1.0mm}{
    \begin{tabular}{l|lllll|llll}
    
    \toprule
        \textbf{Models} & \textbf{Date}& \textbf{Resolution(T2V)}&  \textbf{FPS}&  \textbf{Frames(T2V)}& \textbf{Frames(I2V)} & \textbf{T2V} & \textbf{I2V} & \textbf{II2V} & \textbf{V2V} \\ \midrule
        Kling 1.0 & 24.06&1280$\times$720 & 30 & 153&153&$\checkmark$&$\checkmark$&$\checkmark$&$\times$ \\ 
         Kling 1.5 & 24.09&1280$\times$720 & 30 & 153&153&$\checkmark$&$\checkmark$&$\times$&$\times$ \\
        Gen-3 & 24.06&1280$\times$768 & 24 &128&125&$\checkmark$&$\checkmark$&$\times$&$\checkmark$ \\ 
        Luma 1.0 & 24.06&1360$\times$752 & 24&126&121&$\checkmark$&$\checkmark$&$\checkmark$&$\times$\\ 
         Luma 1.5&24.09&1360$\times$752& 24&126&121&$\checkmark$&$\checkmark$&$\checkmark$&$\times$\\ 
        Vidu &24.07& 688$\times$384  & 16&60&60&$\checkmark$&$\checkmark$&$\times$&$\times$\\ 

       Qingying &24.07&1440$\times$960  & 16&96&96&$\checkmark$&$\checkmark$&$\times$&$\times$\\
        Hailuo&24.09&1280$\times$720 & 25&141&-&$\checkmark$&$\times$&$\times$&$\times$\\
        Wanxiang&24.09&1280$\times$720  & 30&152&137&$\checkmark$&$\checkmark$&$\times$&$\times$\\
        Pika 1.5&24.10&-& 24&-&-&$\checkmark$&$\checkmark$&$\times$&$\times$\\
        \midrule
        OpenSora-1.2 &24.06&Multi, max:720p&24&102 (max:408)&102 (max:408)&$\checkmark$&$\checkmark$&$\times$&$\checkmark$\\
        EasyAnimate-v4&24.08&Multi, max:$1024^{2}$&24&144&144&$\checkmark$&$\checkmark$&$\checkmark$&$\checkmark$\\
        CogVideoX-5B&24.08&720$\times$480&8&49&49&$\checkmark$&$\checkmark$&$\times$&$\checkmark$\\
        
        \bottomrule
    \end{tabular}}}
    \label{tab:cost1}
\end{table}

\paragraph{Qingying (Zhipu)~\cite{zhipu2024qingying}} was released in late July 2024, boasting features such as text-to-video and image-to-video (using a start image) functionalities on its App and website. The videos it generates are 6 seconds at 16 fps. As per the details provided in CogVideoX \cite{yang2024cogvideox}, this model was also developed using a large-scale diffusion transformer model for generating videos from text prompts. The key architecture designs include a 3D Causal Variational Autoencoder (VAE), an Expert Transformer equipped with adaptive LayerNorm to enhance text-video alignment, Explicit Uniform Sampling, and progressive training to yield coherent, long videos with dynamic motions. Additionally, an effective text-video data processing pipeline has been introduced to improve the quality of video captions, video quality, and semantic alignment.

\paragraph{Hailuo (MiniMax)~\cite{minimax2024hailuo}} was released in early September 2024 with only a text-to-video function on the website. The generated videos are 5 seconds with 25 fps. There is a limited introduction to the techniques. The key features are high compression efficiency, strong text alignments, complex motion, emotion generation, and diverse stylistic generalization capability. These features enable the model to generate high-resolution, high-frame-rate videos with a cinematic quality. 

Notably, due to the insufficient information about these closed-source models, including various hyperparameters that control the generation results, as products, they may have some special considerations for the model's settings. This can give the model certain characteristics, such as exaggerated motion shots, video styles, etc., which can also influence the generated results to some extent. Therefore, these results potentially have a certain degree of bias.

\paragraph{Wanxiang (Ali Tongyi)~\cite{tongyi2024wanxiang}} was released in late September 2024 with text-to-video and image-to-video (using a start image) functionalities with \emph{audio effects} on the website. The generated videos are 5 seconds with 30 fps. The key features support native 1080P 20s video generation, powerful motion generation and concept combination capability, and master a variety of artistic styles. In addition, this is the first model to generate video and the corresponding audio effects together. However, due to a lack of references, the method of generating the audio is unclear.

\paragraph{Pika (Pika Labs)~\cite{pika2024pika}} has released its new version 1.5 in early October 2024 with more realistic movement, big screenshots, and mind-blowing "Pikaffects" that break the laws of physics. The videos it generates are 5 seconds at 24 fps. Specifically, the released function is an image-to-video generation with six well-defined special effects, such as inflate, melt, explode, squish, crush, and cake-ify it for the main object of the input image. There is only one choice for one video generation among effects. 

\paragraph{Ongoing Models.} 

i) In late September 2024, ByteDance released two models, Seaweed and PixelDance, demonstrating enhanced video creation capabilities through sophisticated multi-shot actions and complex interactions among multiple subjects. 
ii) In early October 2024, Meta introduced their advanced media foundation AI models: \emph{Movie Gen}~\cite{meta2024moviegen}, a cast of foundation generative models, encompassing a 30B DiT-based model that can generate both images and videos of up to 16 seconds at 16 fps, along with a 7B V2V model for video super-resolution. In addition, it introduces a 13B parameter foundation model for video- and text-to-audio generation named \emph{Movie Gen Audio}, which can generate 48kHz high-quality cinematic sound effects and music synchronized with the video or text inputs. These models are trained and fine-tuned to possess four key capabilities: text-to-video generation, personalized video generation, precise video editing, and audio generation.

We will continue to update these results until they are ready to be public, and we summarize the above model's information in Table \ref{tab:cost1} for a quick check.

\subsubsection{Open-source Models}

\paragraph{Open-Sora \cite{opensora}} was published in March 2024 as an open-source project for efficient text-to-video and image-to-video generation, including a complete data preprocessing, acceleration, training, and inference pipeline. Until June 2024, Open-Sora has progressed from version 1.0 to 1.2. In the latest Open-Sora 1.2, this model trained a 1.1B-parameter model on 30M data points (~80K hours), using 35K H100 GPU hours. It supports 0-16s video generation from 144p to 720p and various aspect ratios. The key improvements are i) 3D video compression network, first compressing the video in the spatial dimension by 8x8 times, then compressing the video in the temporal dimension by 4x times; ii) implementations on several model adaptation and rectified flow \cite{liu2022flow,lipman2022flow} instead of DDPM following stable diffusion 3 \cite{esser2024sd3}, a Multimodal Diffusion Transformer (MMDiT) text-to-image model; iii) more data, better caption, and three-stage multi-scale training.

\paragraph{Open-Sora-Plan \cite{yuan_opensora}} was also published in March 2024 as an open-source project for text-to-video and image-to-video generation. Until August 2024, Open-Sora-Plan has progressed from version 1.0.0 to 0.1.2. The latest Open-Sora-Plan 0.1.2 utilizes a 3D full attention architecture instead of 2+1D and releases a true 3D video diffusion model trained on 4s 720p. The key improvements are: i) better compressed visual representations, they optimized the structure of CausalVideoVAE, which now delivers enhanced performance and higher inference efficiency; ii) better video generation architecture, they used a diffusion model with a 3D full attention architecture, which provides a better understanding of the world. The two Open-Sora projects were originally built based on Latte's~\cite{ma2024latte_videogen} source code and research findings.

\paragraph{EasyAnimate \cite{xu2024easyanimate}} was published in April 2024 as an open-source project for text-to-video, image-to-video, and video-to-video generation. Until September 2024, EasyAnimate was updated to the fourth version, supporting a maximum resolution of 1024x1024, 144 frames, 6 seconds, and 24fps video generation or 1280x1280 with video generation at 96 frames. The input can be text, images, and videos. The codebase also provides detailed data preprocessing, slice VAE, and DiT training. In addition, EasyAnimate introduces a specialized motion module, named the Hybrid Motion Module, to guarantee uniform frame production and transition of movements.

\paragraph{CogVideo-X \cite{yang2024cogvideox}} was published in August 2024 as an open-source project for text-to-video and image-to-video generation, the open-source version of the Qingying, with two versions CogVideoX-2B and CogVideoX-5B. The larger model exhibits higher quality and better visual effects with higher inference costs. It supports generating 48-frame videos at a resolution of 720x480 with 8fps. From the report~\cite{yang2024cogvideox}, CogVideo-X-5B significantly surpasses the previous open-source models (\emph{i.e.}, Open-Sora v1.2~\cite{opensora} and VideoCrafter2~\cite{chen2024videocrafter2}) from the various generation qualities on human actions, multiple objects, dynamic quality, and GPT4o-MTScore. Qingying and CogVideoX are the only paired closed-source and open-source models, making analysis and comparison easier.

From our comprehensive evaluation among existing \emph{open-sourece} DiT-based models, CogVideo-X-5B shows consistently better spatio-temporal generated quality. Thus, we mainly demonstrate CogVideo-X-5B results compared with \emph{closed-source} DiT-based models.

\subsection{Evaluation Process}

\subsubsection{Input Preparation}
We systematically design the input text and image prompts using the following strategies.

\begin{itemize}
\item Existing vertical-application video generation tasks, especially for the image-to-video task, such as 1) pose-controllable character and portrait animation \cite{hu2024animate_anyone,ma2024emoji}, 2) audio-driven portrait and upper-body gesture video generation~\cite{xu2024vasa,tian2024emo,lin2024cyberhost,corona2024vlogger}, 3) embodied avatar synthesis \cite{lin2024cyberhost,corona2024vlogger,tian2024emo,xu2024vasa}, 4) robot operation generation \cite{wang2024language, xiang2024pandora}, 5) animation video generation \cite{xing2024tooncrafter}, 6) world model simulator \cite{xiang2024pandora,yang2024worldgpt,bruce2024genie}, 7) autonomous driving \cite{gao2024vista,wen2024panacea, yang2024drivearena, zhao2024drivedreamer}, and 8) camera-controllable image animation \cite{wang2024humanvid,wang2024motionctrl,he2024cameractrl}.

\item Diverse objectives in videos to evaluate the visual and motion quality, including various functionalities (\emph{e.g.}, text-alignment, composition, transition, creativity, stylization, consistency, text generation, stylization, stability, interaction and relation, styles, reasoning ability, etc.) and contents (\emph{e.g.}, human, animal, object, city, nature, culture, etc.)
This part is similar to existing benchmarks \cite{ji2024t2vbench,feng2024tc_bench,sun2024t2v_bench,hu2023t2v_benchmark,li2024videoeval_bench,liu2024fetv_bench,liu2024evalcrafter_bench,huang2024vbench,mao2024tavgbench} but different prompts and taxologies. Our prompts are longer, more diverse, and provide more detailed descriptions, with 31 words on average. We also conduct a risk assessment test, including violence and hate, political topics, misinformation, privacy, discrimination, and the generation of pornographic content.

\item Prompts are collected by hundreds of online users and creators, and then we put these prompts into GPT-4o to re-write 50 new prompts with all mentioned compositional elements.

\item Prompts from ten kinds of real-life video applications, including advertising film or movie, anime, game, education, autonomous driving, embodied robots, documentaries, eat shows, and short videos, with extracting the first image and the corresponding motion descriptions.

\end{itemize}

\subsubsection{Model Setting Details}
Notably, due to the unavoidable instability of the generation process, all of our experiments used once-generated results without cherry-picking videos, except for experiments exploring the stability of multiple generations. Additionally, all closed-source models can be regarded as products whose performance and features will be affected by different model preference designs.

\begin{itemize}
\item For Kling 1.0, it has two kinds of models (\emph{i.e.}, a high-efficiency model and a high-performance model) with either 5-second or 10-second outputs. We use a 5-second high-efficiency model with enhanced prompts by default for comparisons, and partial results are generated by a high-performance (highlighted via \emph{HP}) model. 
\item For Kling 1.5, we evaluate the 5-second high-performance model with enhanced prompts.
\item For Dream Machine, we evaluate both the initial 1.0 version and the recent 1.6 version of Dream Machine with enhanced prompts.
\item For Gen-3, most text-to-video generation uses 5-second Gen-3 Alpha with enhanced prompts, and partial text-to-video videos are 10 seconds. Most image-to-video generation uses 5-second Gen-3 Alpha, and partial videos are from the Gen-3 Alpha Turbo model due to its fast and competitive performance. We also evaluate the video-to-video function and explore further usages for the research community.

\item For Vidu, most of our evaluation results are 4-second videos with enhanced prompts and the general style. Notably, Vidu provides the upscaling function after the video generation; some imperfections will be corrected, and the clarity will be improved. However, due to the additional efforts and costs, we simply use the original output videos without the upscaling.

\item For MiniMax, since only a text-to-video function is available, we evaluate text-to-video performance at this stage.

\item For Qingying, similar to Kling, it has two kinds of models (\emph{i.e.}, a high-efficiency model and a high-performance model) with 6-second outputs. We use a 6-second high-efficiency model with enhanced prompts by default for comparisons.

\item For Wanxiang, we generate the video with audio and the inspiration mode by default. The aspect ratio of the text-to-video generation is 16:9.

\item For Pika 1.5, we only evaluate the provided six special effects and will add the results until the new text-to-video model is released.

\item Notably, for image-to-video generation, Gen-3 only supported 1280x768 resolution before September 2024, Qingying supported 1440x960, while input images for human animation tend to be vertical. Thus, we reshaped the input image to make the input video visible.

\item For open-source models, we evaluate three models, Open-Sora 1.2, EasyAnimate v4, and CogVideo-X-5B, on text-to-video and image-to-video tasks.

\end{itemize}

\subsubsection{Model Results and Comparisons}

We select and demonstrate part of the generated videos in the following sections. \textbf{For each case, we provide its task (\eg, I2V), test ID number, and the prompt in the caption.}

\textbf{Visualization.} To observe the visual quality, we put comprehensive qualitative results in this report. For each video, we \emph{uniformly} sample the generated frames. This is a relatively fair way for comparison, but it may lead to some key generated frames being ignored in the report. However, the motion quality and much detailed information are still hard to reflect in static images; thus, \textbf{we encourage the readers to watch the generated videos on our website}\footnote{https://ailab-cvc.github.io/VideoGen-Eval/}.

\textbf{Comparison.} Beyond the SORA-like models, we also demonstrate the frames from the professionally generated videos (\emph{i.e.}, from movie, animation, and advertising videos) or generated videos from vertical-application models (\emph{e.g.}, Animate Anyone \cite{hu2024animate_anyone} for pose-controllable image animation).

\textbf{Output.} We provide all the text, image prompts, and corresponding generated videos to serve further research. In the future version, we will keep exploring effective quantitative evaluation from subjective metrics (\emph{e.g.}, temporal consistency, numeracy, and multi-shot detection) and multi-video Arena comparisons.

\section{General-purpose SORA-like Models v.s Vertical-Domain Models}
\label{sec:2}

With the development of generic base models for images and videos (\emph{e.g.}, SD, SVD, DiT) to better adapt to the usage requirements of draping scenarios, including fine-grained controllability and high-quality video generation, video generation models in various focused draping domains have emerged. However, most of them are still based on UNet-based foundation models and small-scale data finetuning. Actually, the upper limit of the vertical model’s capability largely depends on the foundation model. With the rapid progress of DiT-based models, we derive a concern about whether better-generalized models can now surpass the existing vertical-domain models in some aspects, and we can further look ahead to the existing models by their generated results. Where the upper limit of the vertical-domain model is, this motivates us to rethink the definition of existing tasks, especially the problems that are not previously focused on or are difficult to solve. 

\begin{figure}[h]
	\centering
	\small
        \begin{overpic}[width=1\linewidth]{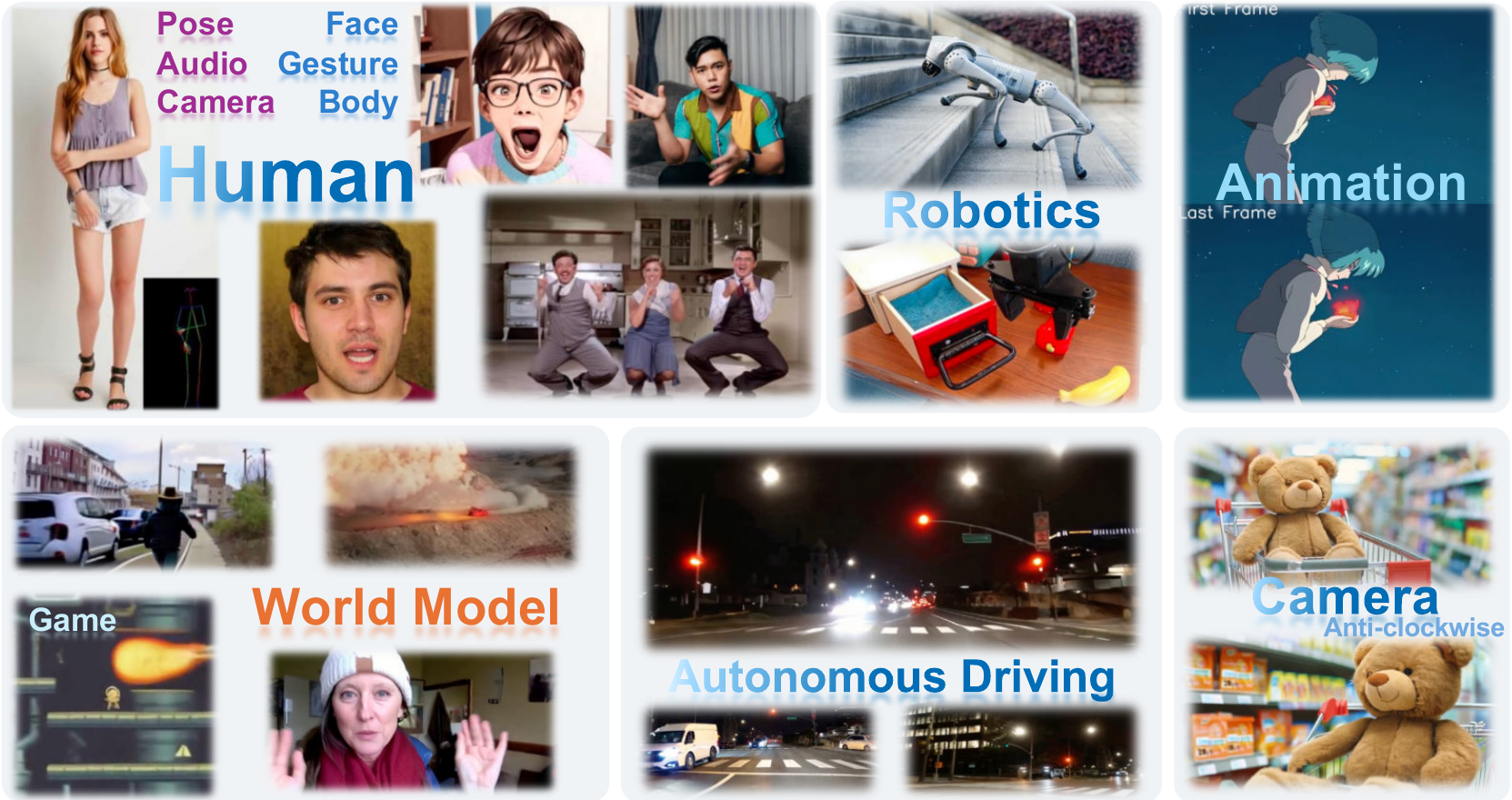}

	\end{overpic}
    \vspace{-0.4cm}
	\caption{Overview of Section~\ref{sec:2}. We compare with existing vertical-domain video models, including human-centric animation, robotics, cartoon animation, world models, autonomous driving, and camera controls in the video generation area.}
 \label{Fig:sec2_teaser}
     \vspace{-0.5cm}

\end{figure}

Based on existing vertical-domain video generation tasks, as illustrated in Figure \ref{Fig:sec2_teaser}, we explore the performance of recent SORA-like models on these tasks. 

\subsection{Human Video Generation}
\label{Sec:Human}
Human video generation aims to synthesize 2D human-centric video sequences under various control conditions, such as pose, audio, camera, and text with an initial input image~\cite{lei2024comprehensive_human_survey,hu2024animate_anyone,wang2024humanvid,ma2024emoji,tian2024emo,corona2024vlogger,xu2024vasa}. This process requires translating these conditions into a video that exhibits natural, temporal-aligned, and dynamic motions with full-body, upper-body, or face appearances. The ability to generate such videos holds immense potential for film, advertisement, games, and virtual communication applications. In this area, creating natural, controllable, realistic, and dynamic human is crucial.

Recent advancements in generic generative image and video models (\emph{e.g.}, Stable Diffusion~\cite{rombach2022sd}, Stable Video Diffusion~\cite{blattmann2023svd}) have significantly contributed to progress, yet human video generation still faces substantial challenges. These include maintaining appearance and motion consistency, addressing finger abnormalities, handling temporal and semantic misalignment, and ensuring natural motion transitions. The complexity of human motions and emotions, low-resolution appearance regions, lack of high-quality data, insufficient capabilities of foundation models, and the tendency to overlook interactions with the environment and objects further compound these difficulties. 

\begin{figure}[!ht]
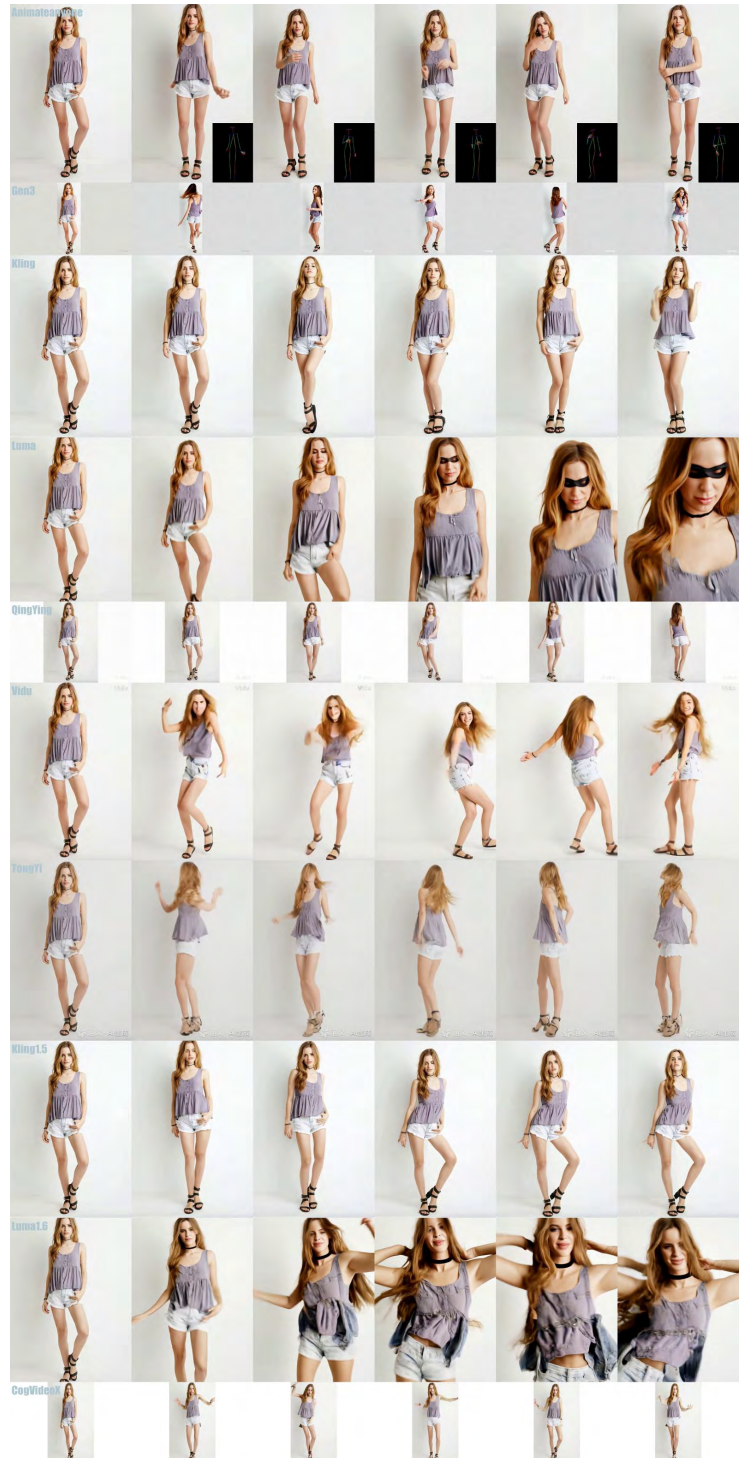

 \vspace{-0.3cm}
	\centering
	\small
        \begin{overpic}[width=0.7\linewidth]{Figs/Sec2/00591.pdf}

	\end{overpic}
    \vspace{-0.3cm}
	\caption{\emph{Comparisons with the pose-controllable image animation (\emph{e.g.}, Animate-Anyone~\cite{hu2024animate_anyone})}. Prompt: (I2V-591) "The camera remains still, swinging the person's left and right hands back and forth. At the same time, the left and right feet move rhythmically." It is hard to generate continuous and complex actions solely through text control, meanwhile, there are still limitations in ID preservation.}
 \label{Fig:human_pose1}
 \vspace{-0.5cm}
\end{figure}

\begin{figure}[!ht]
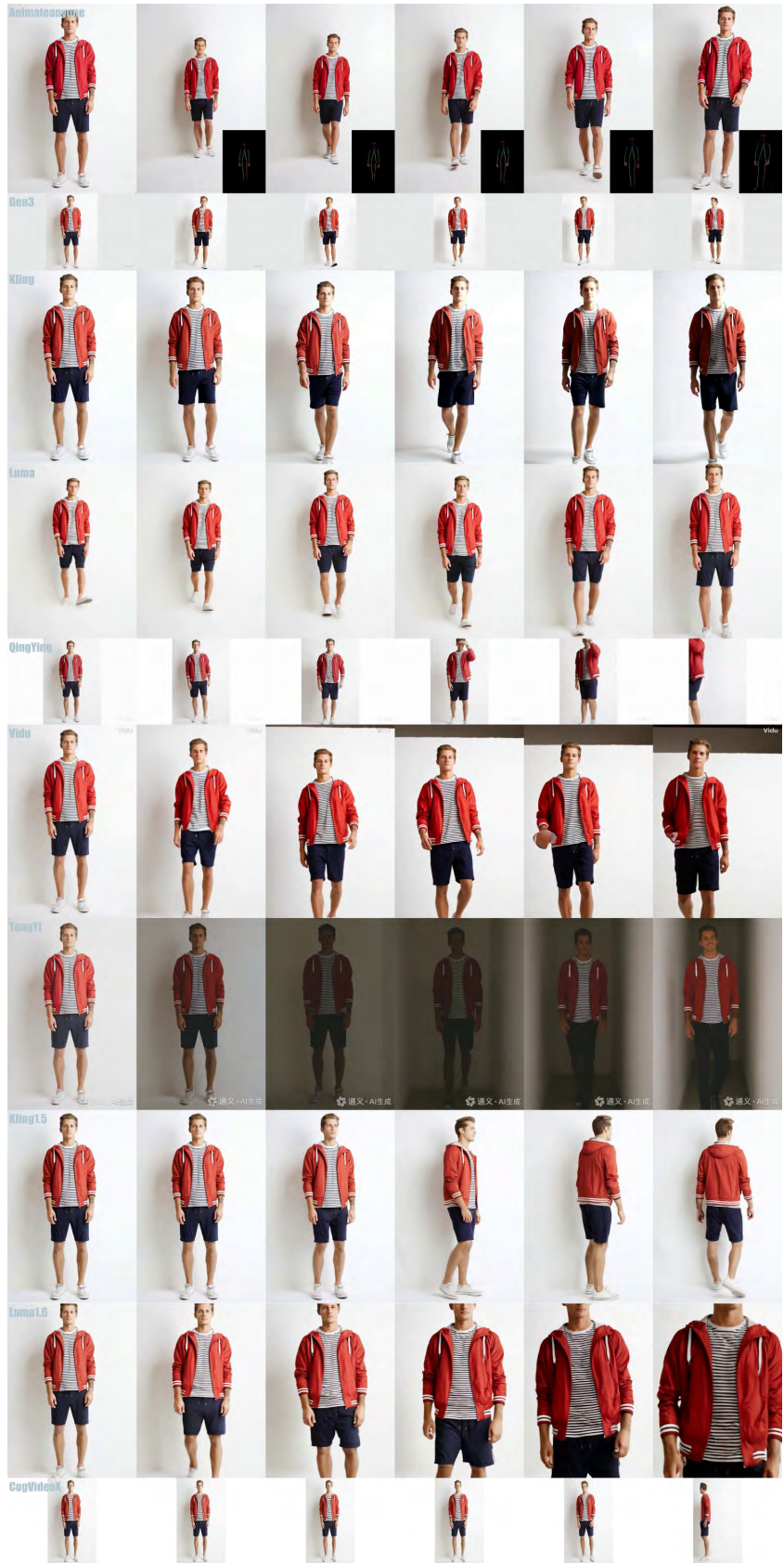

 \vspace{-0.5cm}
	\centering
	\small
        \begin{overpic}[width=0.8\linewidth]{Figs/Sec2/00593.pdf}

	\end{overpic}
  \vspace{-0.2cm}
    \caption{\emph{Comparisons with the pose-controllable image animation (\emph{e.g.}, Animate-Anyone~\cite{hu2024animate_anyone})}. Prompt: (I2V-593) "The camera stays still as the man walks to the camera from a distance." When performing simple motions such as walking, most models can generate plausible results, but some may generate actions but do not follow the direction of the instructions, \eg, QingYing and Kling1.5.}  
    \label{Fig:human_3}
     \label{Fig:human_pose2}
\end{figure}

\begin{figure}[!ht]
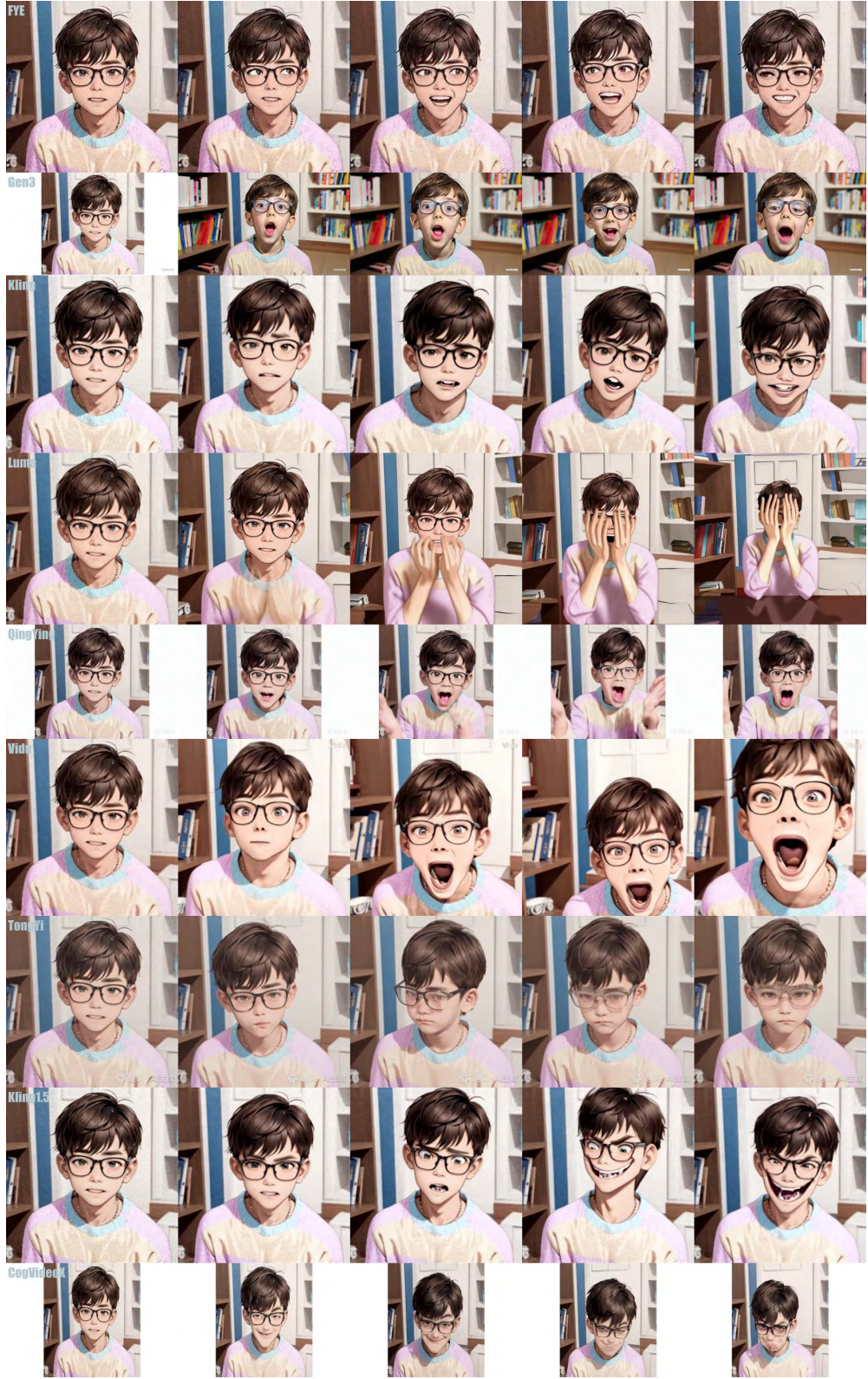

	\centering
	\small
        \begin{overpic}[width=0.95\linewidth]{Figs/Sec2/00598.pdf}

	\end{overpic}
	\caption{\emph{Comparisons with the pose-controllable portrait animation (\emph{e.g.}, Follow-your EMOJI~\cite{ma2024emoji}) in a photo-realistic style}. Prompt: (I2V-598) "The boy makes an exaggerated expression on his face." The models have generally generated content that aligns with the intended facial expressions, but it is difficult to maintain facial identity under large expressive movements. 
 }
  \label{Fig:human_pose5}
\end{figure}

\begin{figure}[!ht]
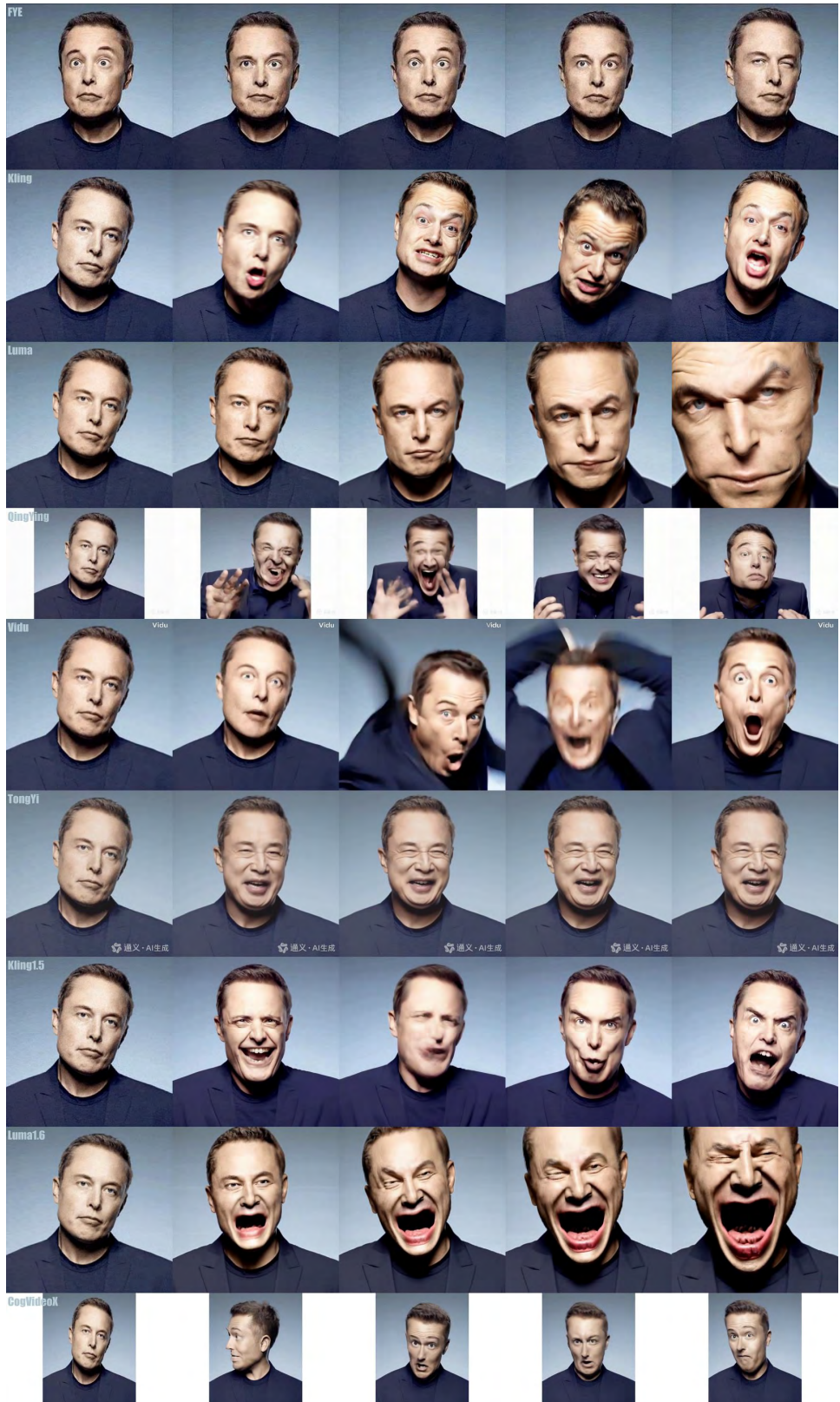

\vspace{-0.3cm}
	\centering
	\small
        \begin{overpic}[width=0.92\linewidth]{Figs/Sec2/00599.pdf}

	\end{overpic}
    \vspace{-0.3cm}
	\caption{\emph{Comparisons with the pose-controllable portrait animation (\emph{e.g.}, Follow-your EMOJI~\cite{ma2024emoji}) in a photo-realistic style}. Prompt: (I2V-599) "The man makes an exaggerated expression on his face." Almost all models fail to maintain the ID and natural motions. Vidu performs well but exhibits some unnecessary actions beyond instructions.}
  \label{Fig:human_pose6}
\end{figure}

\begin{figure}[!ht]
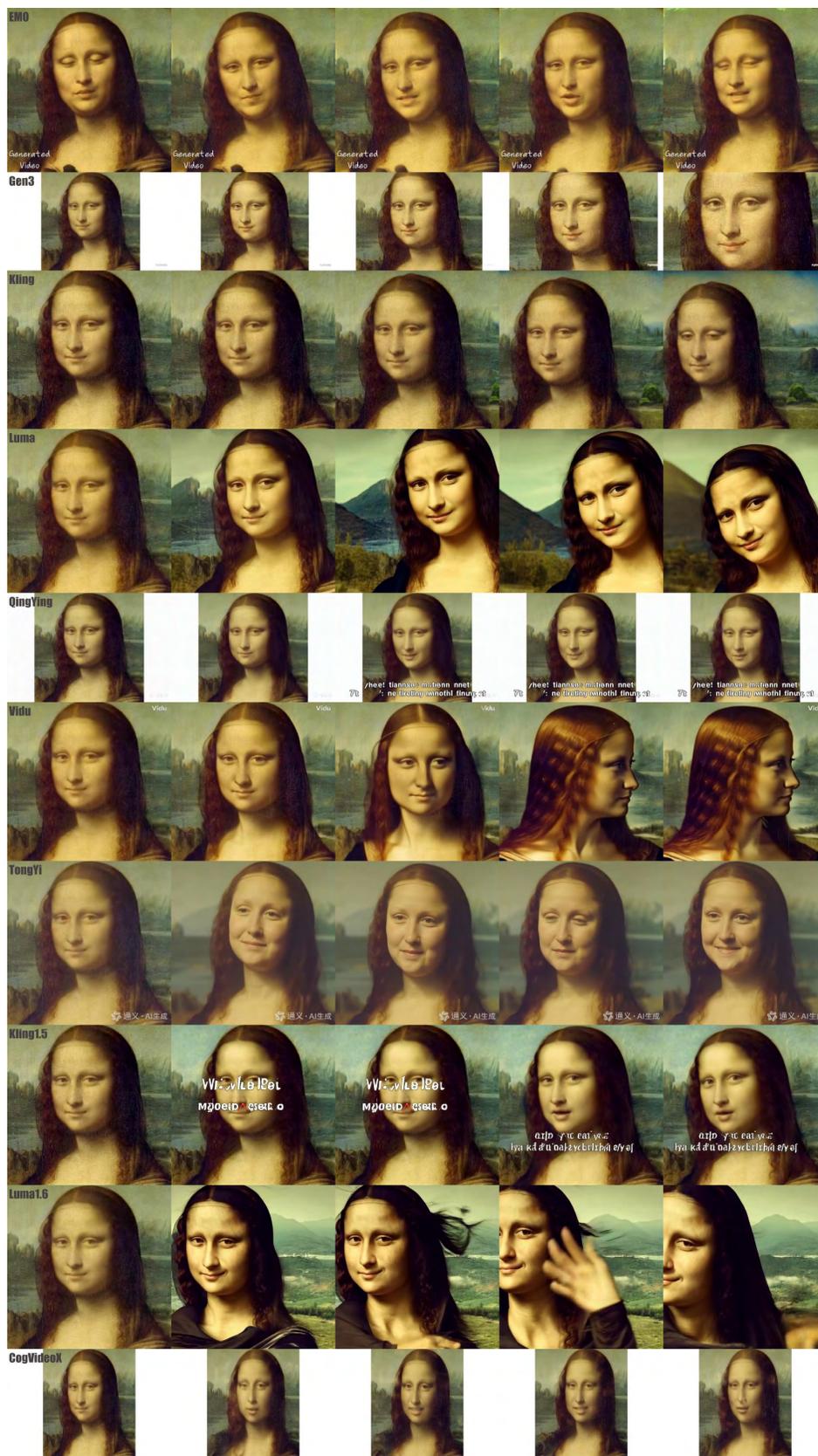

\vspace{-0.3cm}
	\centering
	\small
        \begin{overpic}[width=0.9\linewidth]{Figs/Sec2/00603.pdf}

	\end{overpic}
 \vspace{-0.3cm}
	\caption{\emph{Comparisons with the audio-driven portrait animation (\emph{e.g.}, EMO~\cite{tian2024emo})}. Prompt: (I2V-603) "The woman is saying the following: ``Yes, one; and in this manner.'' We try to convert the speech contents to text input to explore if existing models could generate corresponding lip movements of the speech. However, all the models even fail at the motion of speaking (motion incompleteness).}   
 \vspace{-0.3cm}
 \label{Fig:human_audio2}
\end{figure}

\begin{figure}[!ht]
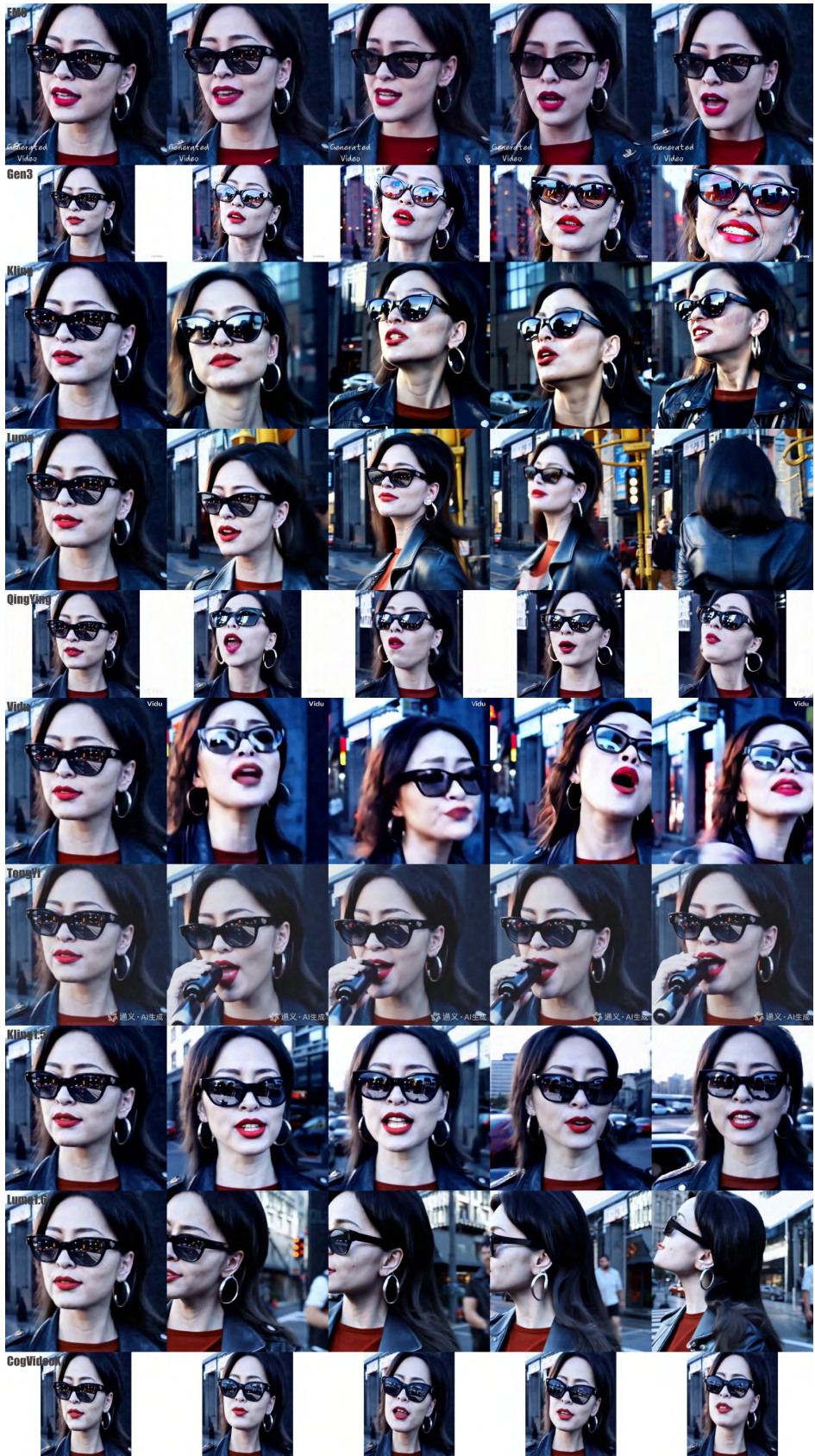

\vspace{-0.3cm}
	\centering
	\small
        \begin{overpic}[width=0.9\linewidth]{Figs/Sec2/00605.pdf}

	\end{overpic}
    \vspace{-0.3cm}
	\caption{\emph{Comparisons with the audio-driven portrait animation (\emph{e.g.}, EMO~\cite{tian2024emo})}. Prompt: (I2V-605) "The woman is singing." Models generate results with open mouths, but some do not sing \eg, Luma, and others hallucinate and generate a microphone.} 
 \label{Fig:human_audio3}
\end{figure}

\begin{figure}[!ht]
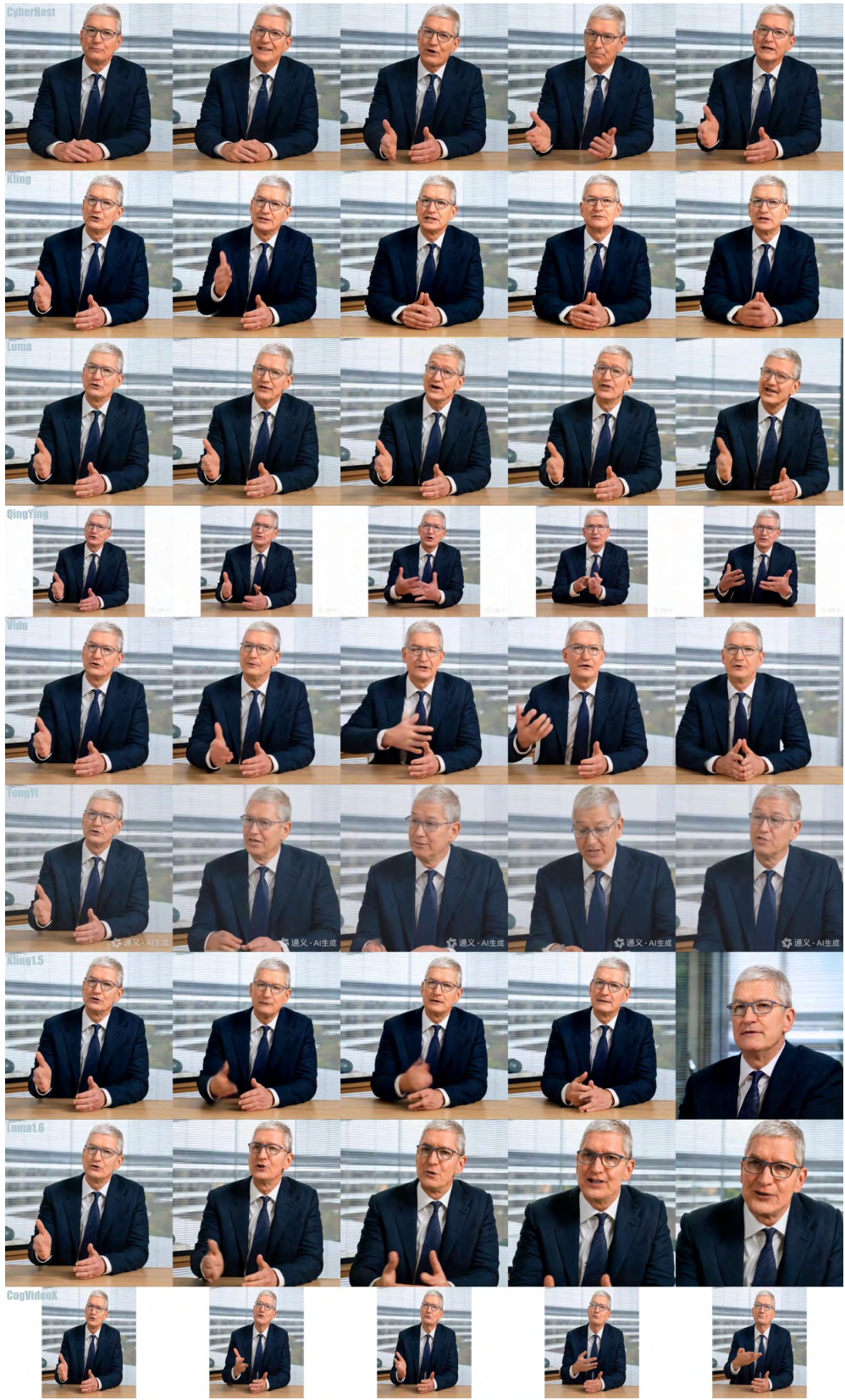

\vspace{-0.3cm}
	\centering
	\small
        \begin{overpic}[width=0.95\linewidth]{Figs/Sec2/00607.pdf}

	\end{overpic}
	\caption{\emph{Comparisons with the audio-driven co-speech gesture animation (\emph{e.g.}, CyberHost~\cite{lin2024cyberhost})}. Prompt: (I2V-607) "He's talking, accompanied by gesture changes." All the models generate plausible results, with Luma showing slightly less variation in gesture changes.}
 \label{Fig:human_audio5}
\end{figure}

\begin{figure}[!ht]
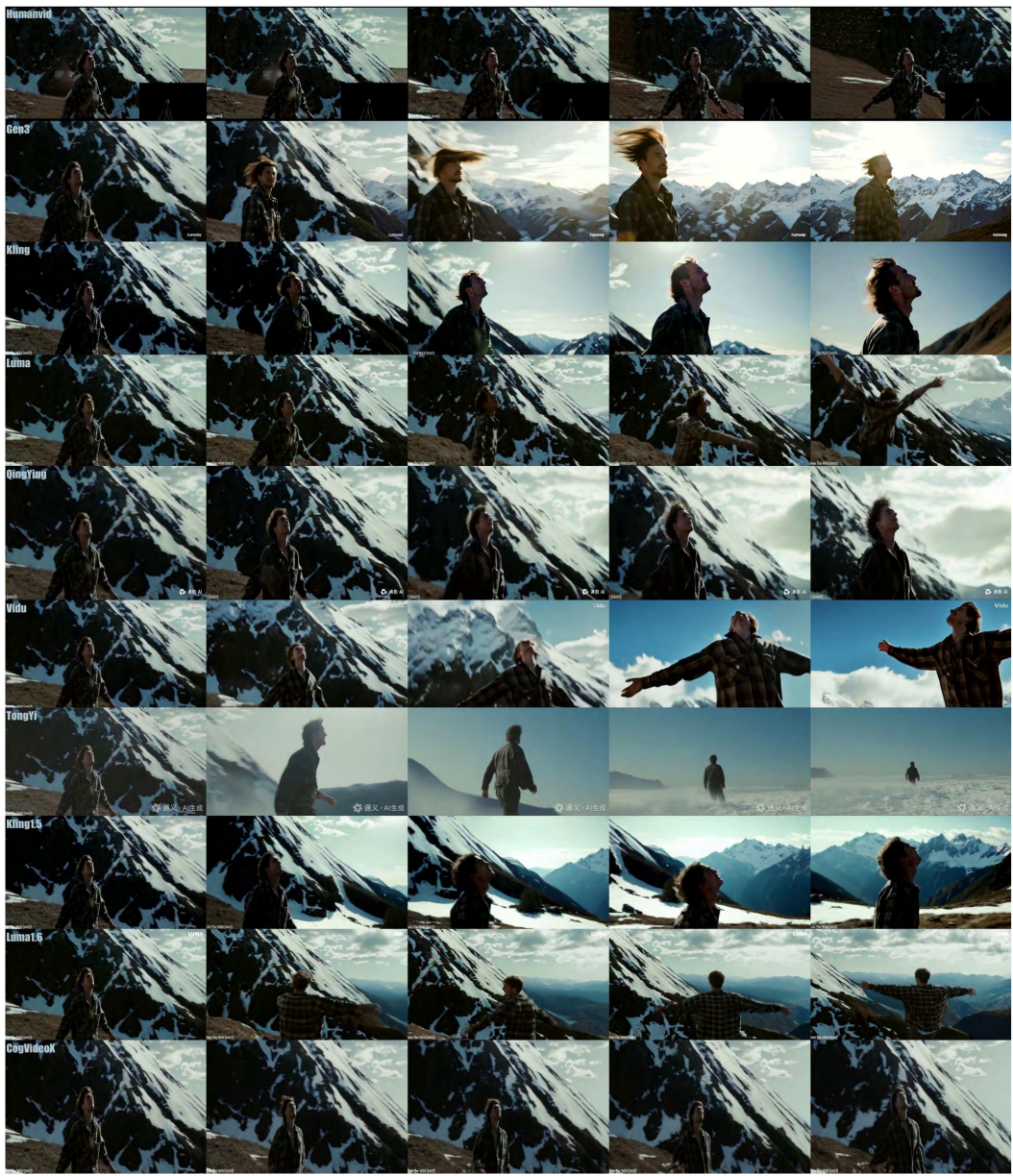

	\centering
	\small
        \begin{overpic}[width=1.\linewidth]{Figs/Sec2/00611.pdf}

	\end{overpic}
	\caption{\emph{Comparisons with the pose-driven and camera-controllable human image animation (\emph{e.g.}, HumanVid~\cite{wang2024humanvid})}. Prompt: (I2V-611) "camera move right, The wind is blowing this man, \emph{the man keeps his motion.}" With only camera motions moving, all existing generated videos struggle to keep the appearance and motion of the person. The more precise prompts will obtain better results; specifically, see the results from Vidu.}
 \label{Fig:human_camera2}
\end{figure}

\begin{figure}[!ht]
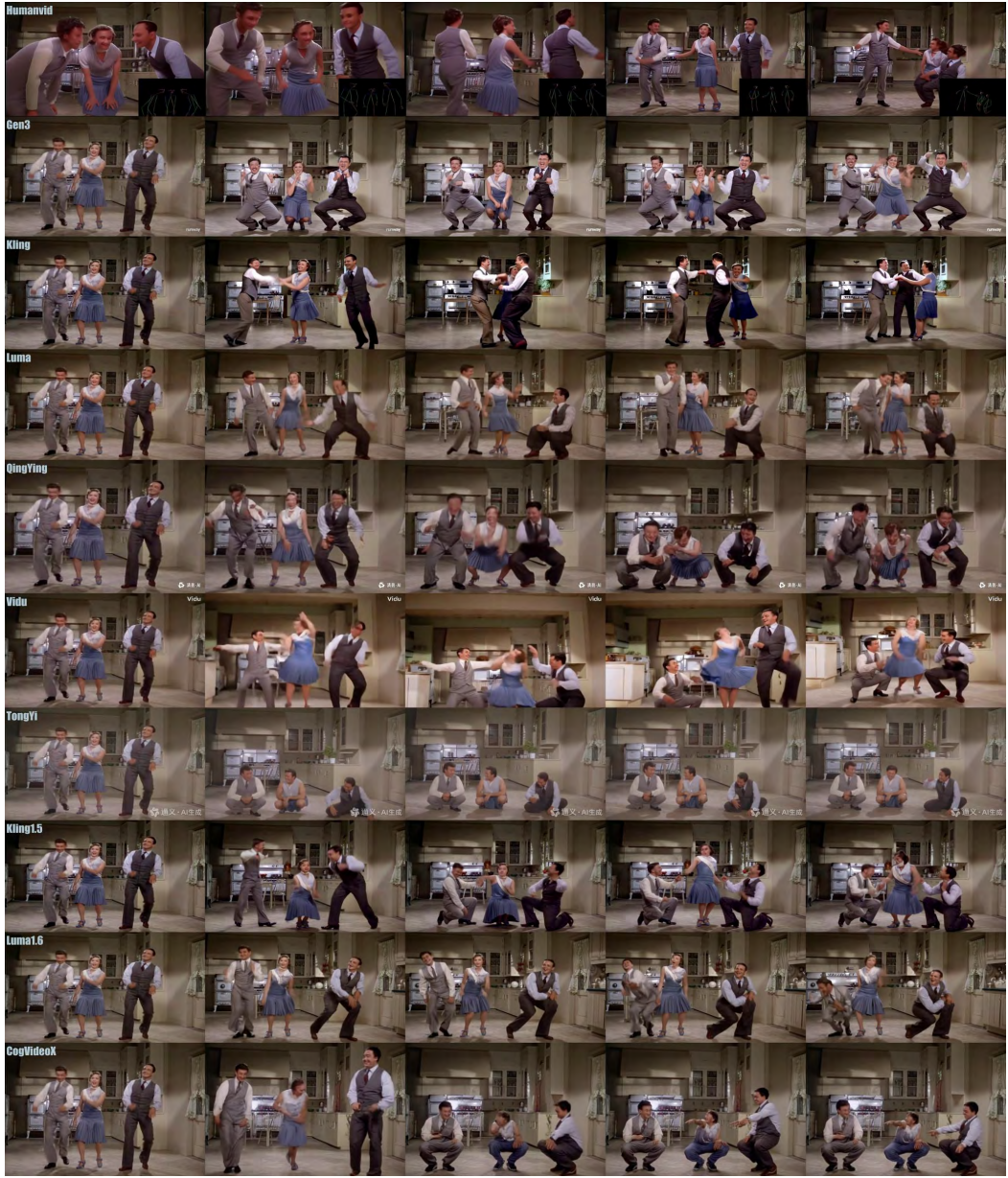

	\centering
	\small
        \begin{overpic}[width=1.\linewidth]{Figs/Sec2/00610.pdf}

	\end{overpic}
	\caption{\emph{Comparisons with the pose-driven and camera-controllable human image animation (\emph{e.g.}, HumanVid~\cite{wang2024humanvid})}. Prompt: (I2V-610) "The three persons talked and laughed and turned to the right together, then the two persons on the right squatted down, and the man on the left pointed to the two persons on the right." The models still have limited ability to generate complex motion for multi-person, none of them generate content according to the instructions.} 
 \label{Fig:human_camera3}
\end{figure}

In this part, we explore the image animation or image-to-video generation abilities of SORA-like models following the input settings of previous representative works. Since the existing model does not support sequential human poses or audio inputs, we describe the motion in words as closely as possible to what it would have been, so the corresponding generated actions may be difficult to reenact the original model. We summarize the key observations on various tasks from pros and cons:
\begin{itemize}

 \item \emph{Pose-controllable human body generation} (Figure \ref{Fig:human_pose1}, \ref{Fig:human_pose2}). The given two cases evaluate the different motion generation capabilities. Kling~\cite{kuaishou2024kling} 1.0 consistently generates text-aligned and robust videos, even simulating physically plausible light changes as the person walks to the camera. But Kling 1.5 fails to generate the text-aligned motions in Figure \ref{Fig:human_pose2}, which may indicate the instability of the generation process. Vidu~\cite{bao2024vidu} tends to generate highly dynamic motions while failing to maintain a consistent appearance and natural limb movements. Although the prompt describes that the camera stays still, Luma~\cite{luma2024dm} tends to generate videos with a moving camera (\emph{e.g.}, zoom in), which may reflect the motion incompleteness of the out-of-distribution motions based on the given image. Other models generate worse results due to different degrees of uncontrollable motions, inconsistent appearance, motion blur, low visual quality, etc. Hopefully, we find that the best models here (\emph{e.g.}, Kling), have significantly improved in terms of the naturalness of motion, and the finer details of body movements, including face and finger quality.

 \item \emph{Pose-controllable portrait image generation} (Figure \ref{Fig:human_pose5}, \ref{Fig:human_pose6}). The two cases evaluate the animation generalization abilities from a cartoon image to a real-life portrait image. Due to the high sensitivity of humans to facial details, driving facial expressions often demands higher accuracy, along with requirements for both appearance consistency and a balance between realistic and exaggerated expressive movements. Existing models show varying degrees of identity distortion and motion deformation in these aspects, making the results less controllable compared to ~\cite{ma2024emoji}.

 \item \emph{Audio-driven portrait animation} (Figure \ref{Fig:human_audio2}, \ref{Fig:human_audio3}). When given generic verb descriptions such as "talk" or "sing," these models can easily generate the corresponding actions, but without much actual meaning. Some models generate subtitles and hand information with hallucinations. Interestingly, when we tried providing prompts with speech-to-text contents (Figure~\ref{Fig:human_audio2}), hoping that the model would generate lip movements based on the input text, none of the models were able to produce speech motions. This indicates that present models do not incorporate the content of speech during captioning, similar to how text generation may require additional descriptions for dense text recognition and translation.

 \item \emph{Audio-driven co-speech gesture animation} (Figure \ref{Fig:human_audio5}). Due to concerns about portrait rights, some models refused to generate the content (\emph{e.g.}, Gen-3). In this case, most models can maintain a good identity consistency, as the action does not require significant movement. Compared to previous pose-controllable models, the SORA-like models avoid explicit pose controls, reducing errors in motion representation (such as skeletal errors during crossed-arm movements) and can generate more diverse and complex gestures.

 \item \emph{Pose-driven and camera-controllable image animation} (Figure \ref{Fig:human_camera2}). Given both human poses and camera movements, this task is more complex, testing the model's 3D modeling capability. Kling also generates better visual and motion quality with good text alignment. 
 Figure~\ref{Fig:human_camera2} adds "the man keeps his motion" to control the human's motion explicitly. After adding the text, Vidu controls the person's visibility, indicating the precise text should be given to control the main subjects.

 \item \emph{Multi-person image animation} (Figure \ref{Fig:human_camera3}). This case demonstrates dynamic motion and multi-person animation. Keeping all appearances with diverse motions is challenging due to occlusions and interactions. Gen-3~\cite{runway2024gen3} generates the result with better text alignment, as well as visual and motion quality. Many models fail to maintain consistency in appearance and generate several motion blur frames.

\end{itemize}

\clearpage

\subsection{Robotics}
\begin{figure}[!ht]
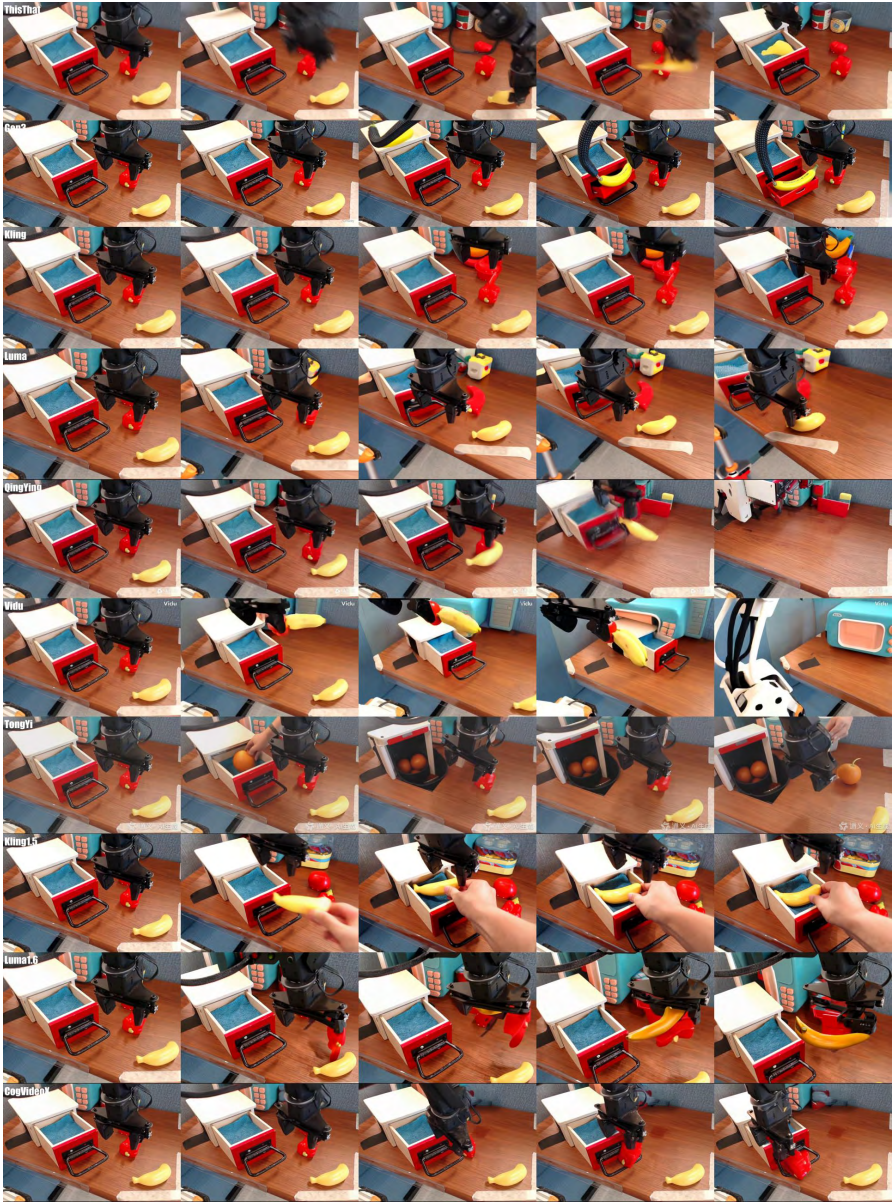

	\centering
	\small
        \begin{overpic}[width=0.85\linewidth]{Figs/Sec2/00404.pdf}

	\end{overpic}
	\caption{\emph{Comparisons with a robot action generation model (\eg, This \& That \cite{wang2024language})}. Prompt: (I2V-404) "the robotic arm puts the banana inside the drawer." As demonstrated, these models fail to execute instructions well, resulting in a banana that appears out of nowhere and a robotic arm that does not follow the plausible motion trajectory, indicating a serious issue in I2V models that do not understand the input image and tend to generate new objects.}
 \label{Fig:robot_1}
 \vspace{-0.4cm}
\end{figure}

\begin{figure}[!ht]
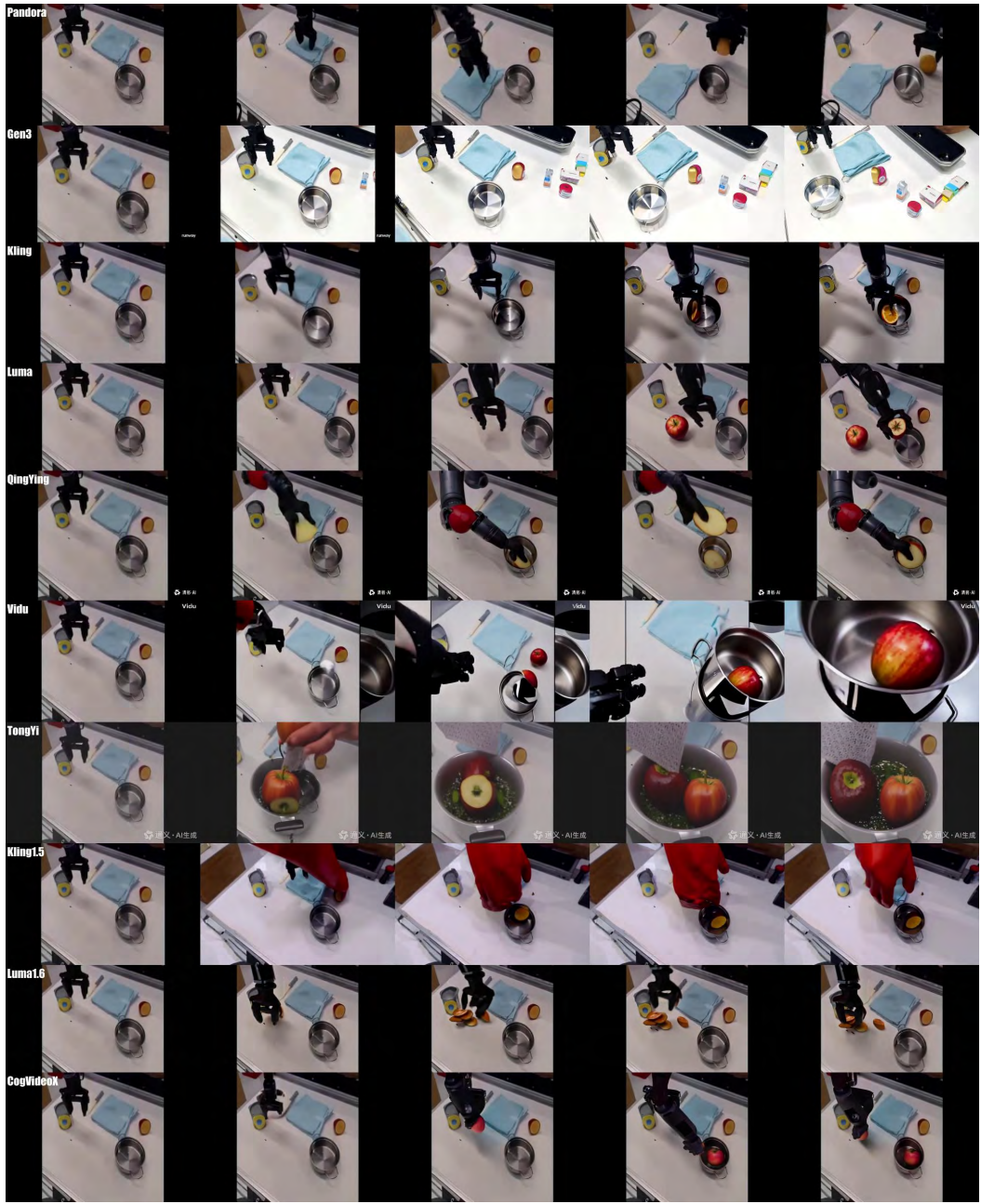

	\centering
	\small
        \begin{overpic}[width=1.\linewidth]{Figs/Sec2/00405.pdf}

	\end{overpic}
	\caption{\emph{Comparisons with a general world model (\eg, Pandora \cite{xiang2024pandora})}. Prompt: (I2V-405) "the robotic arm moves the towel, puts the apple into the pot, and takes the apple out of the pot." This instruction contains multiple operations, but the results show that these models did not complete even one operation, and method \eg, Vidu generates hallucinated camera movements.}
 \label{Fig:robot_2}
\end{figure}

Video diffusion models have demonstrated the potential to facilitate robotics, encompassing planning \cite{chi2024diffusionpolicy,du2023video}, generalization to new scenarios \cite{du2024learning}, and generating human actions for imitation \cite{liang2024dreamitate}. Among these methods, the video generation stage involves inputting instructions \eg, natural language and an image depicting the initial state \eg, a robotic arm in an operational environment; the model then generates videos illustrating the process of robot operations, especially trajectories of the configuration. Regarding this process, the key elements involve ensuring the consistency and suitable motion speed of the configuration, maintaining a fixed camera, and being semantic-aware of the given initial state (\eg, an image). We provide a comparison between the closed-source models and several specific methods \cite{wang2024language, xiang2024pandora}, please refer to Figure \ref{Fig:robot_1}, \ref{Fig:robot_2}. 

\clearpage

Existing I2V models fail to execute instructions, not to mention that the instruction contains multiple operations. There is no complete even one operation and interaction, and a robotic arm that does not follow the plausible motion trajectory. These results also indicate that existing I2V models find it difficult to understand the spatial relationships and object information in input images, especially when the input images contain out-of-distribution (OOD) objects or when the objects to be controlled are relatively small. This increases the difficulty of controlling and animating the spatial information.

\subsection{Cartoon Animation}
\begin{figure}[!ht]
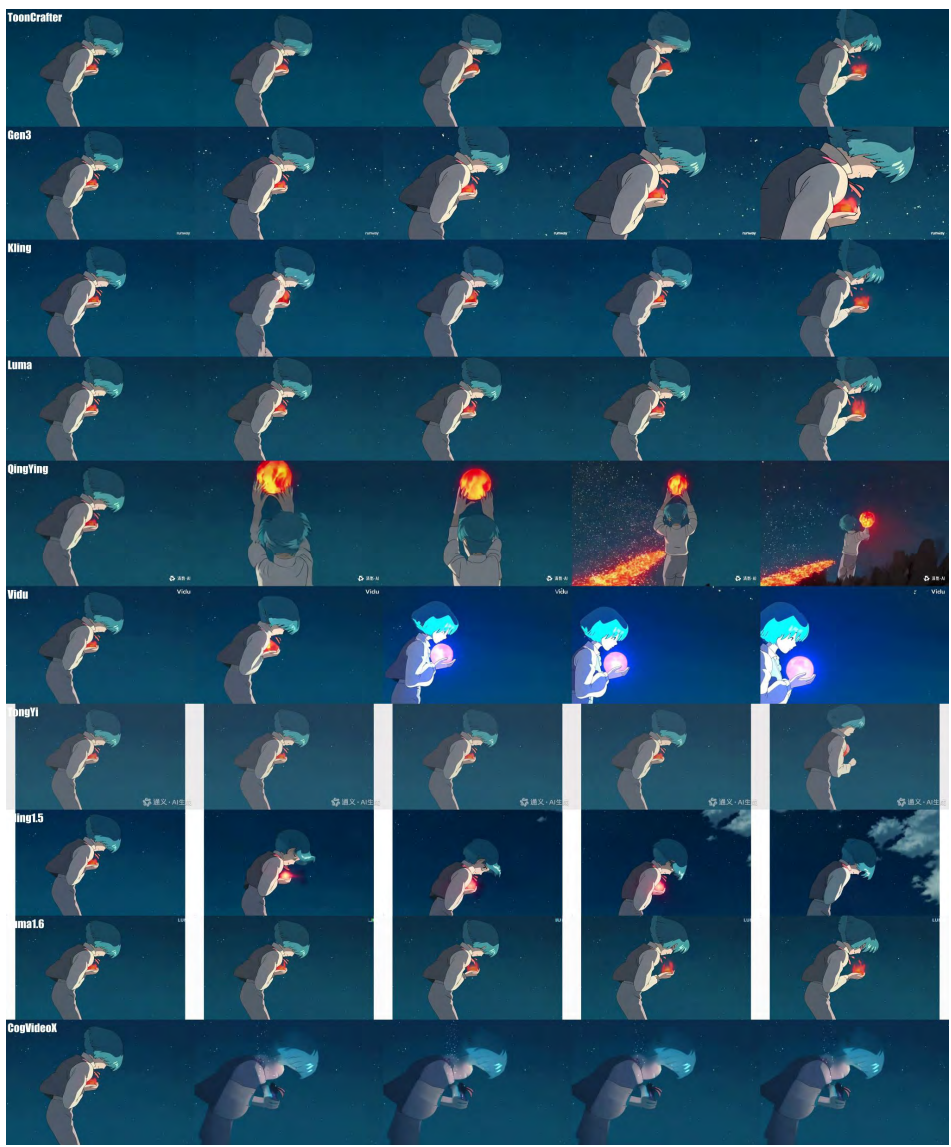

	\centering
	\small
        \begin{overpic}[width=0.9\linewidth]{Figs/Sec2/00615.pdf}

	\end{overpic}
	\caption{\emph{Comparisons with the generative cartoon Interpolation (\emph{e.g.}, ToonCrafter~\cite{xing2024tooncrafter})}. Prompt: (II2V-615): " ". No additional textual prompt. Most content can be generated reasonably, but some involve creative content, while others focus on camera movement. } 
 \label{Fig:anime1}
\end{figure}

\begin{figure}[!ht]
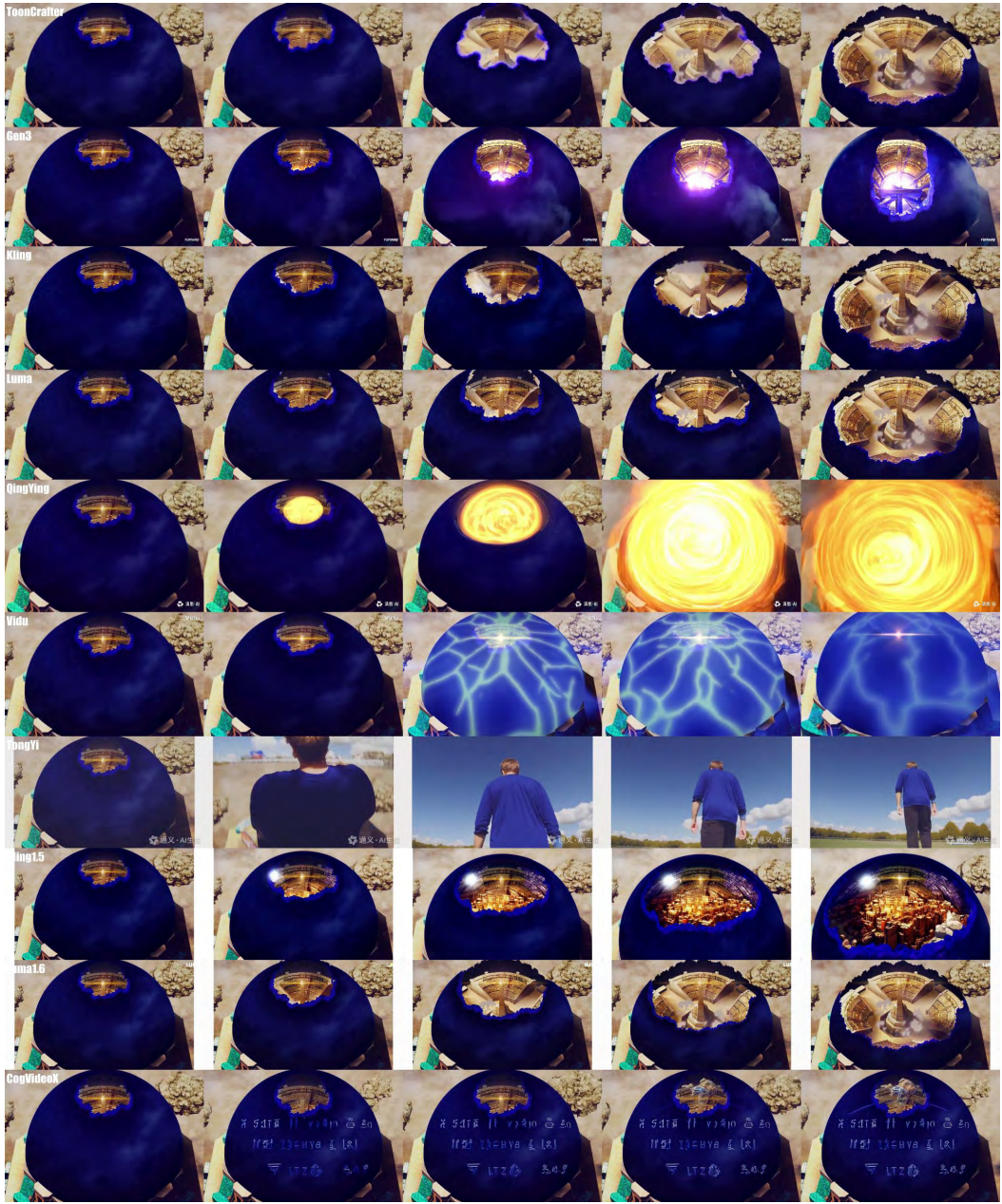

	\centering
	\small
        \begin{overpic}[width=1\linewidth]{Figs/Sec2/00617.pdf}

	\end{overpic}
	\caption{\emph{Comparisons with the generative cartoon Interpolation (\emph{e.g.}, ToonCrafter~\cite{xing2024tooncrafter})}. Prompt: (II2V-617): " ". No additional textual prompt. Gen-3, Kling, and Luma align with the content between the first and last frames, while QingYing, Vidu, and Kling 1.5 lean more toward creativity.} 
 \label{Fig:anime2}
\end{figure}

For animation, the high-frequency details of characters and scenes are relatively fewer compared to real-world scenarios. 
Besides, exaggerated expressions and actions do not come across as too abrupt. For this type of generation, in addition to maintaining character and motion consistency (mentioned in Sec. \ref{Sec:Human}), the ability to generate animations in various styles is a crucial aspect. 
Most efforts have concentrated on linear interpolation between two given images, assuming simple underlying motions. Typically, these methods identify correspondences, such as optical flow, between two frames and perform linear interpolation. However, this linear assumption fails to capture complex motions and heavy occlusions. Inspired by large-scale data-trained models that can generate diverse and realistic videos from images, some works~\cite{xing2024tooncrafter} explore how the rich motion priors learned from these video models can be leveraged for generative cartoon interpolation.
To further compare the effectiveness of existing SORA-like models in this task, we show some cases compared with the ToonCrafter \cite{xing2024tooncrafter}. The input involves the first and last frames (Figure \ref{Fig:anime1}, \ref{Fig:anime2}). Notably, some models do not provide the last frame function (\emph{i.e.}, Gen-3, Qingying, and Vidu); thus, we only input the first frame.

The two examples demonstrate that the naturalness and continuity of the motion interpolation from SORA-like models are good, particularly in cases where two controllable frames (start and end) are provided. For such scenarios, it is essential to define more challenging and practically meaningful evaluation test sets to continually explore the boundaries of the models. When given the first frame, some models (\emph{e.g.}, Kling 1.5) can even show creativity, generating content that surpasses the controllable video's richness.

\clearpage

\subsection{World Model}
\begin{figure}[!ht]
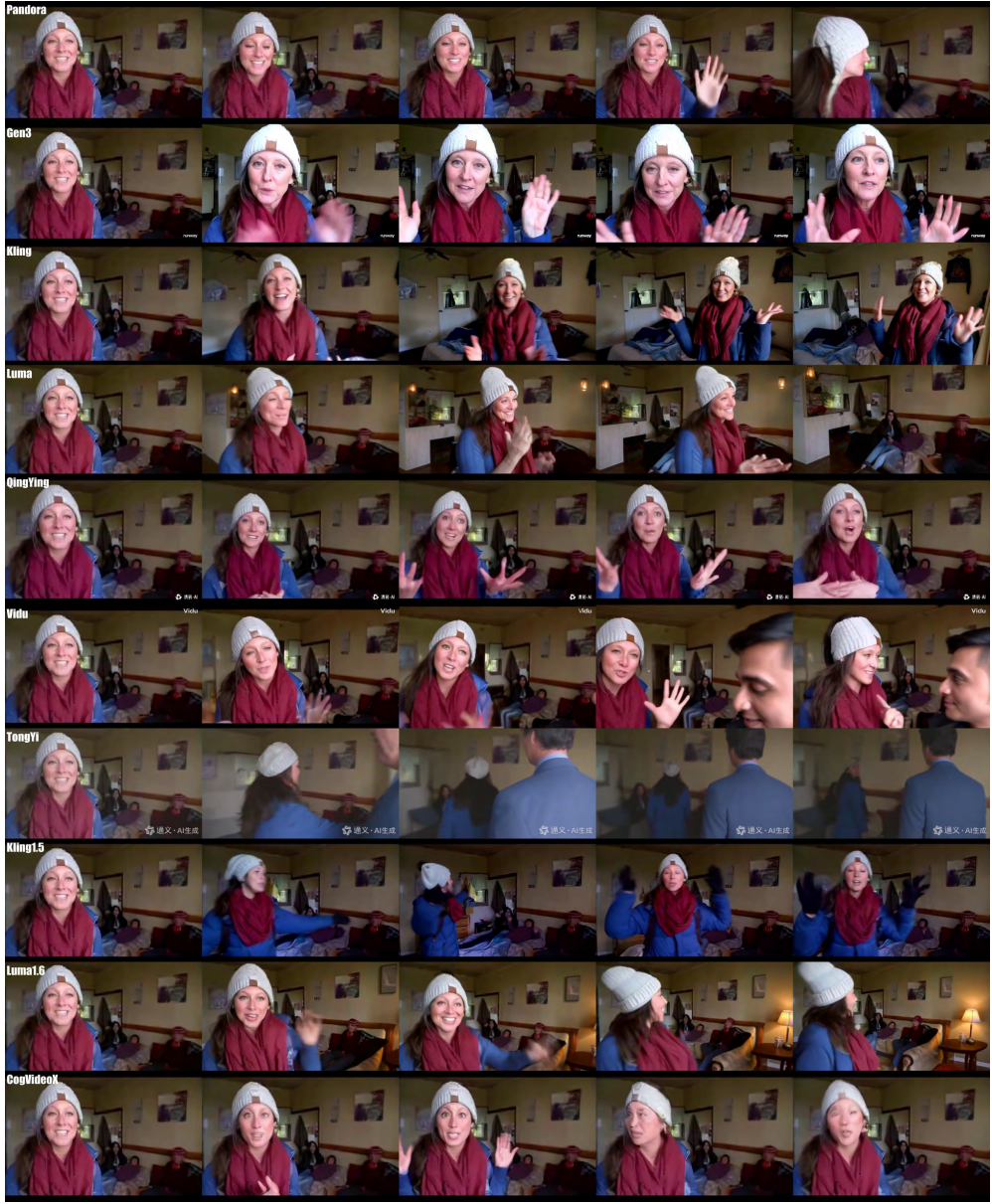

	\centering
	\small
        \begin{overpic}[width=0.95\linewidth]{Figs/Sec2/00618.pdf}

	\end{overpic}
	\caption{\emph{Comparisons with a general world model (\emph{e.g.}, Pandora \cite{xiang2024pandora})}. Prompt: (I2V-618) "Initially, the woman is talking, then she waves her hands, and last, she turns her head to the man." The models can generate the component motions but struggle to execute all instructions accurately in the correct order, which may stem from the lack of densely temporal-aligned captions.}
 \label{Fig:world1}
\end{figure}

\begin{figure}[!ht]
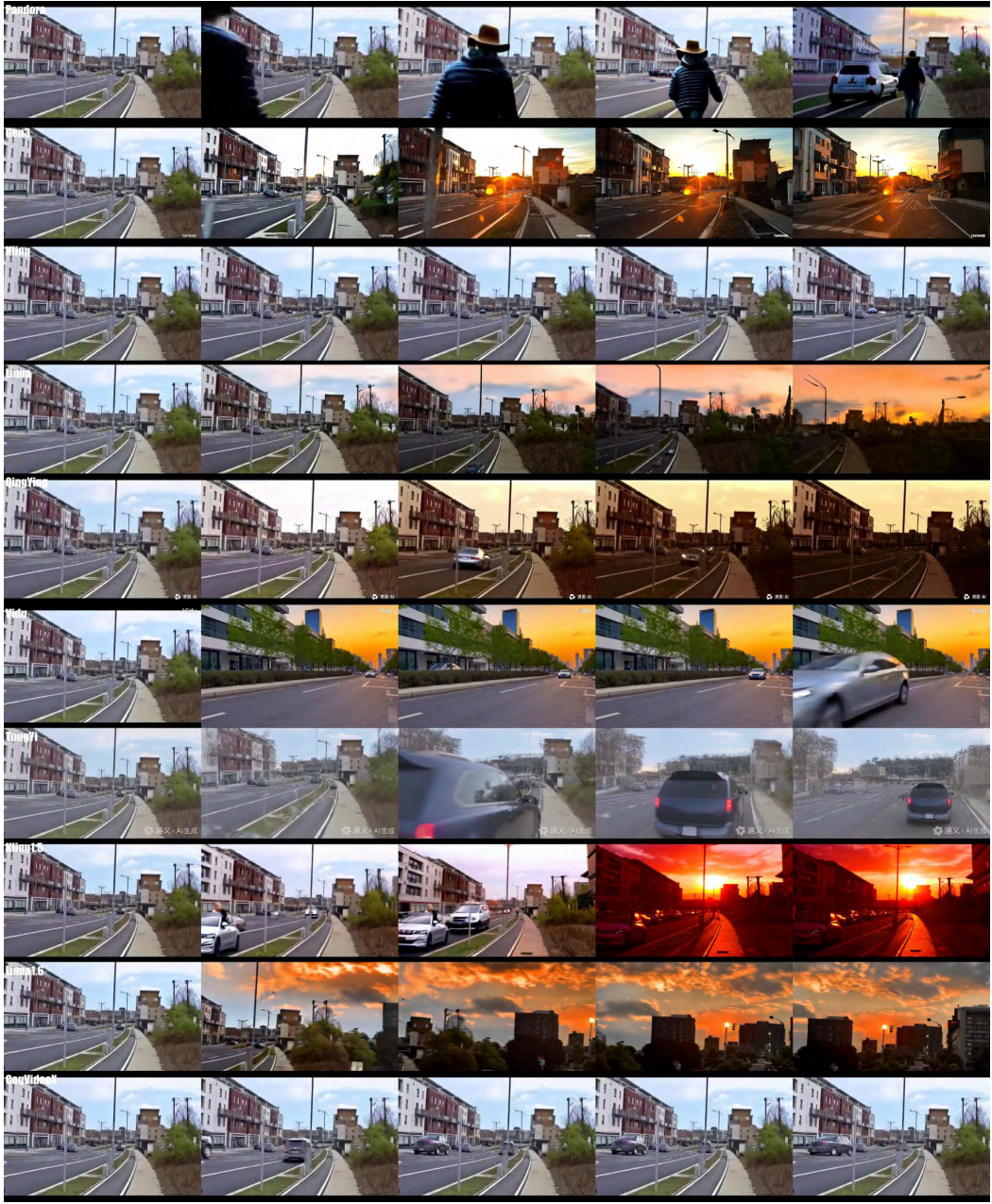

	\centering
	\small

        \begin{overpic}[width=1.\linewidth]{Figs/Sec2/00619.pdf}

	\end{overpic}
	\caption{\emph{Comparisons with a general world model (\emph{e.g.}, Pandora \cite{xiang2024pandora})}. Prompt: (I2V-619) "Initially, a person comes in from the back, then a car comes in from the back, after that, the weather changes to sunset." None of the models generate content as instructed, such as the emergence of a person from the back. Some models can generate some mentioned components, such as a car from the back and sunset.} 
 \label{Fig:world2}
\end{figure}

World models seek to simulate future states based on the current state and actions. Some approaches \cite{ao2024body, bruce2024genie, valevski2024diffusion, wang2024worlddreamer, xiang2024pandora} have explored different architectures to build the world model, and one crucial feature among them is the interactive generation. From this perspective, the mainstream full-sequence denoising methods may struggle to meet this definition. However, the ultimate goal is obtaining a continuous sequence that aligns with instructions (perhaps given in separated states). Therefore, we provide the initial state (an image) and combine instructions (or called actions) from multiple states into a sequentially ordered prompt and test it on existing closed-source models that apply full-sequence denoising, compared with Pandora \cite{xiang2024pandora}; please refer to Figure \ref{Fig:world1}, \ref{Fig:world2}.

Existing video generation models still have many limitations in encapsulating the physical and dynamic properties of the world, anticipating future states, reasoning about outcomes, and improving decision-making. Improving the fine-grained video understanding with the reason capabilities and interactive long video generation could benefit the target.

\clearpage

\subsection{Autonomous Driving}

\begin{figure}[!ht]
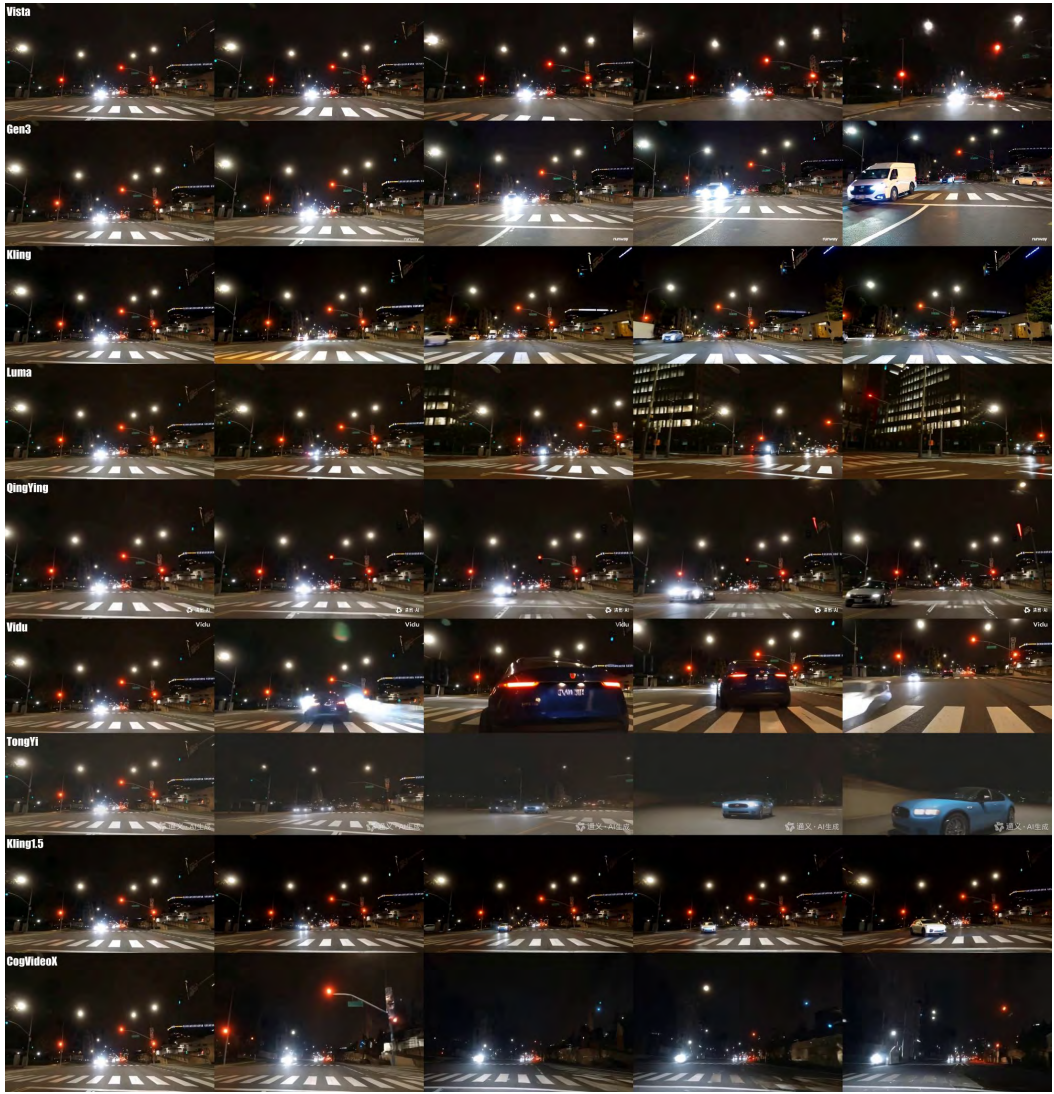

	\centering
	\small
        \begin{overpic}[width=1.\linewidth]{Figs/Sec2/00613.pdf}

	\end{overpic}
	\caption{\emph{Comparisons with a driving world model (\emph{e.g.}, Vista \cite{gao2024vista})}. Prompt: (I2V-613) "The ego-car turns \emph{left}." When the instruction contains a turn, models almost fail to generate the controllable motions and the corresponding environment.}
  \label{Fig:auto_drive_1}
\end{figure}

Regarding autonomous driving, high-quality data for training large end-to-end models is extremely costly. The simulator is an alternative option to collect data, but it usually requires crafted designs for controllable conditions, \eg, a complex environment. It also involves overcoming the sim-to-real gap. Thus, some methods \cite{wen2024panacea, yang2024drivearena, zhao2024drivedreamer} attempt to take video generation models to efficiently create various and more realistic autonomous driving data for training and planning. For this field, control generation mainly lies in two aspects: the embodied car and the external environment (\eg, the weather, the road conditions, other vehicles, and pedestrians). We test the controlling of the embodied car from an ego-car perspective and compare them with a specific method Vista \cite{gao2024vista}, please refer to Figure \ref{Fig:auto_drive_1}. Other aspects, \eg, the weather, etc., are shown in the application (Sec. \ref{Sec:application}).

\clearpage

\subsection{Camera Control}

\label{Sec:camera_control}
\begin{figure}[!ht]
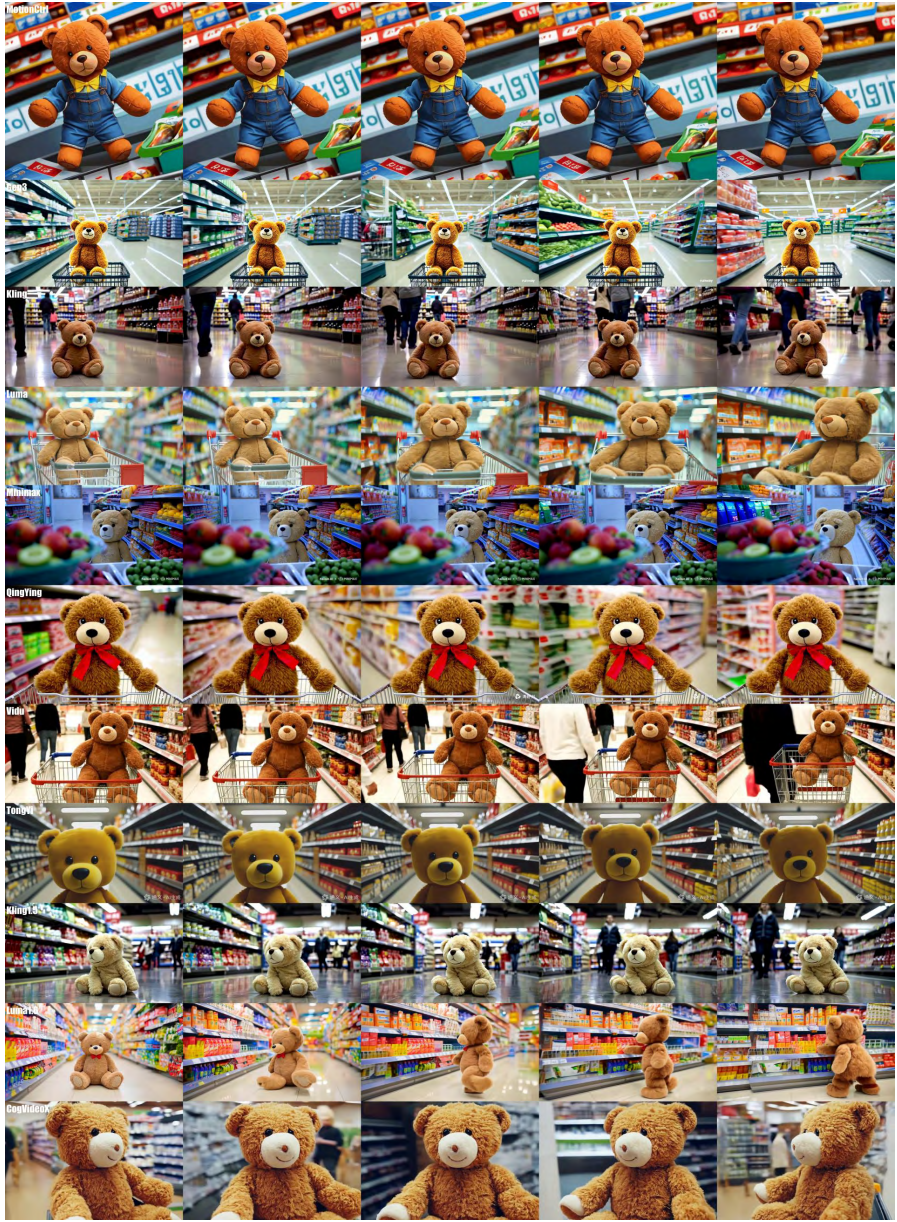

	\centering
	\small
        \begin{overpic}[width=0.85\linewidth]{Figs/Sec2/00657.pdf}
	\end{overpic}
	\caption{\emph{Comparisons with the camera-controllable video generation (\emph{e.g.}, MotionCtrl~\cite{wang2024motionctrl}) based on AnimateDiff~\cite{guo2023animatediff}}. Prompt: (T2V-657) "A teddy bear at the supermarket. The camera is moving \emph{anti-clockwise}." The camera moves to a certain extent according to instructions, especially Luma, Minimax, and CogVideoX.}
 \label{Fig:camera_motion1}
\end{figure}

\begin{figure}[!ht]
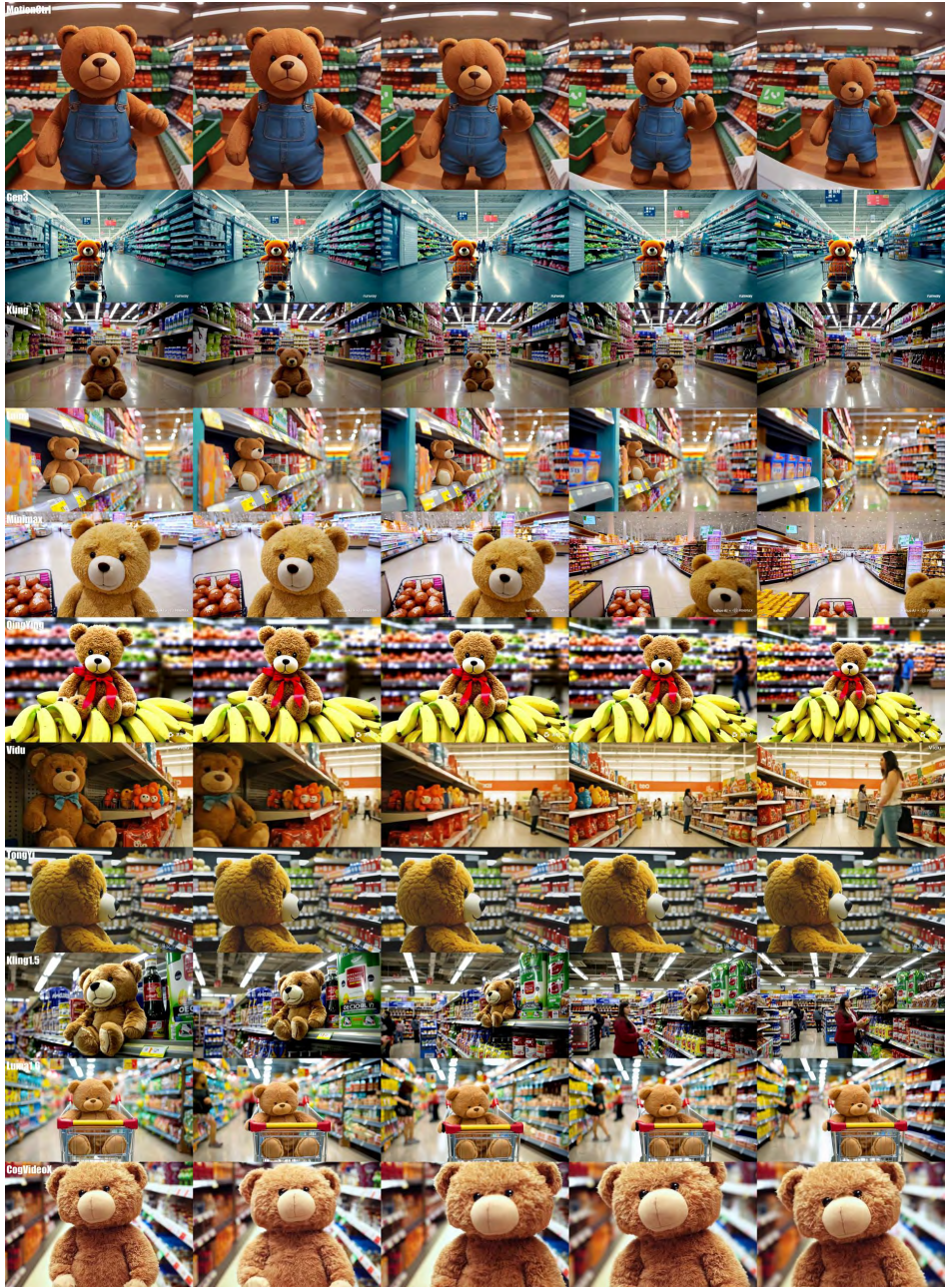

 \vspace{-0.7cm}
	\centering
	\small
        \begin{overpic}[width=0.9\linewidth]{Figs/Sec2/00658.pdf}
	\end{overpic}
  \vspace{-0.3cm}
	\caption{\emph{Comparisons with the camera-controllable video generation (\emph{e.g.}, MotionCtrl~\cite{wang2024motionctrl}) based on AnimateDiff~\cite{guo2023animatediff}}. Prompt: (T2V-658) "A teddy bear at the supermarket. The camera is \emph{zooming out}." All made corresponding camera motions according to the instructions.}
 \label{Fig:camera_motion2}
 \vspace{-0.3cm}
\end{figure}

Camera control plays a crucial role in the film industry. Actually, camera movement highlights the inherent 3D nature. As the camera moves, both the foreground subject and the background should change according to the camera motion while maintaining consistency. For camera control in I2V, the provided 2D image is merely a projection from an angle. When the camera moves, the model should generate unseen or novel views, intensifying the challenge with the given condition. Here, we test the text-controlled camera motion generation and compare it with MotionCtrl \cite{wang2024motionctrl}, please refer to Figure \ref{Fig:camera_motion1}, \ref{Fig:camera_motion2}. Luma~\cite{luma2024dm}, MiniMax~\cite{minimax2024hailuo}, and CogVideoX~\cite{yang2024cogvideox} successfully control the proper "anti-clockwise" camera movement. For the commonly used "zooming out" guidance, more models can control the corresponding camera movements.

\clearpage

In summary, existing SORA-like models still lack densely spatio-temporal fine-grained text annotations and insufficient descriptive focus on domain-specific information, such as facial expressions, spoken language, fine-grained gesture actions, precise camera and subject motion, simultaneously, and professional descriptions in many scenes, achieving precise video generation control in specialized domains through I2V combined with input text remains challenging. Thankfully, general-purpose models still enhance the fundamental capabilities of overall modeling in aspects like generalization, consistency, composition, and diversity. They excel in world-simulated modeling (combining humans, objects, environments, etc.). Unlike models specifically tailored for human-centric video generation, they generalize effectively across various scenarios and better understand the interactive information between humans, objects, and environments. For instance, in contrast to current pose-conditioned video generation methods that rely on explicit keypoint or motion guidance, whose errors would significantly impact the animation quality (\emph{e.g.}, fine-grained gestures and interaction under heavy occlusions), general-purpose models exhibit more robustly in expressing human motion modeling.

\clearpage

\section{Objective Video Generation Capability}
\label{sec:3}
\begin{figure}[h]
	\centering
	\small
        \begin{overpic}[width=1\linewidth]{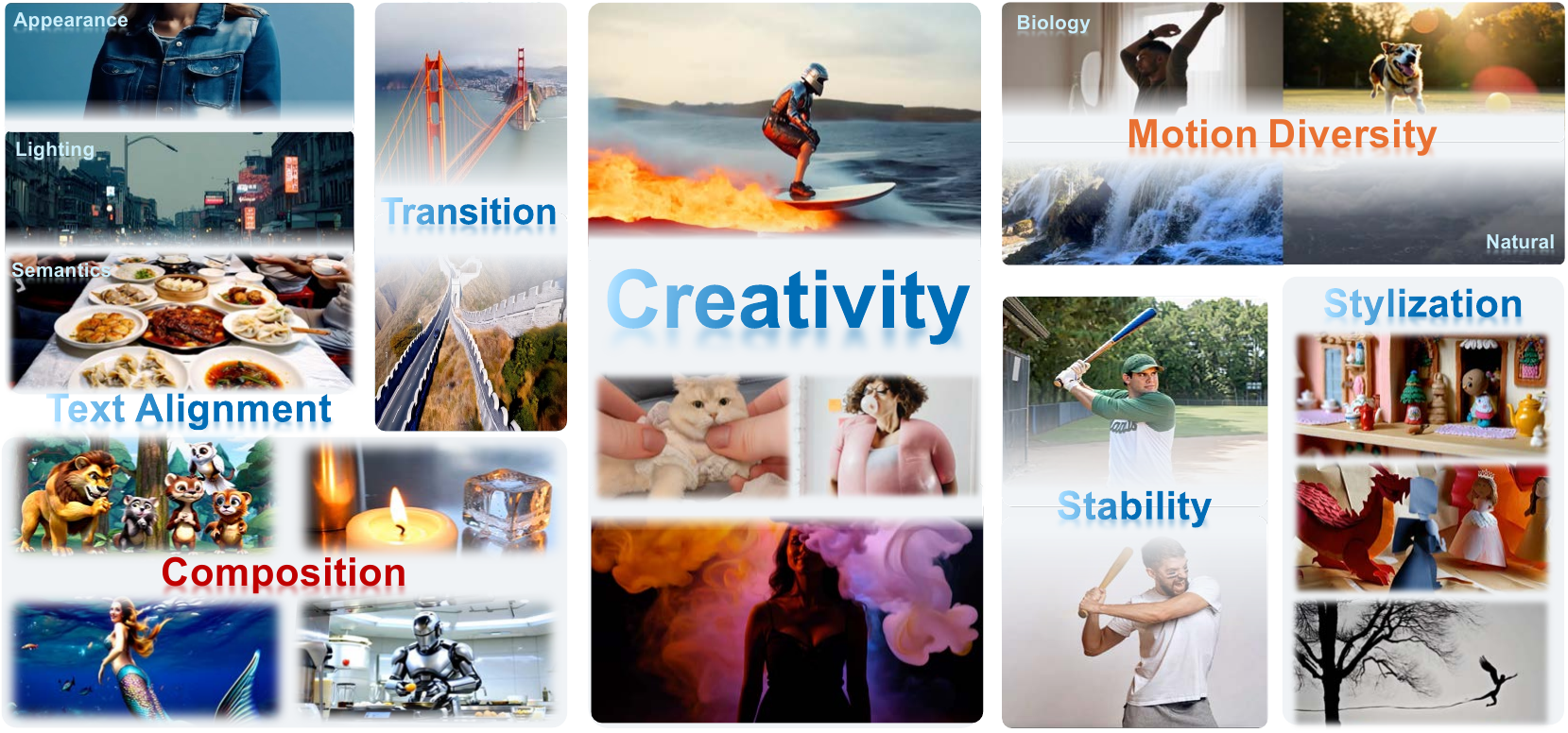}

	\end{overpic}
        \caption{\textbf{Overview of Section~\ref{sec:3}.} We evaluate various objective abilities in video generation, such as text alignment, transition, composition ability, creativity, motion diversity, stability, and stylization.}
        \label{Fig:ability_teaser}
\end{figure}

In this section, we explore various key objective abilities (Figure \ref{Fig:ability_teaser}) for video generation models, providing intuitive examples for observation.

\subsection{Text-alignment}
So far, text-to-video (T2V) is the primary task for training video generation models to semantically control the video generation via user-friendly inputs. Other video generation tasks \eg, image-to-video (I2V) and video-to-video (V2V), could be further or simultaneously trained based on T2V. Thus, aligning the generated video with the input text is the fundamental ability. 

Regarding alignment, it involves many aspects, which can be divided into explicit alignment, \eg visual appearance, as well as alignment implied in the text, \eg, causal phenomenon, etc. In our designed prompts, we maintain diversity and represent a wide range of scenarios, focusing on different aspects, such as nature and environment, people and emotions, technology and imagination, urban and architecture, art and culture, kinds of animals, various food, and diverse objects, the combination, and their corresponding motions and interactions.

\begin{figure}[!ht]
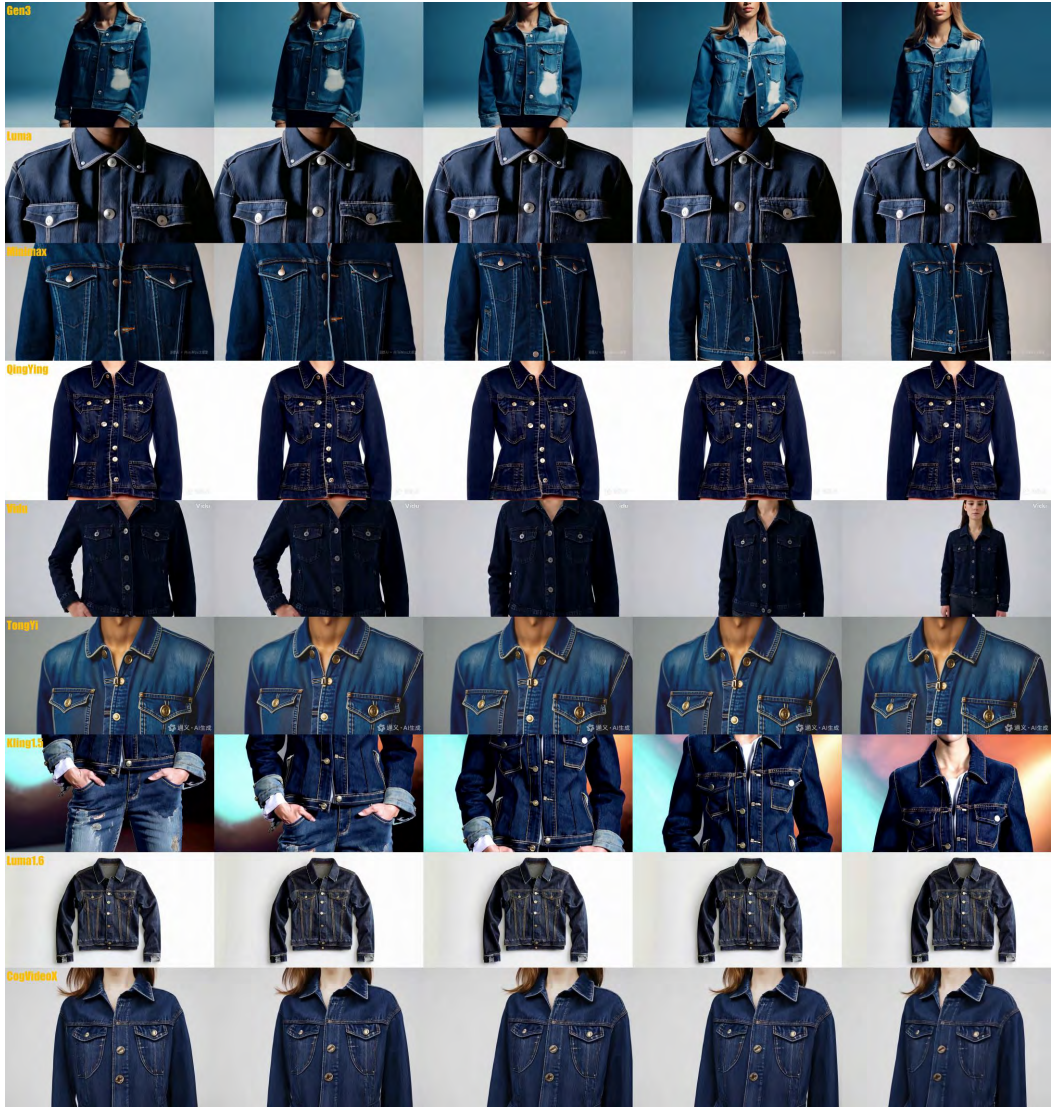

	\centering
	\small
        \begin{overpic}[width=1.\linewidth]{Figs/Sec3/00126.pdf}

	\end{overpic}
	\caption{\emph{Text alignment, appearance.} Prompt: (T2V-126) "Static camera, a model wearing a dark blue denim jacket. The jacket should have a classic collar design, metal buttons, and two chest pockets. The hem and cuffs of the jacket should have a worn-out effect, giving it a fashionable distressed look." Gen-3~\cite{runway2024gen3} generates a more suitable appearance with a better aesthetic. }
 \label{Fig:text-align 1}
\end{figure}

\begin{figure}[!ht]
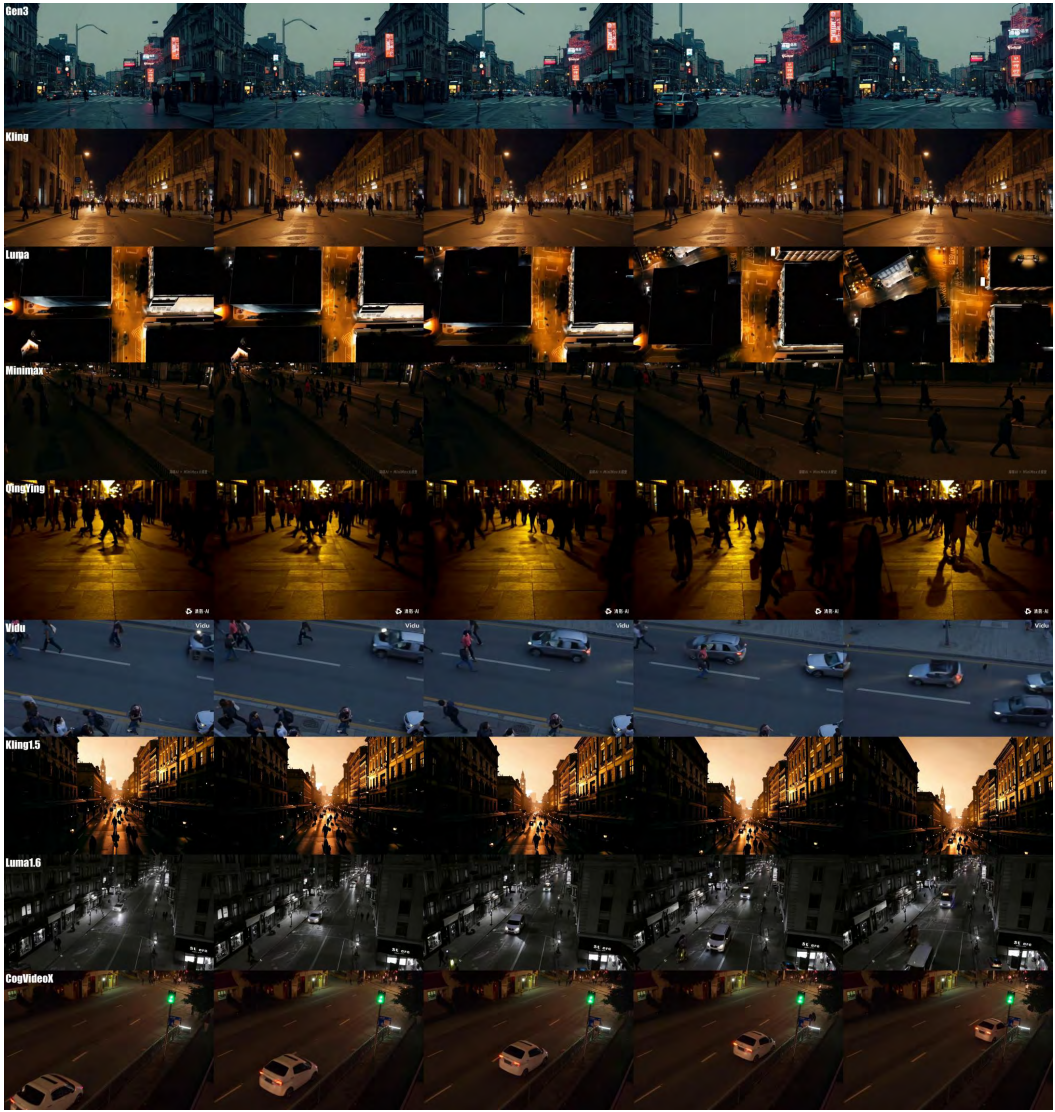

	\centering
	\small
        \begin{overpic}[width=1.\linewidth]{Figs/Sec3/00124.pdf}

	\end{overpic}
        \caption{\emph{Text alignment, viewpoint and lighting.} Prompt: (T2V-124) "Overlooking the street, pedestrians walking on the road, dim lighting." Luma~\cite{luma2024dm} shows better viewpoint generation; all models can generate the proper lighting. They vary greatly in visual content and motion.}
 \label{Fig:text-align 2}
\end{figure}

\begin{figure}[!ht]
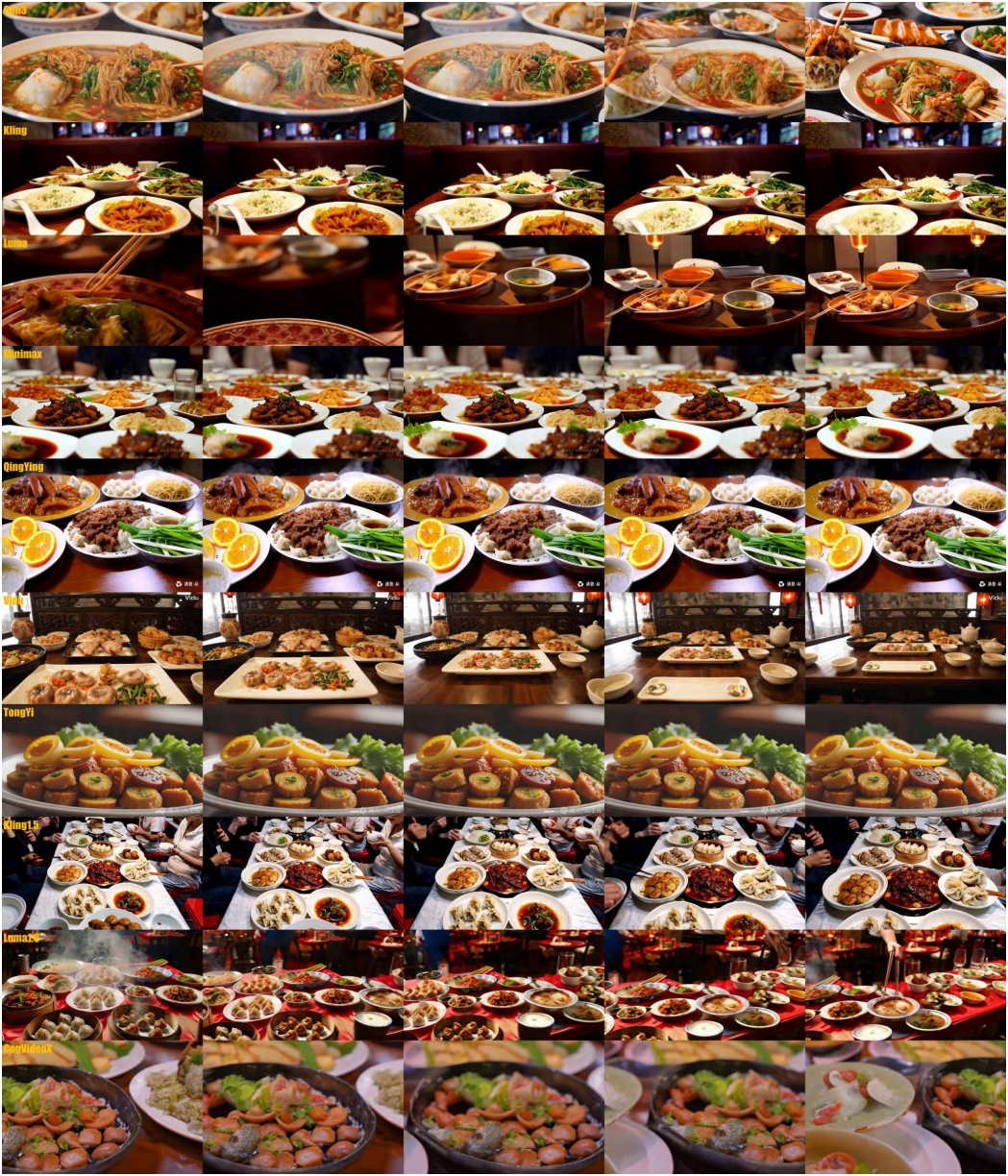

	\centering
	\small
        \begin{overpic}[width=1.\linewidth]{Figs/Sec3/00137.pdf}

	\end{overpic}
        \caption{\emph{Text alignment, semantics.} Prompt: (T2V-137) "Close-up shot of a table in the restaurant filled with Chinese cuisine." Kling~\cite{kuaishou2024kling} and MiniMax~\cite{minimax2024hailuo} can generate better semantic visual contents with a better understanding of Chinese cuisine.}
 \label{Fig:text-align 3}
\end{figure}

In this part, we select some representative intuitive and explicit aspects to demonstrate, including the appearance, environmental lighting, semantics, and camera motion, while the implicit aspects are discussed in later sections. Cases for illustrating the explicit alignment are shown below (Figure \ref{Fig:text-align 1}, \ref{Fig:text-align 2}, \ref{Fig:text-align 3}). More examples can be viewed on the website.
Figure \ref{Fig:text-align 1} explores the basic appearance generation ability, which reflects the text-aligned ability as the text-to-image generation task due to no additional motion description. However, existing T2V models still struggle to achieve the same level of appearance accuracy as state-of-the-art image generation.

\clearpage

\subsection{Composition}
\begin{figure}[!ht]
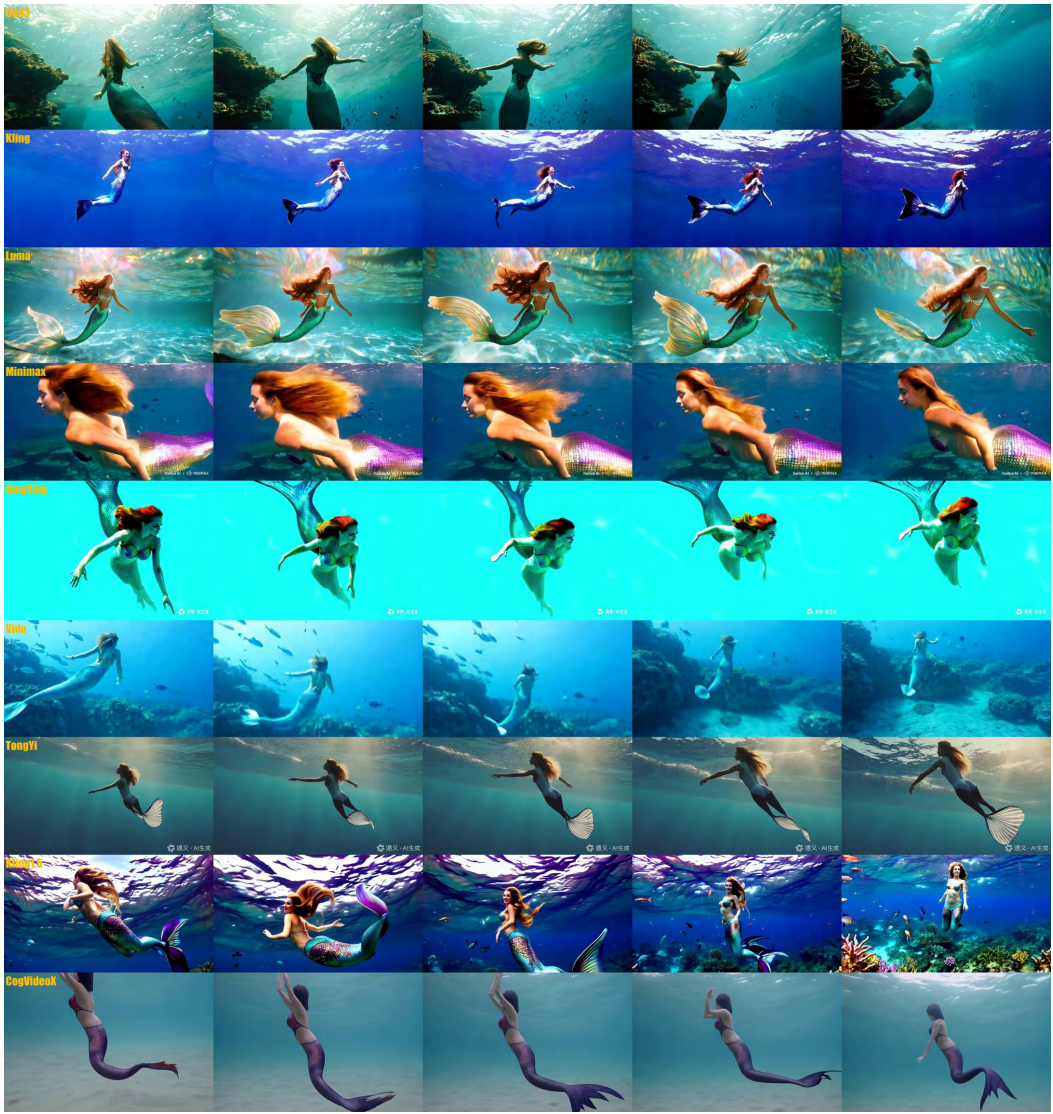

	\centering
	\small
        \begin{overpic}[width=1.\linewidth]{Figs/Sec3/00721.pdf}

	\end{overpic}
        \caption{\emph{Composition of different body parts and environment.} Prompt: (T2V-721) "A mermaid swims in the sea, with her upper body and fish tail in a beautiful and exquisite style." Generating with a fish tail instead of two legs is still hard for some models (\emph{e.g.}, Tongyi and Kling). Additionally, different models have significant differences in aesthetics and motion dynamism.}
 \label{Fig:composition_1}
\end{figure}

Building on text alignment, the ability to simultaneously assemble multiple classes, components, instances, and motions is crucial for generating diverse and complex videos. To achieve this, the model should discern distinct semantic concepts and handle the combination patterns between them. Here, we showcase that assembles multiple parts into an instance (Figure \ref{Fig:composition_1}), and more results including assembling multiple instances and motions into a scene can be found on the website.

\clearpage

\subsection{Transition}

\begin{figure}[!ht]
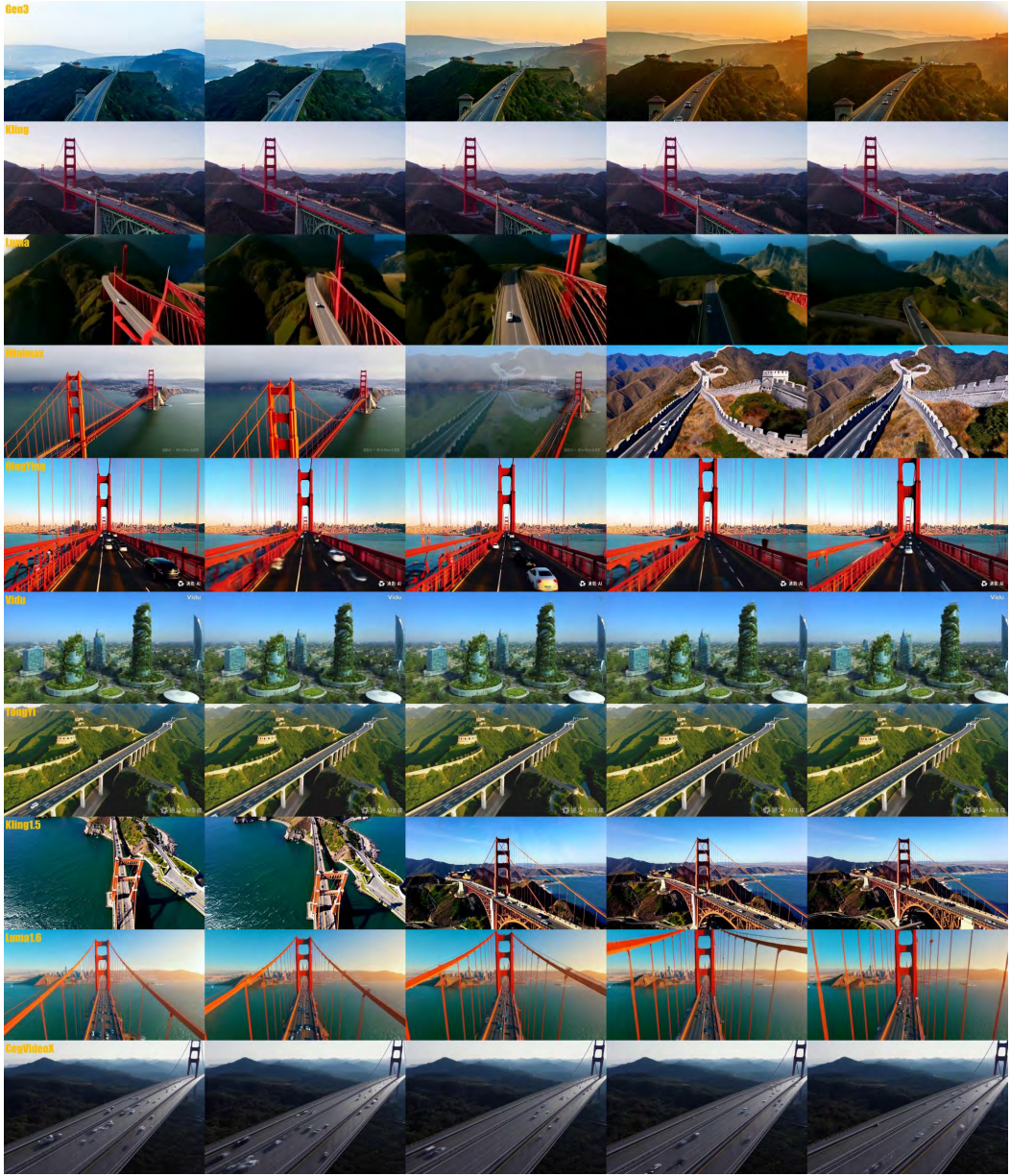

	\centering
	\small
        \begin{overpic}[width=1.\linewidth]{Figs/Sec3/00324.pdf}

	\end{overpic}
        \caption{\emph{Transition.} Prompt: (T2V-324) "Showcase the world's most iconic bridges and highways, from the Golden Gate Bridge to the Great Wall of China. The camera follows vehicles as they traverse these structures, highlighting their architectural brilliance and the landscapes they connect. Use a mix of drone shots, on-the-road footage, and time-lapses to capture the movement and functionality of these infrastructures." Only MiniMax generates the proper scenarios with transitions.}
         \label{Fig:transition_1}
\end{figure}

The transition is a commonly used narrative technique in the film industry and content creation. The slight difference from the combination is that it involves merging multiple whole scenes while adhering to the ordered sequence specified by the instructions. It also requires grasping various semantic concepts. Figure \ref{Fig:transition_1} demonstrates a case of this ability, which performs a transition between two landmarks.

\clearpage

\subsection{Creativity}
\begin{figure}[!ht]
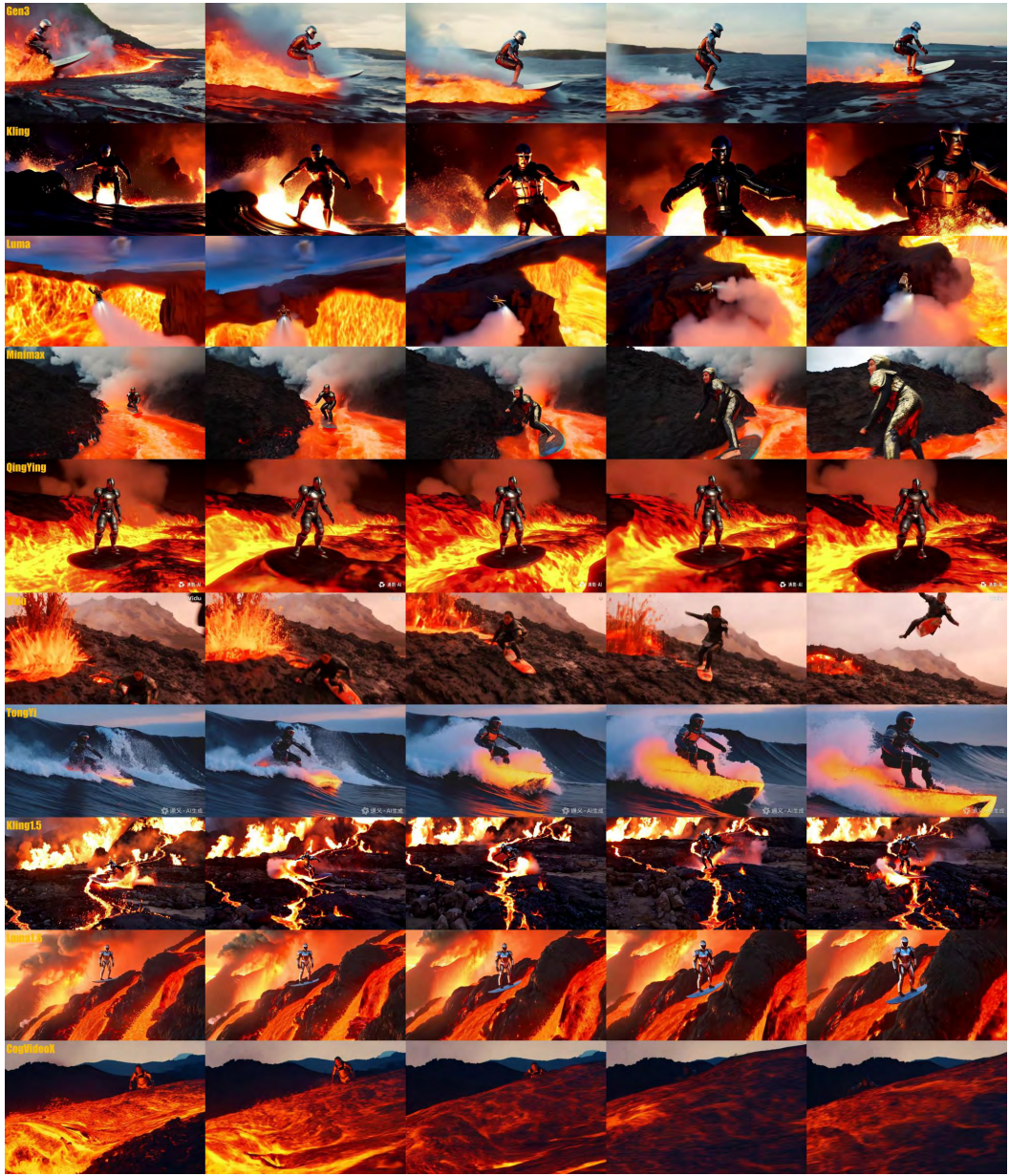

	\centering
	\small
        \begin{overpic}[width=1.0\linewidth]{Figs/Sec3/00276.pdf}

	\end{overpic}
        \caption{\emph{Creativity, unrelated composition creation.}  Prompt: (T2V-276) "A surfer in a suit of armor rides a lava flow from an active volcano, the camera follows closely, capturing the heat and intensity of the moment." All models can combine the concepts with various motions and aesthetics.}
        \label{Fig:Creativity_1}
\end{figure}

Creativity usually involves the combination of unrelated concepts and producing novel, imaginative, and unique content, such as surfing on the fire. This ability is built upon the foundation of alignment and composition. Whether the combination of unrelated concepts is visually plausible, and whether the motion pattern is reasonable when exhibited in abnormal concept combinations are key criteria for evaluating this capability (Figure \ref{Fig:Creativity_1}). In addition, the subjective nature of creativity should be evaluated via originality, aesthetic innovation, contextual awareness, adaptive storytelling, expressive and emotional range, etc. We show a case below.

\begin{figure}[!ht]
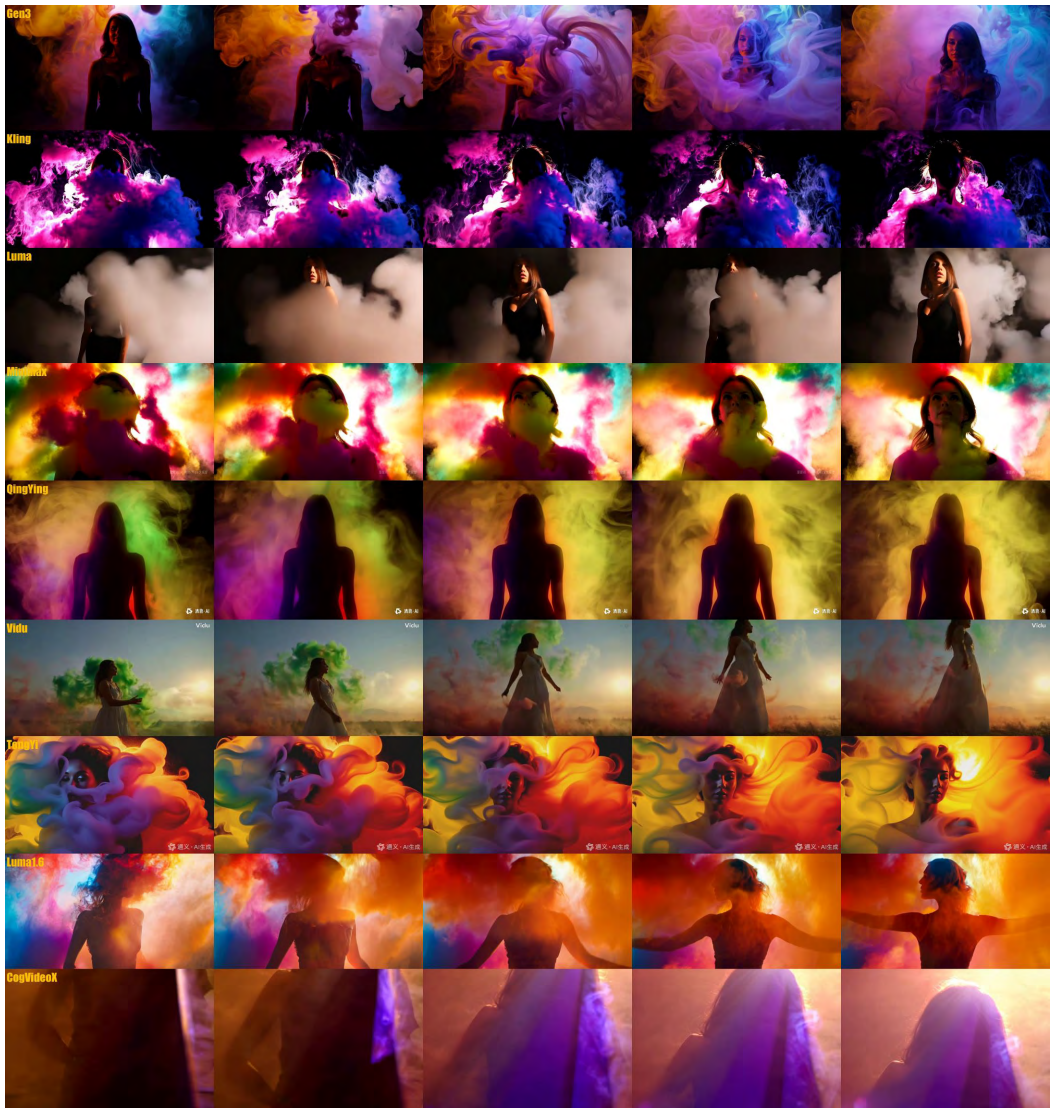

	\centering
	\small
        \begin{overpic}[width=1.\linewidth]{Figs/Sec3/00435.pdf}

	\end{overpic}
        \caption{\emph{Creativity, special effects with smoke simulation.} Prompt: (T2V-435) "In this medium shot video, a woman is enveloped in a vibrant haze of colorful smoke, bathed in a warm, ethereal light. As the camera slowly tilts upwards, her silhouette, partially obscured by the swirling smoke, gradually comes into focus." All models can generate smoke effects with various appearances, motions, and aesthetics.}
        \label{Fig:Creativity_2}
\end{figure}

\begin{figure}[!ht]
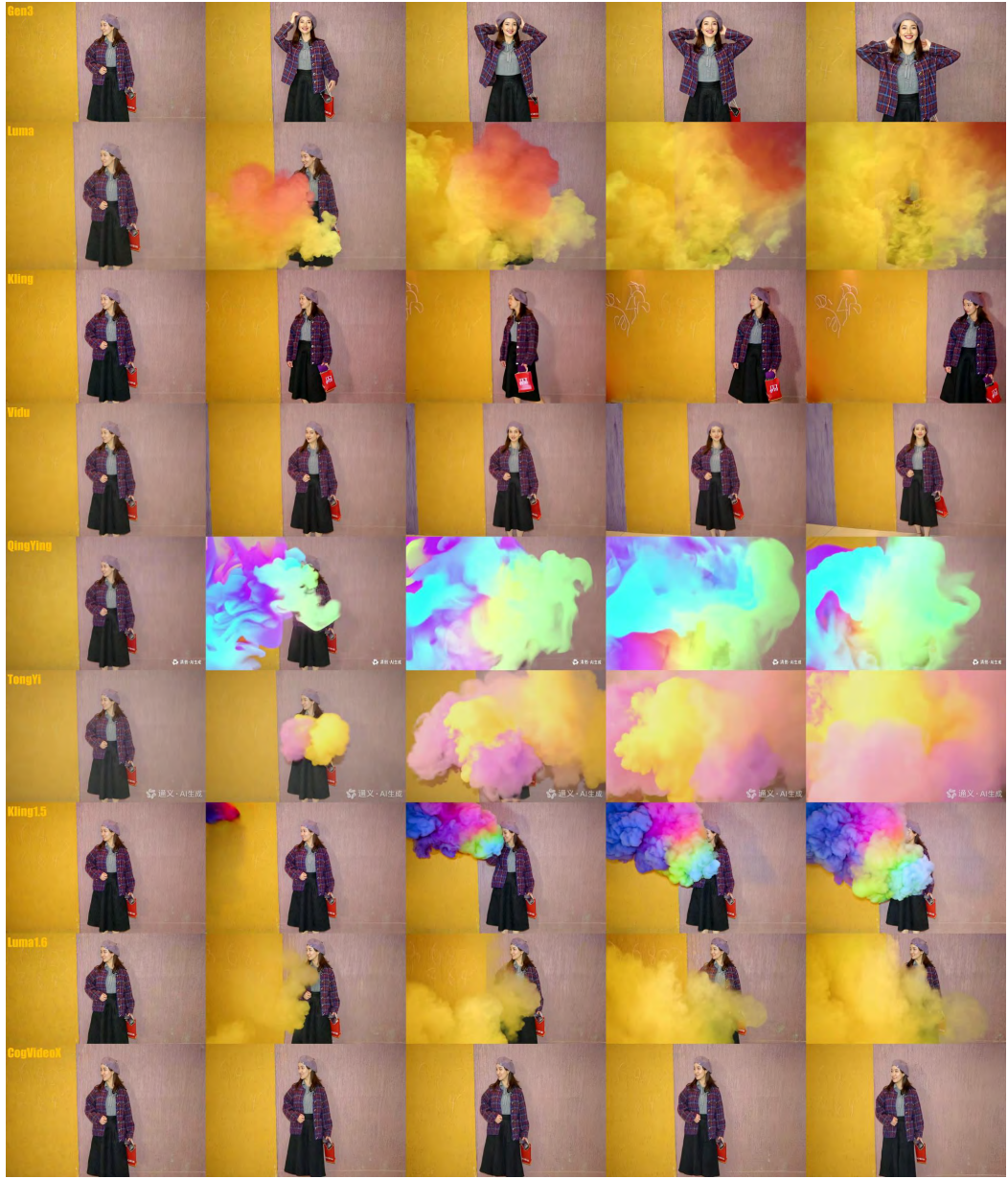

	\centering
	\small
        \begin{overpic}[width=1.\linewidth]{Figs/Sec3/00720.pdf}

	\end{overpic}
	\caption{\emph{Creativity, special effects with smoke simulation.} Prompt: (I2V-720) "In this medium shot video, a woman is enveloped in a vibrant haze of colorful smoke, in a warm, ethereal light." When the task transitions from T2V to I2V, with the same smoke effect added, some models fail to generate the effect due to the prompts' injection and fusion ways.}
    \label{Fig:Creativity_3}
\end{figure}

\begin{figure}[!ht]
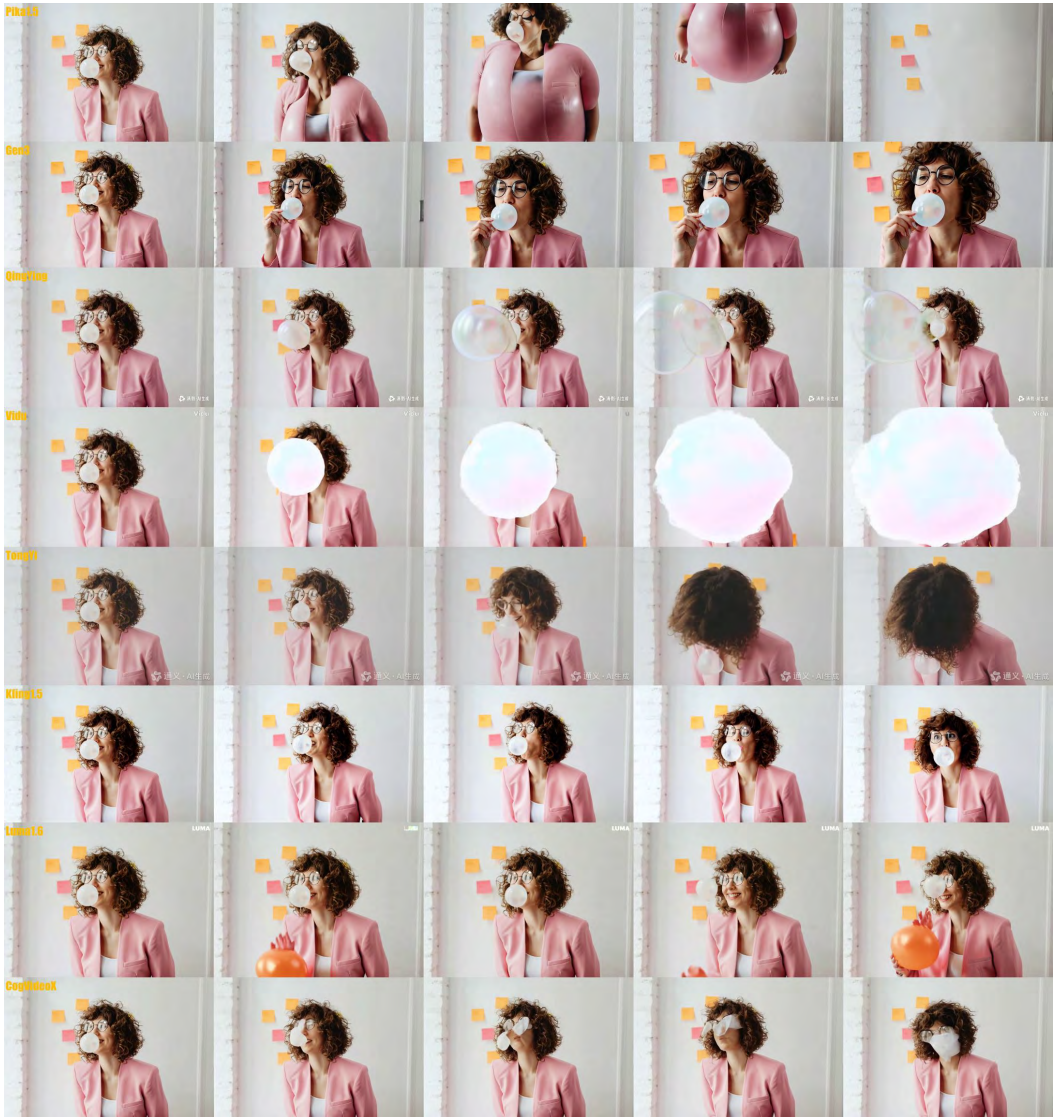

	\centering
	\small
        \begin{overpic}[width=1.\linewidth]{Figs/Sec3/00724.pdf}

	\end{overpic}
        \caption{\emph{Creativity, special effects with inflation.} Prompt: (I2V-724) "Inflate it." General-purposed models fail to generate the special effects, while Vidu tries to generate such effects in another way.}
        \label{Fig:Creativity_4}
\end{figure}

\begin{figure}[!ht]
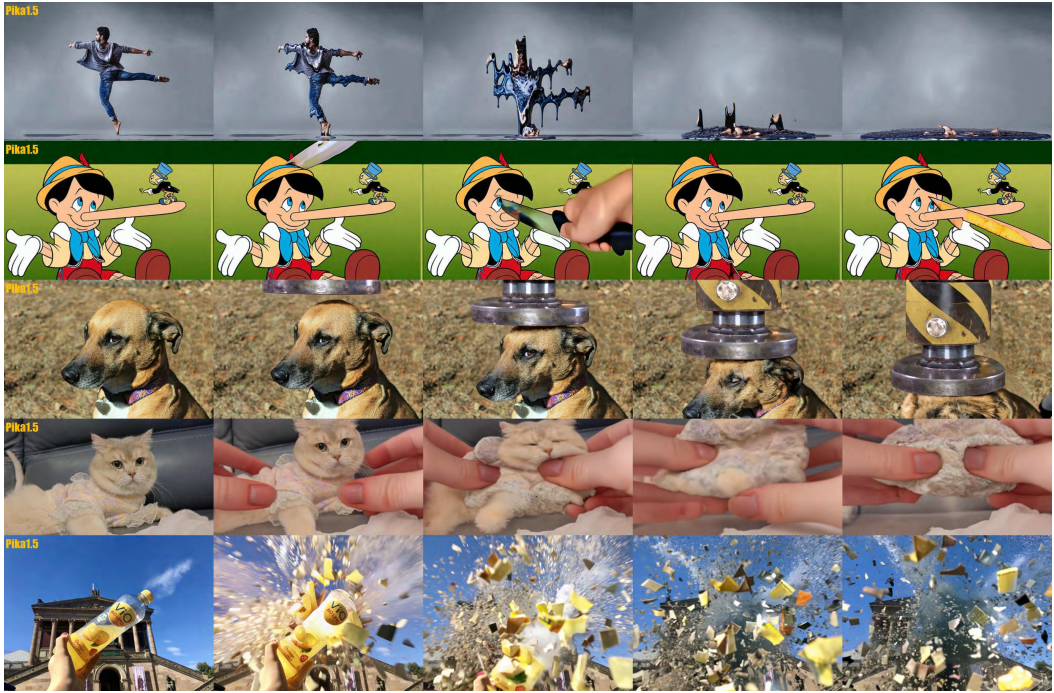

	\centering
	\small
        \begin{overpic}[width=1.\linewidth]{Figs/Sec3/pika.pdf}

	\end{overpic}
        \caption{\emph{Creativity, other five special effects introduced in Pika 1.5~\cite{pika2024pika}.} Prompt: (I2V-725:729) 1) "Melt it."; 2) "Cake-ify it"; 3) "Crush it"; 4) "Squish it"; 5) "Explode it." With five input images, we evaluate them on human, cartoon, animal, and scene contents. "Cake-ify" is the most challenging task, and other effects have a high success ratio, even with low-resolution subjects or subjects not in the center of the image.}
        \label{Fig:Creativity_5}
\end{figure}

\textbf{Special Effects.} Moreover, special effects generation is also an aspect of creativity, which poses significant challenges to the model’s conceptual combination, generated visual quality, and physical plausibility. Cases for this aspect are evaluated in Figure \ref{Fig:Creativity_2}, \ref{Fig:Creativity_3}, \ref{Fig:Creativity_4}, \ref{Fig:Creativity_5}.

\clearpage
Recently, Pika 1.5~\cite{pika2024pika} introduced six types of video effects (\emph{i.e.}, inflate, melt, explode, squish, crush, cake-ify) based on image-to-video (I2V) models, enabling precise specific effect generation. In Figure~\ref{Fig:Creativity_4}, we compare the performance of one of these effects (\emph{e.g.}, inflation) across all SORA-like models, and the general-purposed models that are not specifically trained for these effects failed to generate them. In Figure~\ref{Fig:Creativity_5}, we test Pika 1.5's performance on the six proposed effects with our designed images as inputs, and regardless of the content of the input image (maintaining a single subject in an image), including variations in subject size and position, Pika 1.5 can generate the effects quite well. Therefore, targeted training for specific effects can greatly improve accuracy and controllability, which is particularly useful for niche scenarios, while ensuring the generated results align with visual and physical effects.

\clearpage

\subsection{Stylization}
\begin{figure}[!ht]
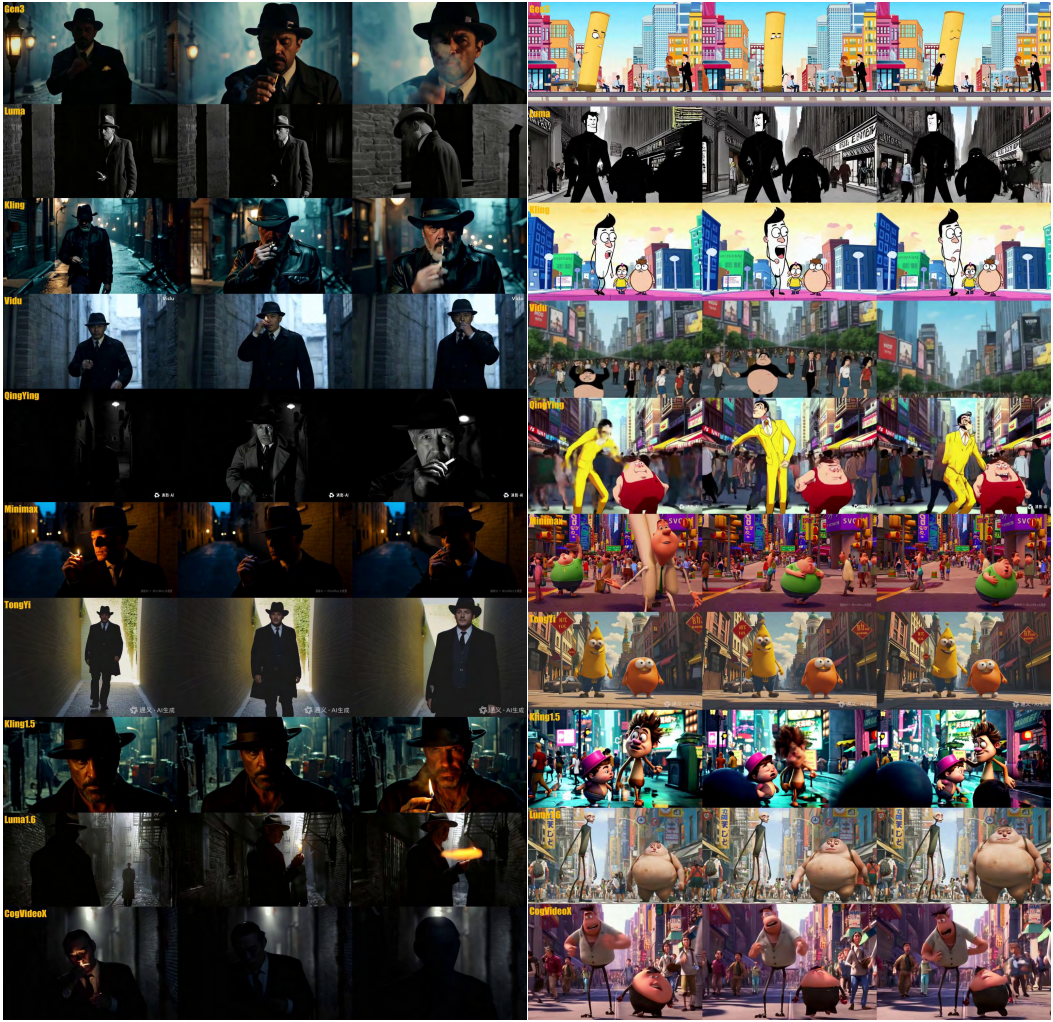

	\centering
	\small
        \begin{overpic}[width=1.\linewidth]{Figs/Sec3/00265_00380.pdf}

	\end{overpic}
	\caption{\emph{Stylization}. Prompt: (T2V) "[Left-265] A noir-style scene with a detective walking down a dimly lit alley, the camera captures the play of shadows across his face as he lights a cigarette, revealing a stern expression. [Right-380] In a cozy, warmly-lit cottage during winter, a grandmotherly figure knits a blanket by the fireplace while a small child curls up beside her with a book. The camera captures the intimate, peaceful atmosphere, focusing on the gentle expressions of both characters. The animation includes subtle, soft movements, like the flickering of the fire and the slight rustle of the blanket, evoking a sense of comfort and warmth."}
    \label{Fig:Stylization_t2v1}
\end{figure}

\begin{figure}[!ht]
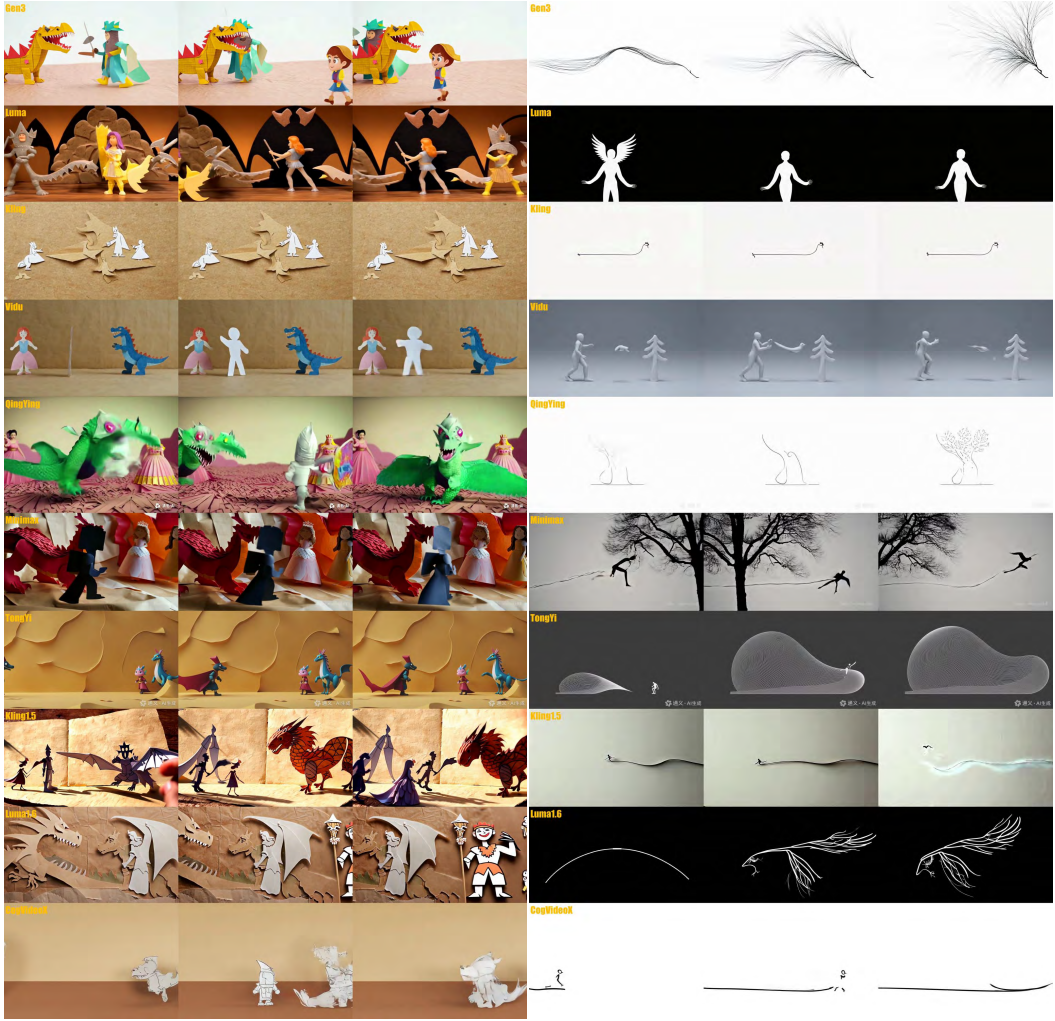

	\centering
	\small
        \begin{overpic}[width=1.\linewidth]{Figs/Sec3/00386_00387.pdf}

	\end{overpic}
	\caption{\emph{Stylization}. Prompt: (T2V) "[Left-386] A whimsical scene where paper cutout characters, including a knight, a dragon, and a princess, move across a textured paper background. The camera follows the knight as he bravely faces the dragon, using simple, jerky motions that mimic the feel of paper being moved by hand. The animation features layered textures, giving depth to the flat, paper world, with playful, childlike movements. [Right-387] A minimalist animation featuring a single, continuous line that morphs into various shapes to tell a story. The line starts as a simple curve, then transforms into a running figure, a tree, and finally, a bird taking flight. The camera remains centered, allowing the fluid, continuous transformation of the line to take center stage. The animation is clean, elegant, and focuses on the simplicity of the art form."}
    \label{Fig:Stylization_t2v2}
\end{figure}

\begin{figure}[!ht]
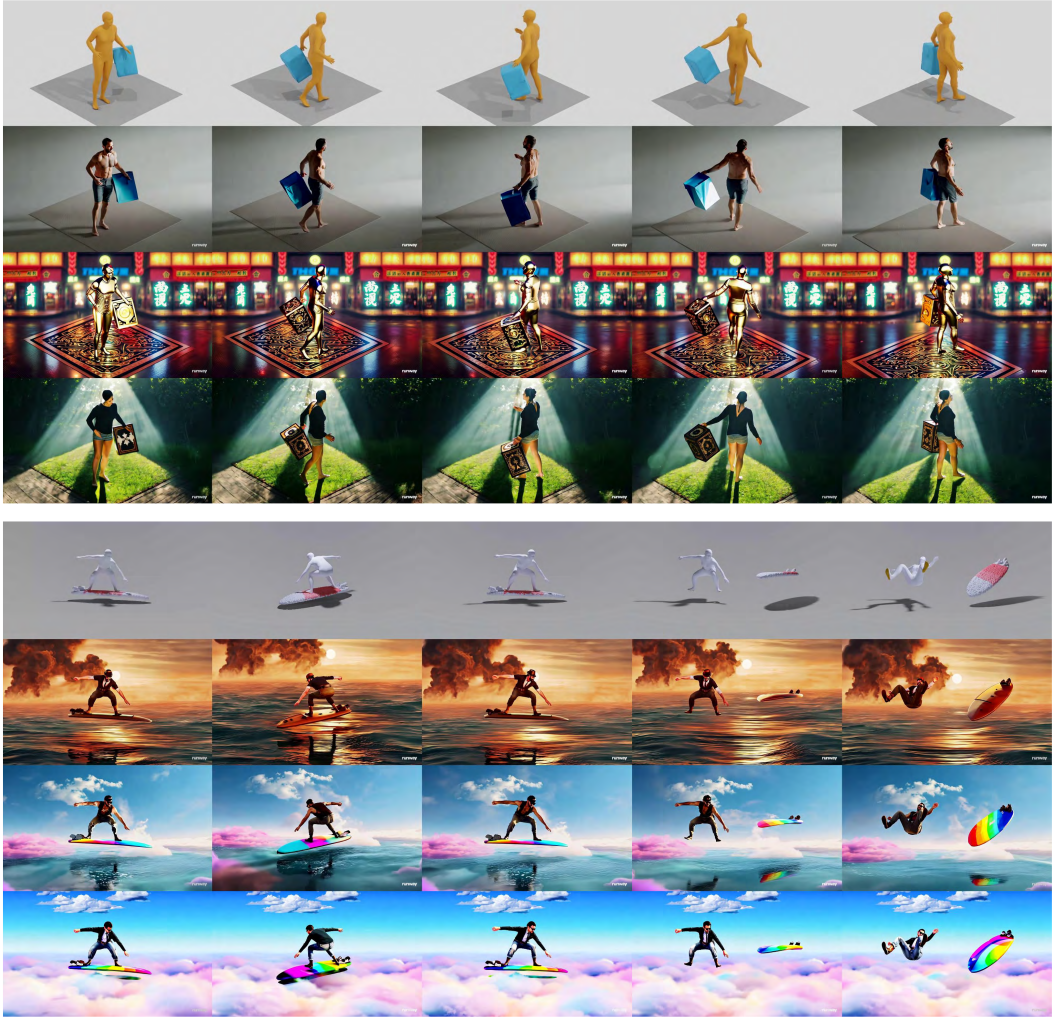

	\centering
	\small
        \begin{overpic}[width=1.\linewidth]{Figs/Sec3/00671_00687.pdf}

	\end{overpic}
	\caption{\emph{Stylization}. Prompt: (V2V-671:673, 687:689) 1) input video; 2) "a man walks in the room."; 3) "Ultraman walks in Chinatown holding a metal treasure box."; 4) "a woman with long black hair walks in a forest with a wool box."; 5) input video; 6) "a cow boy and surfacing board, below is a sea of fire. the man is screaming and jumping up."; 7) "a cow boy with black sunglasses and surfacing board, on a rainbow clouds, blue sky."; 8) "a cow boy with black sunglasses and rainbow surfacing board, on a rainbow clouds, blue sky." The input video demos are sourced from HOIDiff \cite{peng2023hoi} and LEMON \cite{yang2024lemon}.}
    \label{Fig:Stylization_v2v1}
\end{figure}

\begin{figure}[!ht]
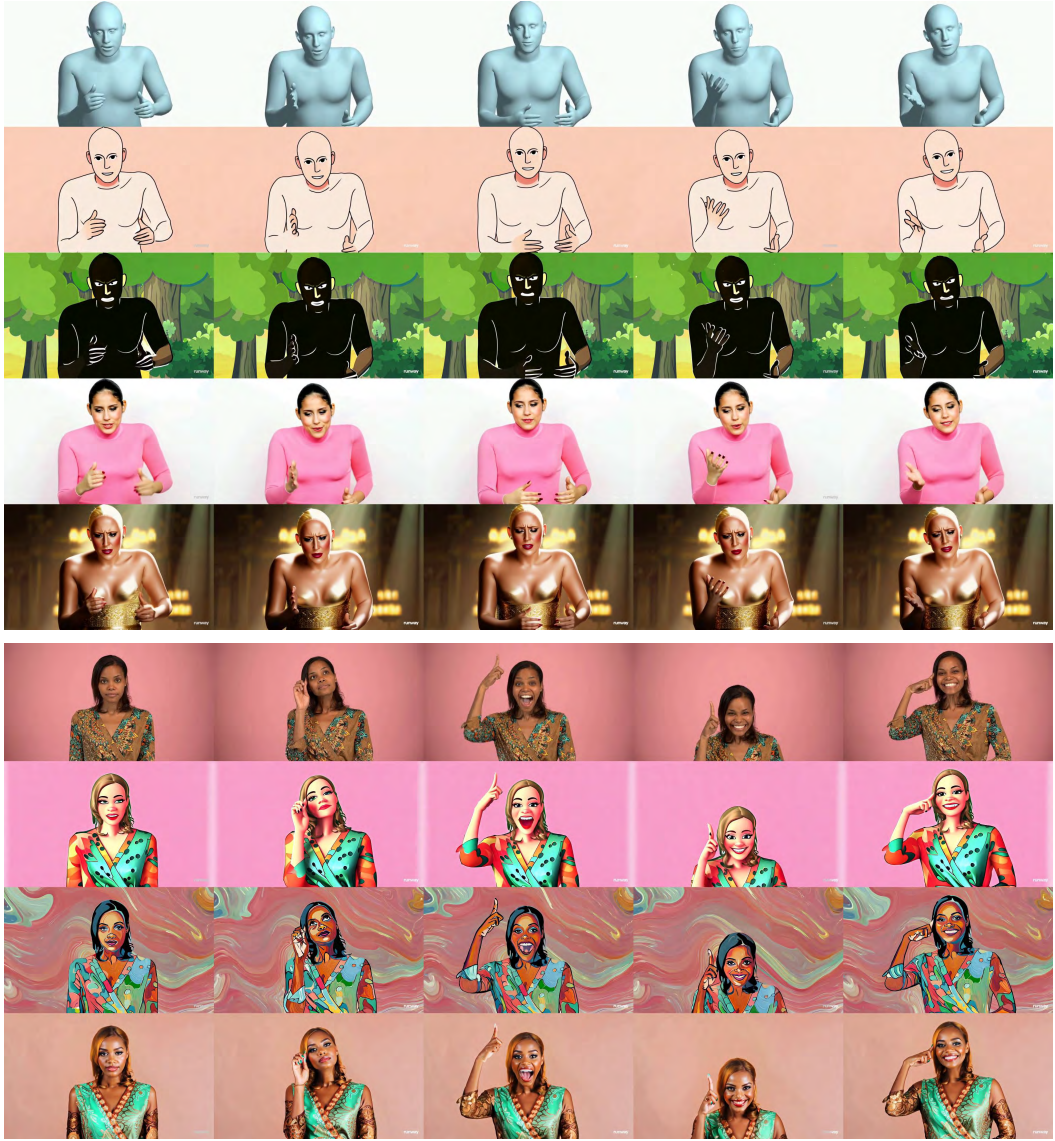

	\centering
	\small
        \begin{overpic}[width=1.\linewidth]{Figs/Sec3/00680_00703.pdf}

	\end{overpic}
	\caption{\emph{Stylization}. Prompt: (V2V-680:683, 703:705) 1) input video; 2) "A cute anime girl, speaking with a cute smile on her face."; 3) "A game boy, speaking with a angry expression, the background is in the green forest, animation style"; 4) "A cute Lolita girl wearing a pink hat, speaking with a smile on her face."; 5) "A princess with crown and golden hair, speaking with anxiety on her face."; 6) input video; 7) "cartoon style and turn the woman's face into princess-like style."; 8) "Van Gogh style and keep the human's appearance feature."; 9) "turn this woman into Rapunzel in Tangled without changing the facial features." The input video demos are sourced from TalkShow~\cite{yi2023generating} and the Pexels video website.}
    \label{Fig:Stylization_v2v2}
\end{figure}

Video stylization seeks to generate or modify the content into any customized style. For this ability, in addition to being semantic-aware of styles (Figure \ref{Fig:Stylization_t2v1}, \ref{Fig:Stylization_t2v2}), decoupling the structure (\eg, depth), appearance, and motion for training might be necessary. Once these elements are decoupled, style can be solely controlled through text or other signals while preserving the original motion and structure. Additionally, motion modeling and stylization can be implemented separately in this manner. Taking human-centric generation as an example, the separate manner brings the advantage of taking 3D models of humans (\eg, 3D parametric human body model, SMPL \cite{loper2023smpl}) and objects to maintain consistency, as well as free-specify camera trajectory. Eventually, the video generation models play the role of a render, styling the contents, as shown in Figure \ref{Fig:Stylization_v2v1}, \ref{Fig:Stylization_v2v2} . We give some examples generated by Gen-3, and the results already show promising effects.

\clearpage

\subsection{Stability}
\begin{figure}[!ht]
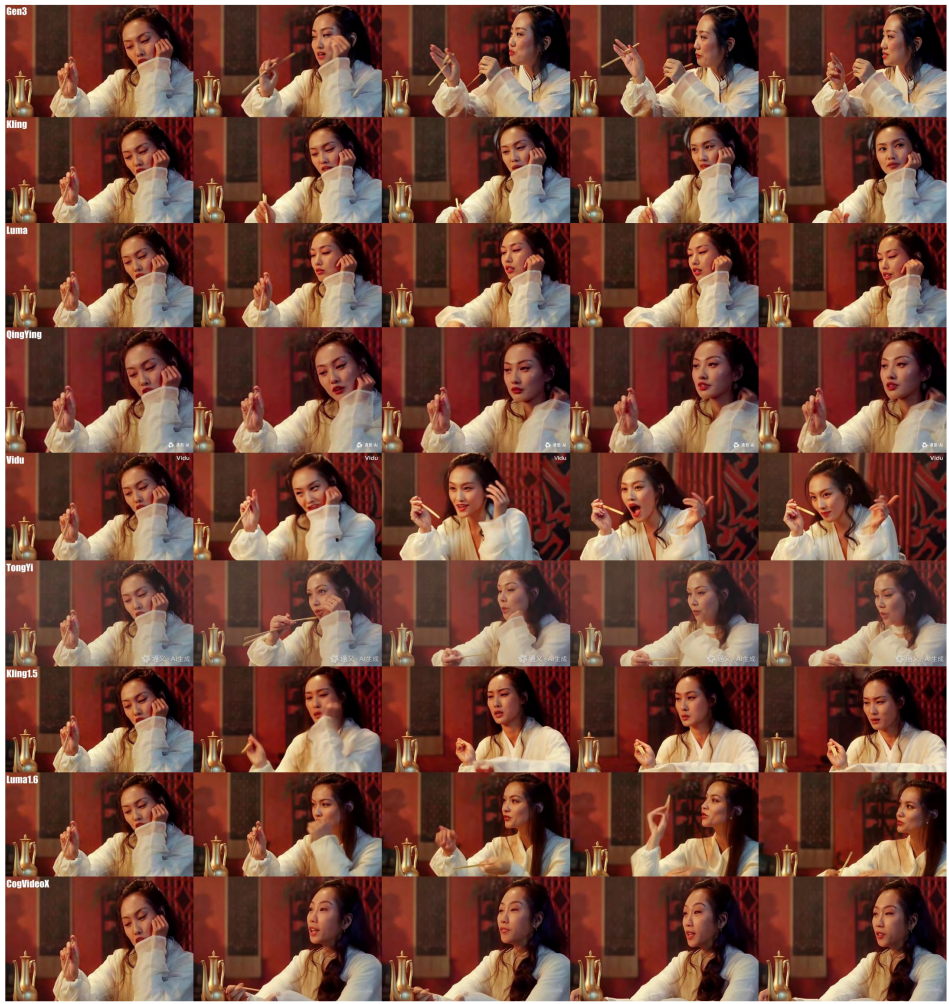

	\centering
	\small
        \begin{overpic}[width=0.9\linewidth]{Figs/Sec3/00145.pdf}

	\end{overpic}
        \caption{\emph{Stability, instance consistency.} Prompt: (I2V-145) "The video is a static medium shot of a woman with long black hair, wearing a white dress, sitting at a table and talking while holding a chopstick. The background features a wall with a red and black design and a silver teapot on the table." Due to partial occlusion and a side view from the given image, many generated videos fail to generate a consistent appearance with natural and text-aligned motions. }
    \label{Fig:Stability_instance}
\end{figure}

\begin{figure}[!ht]
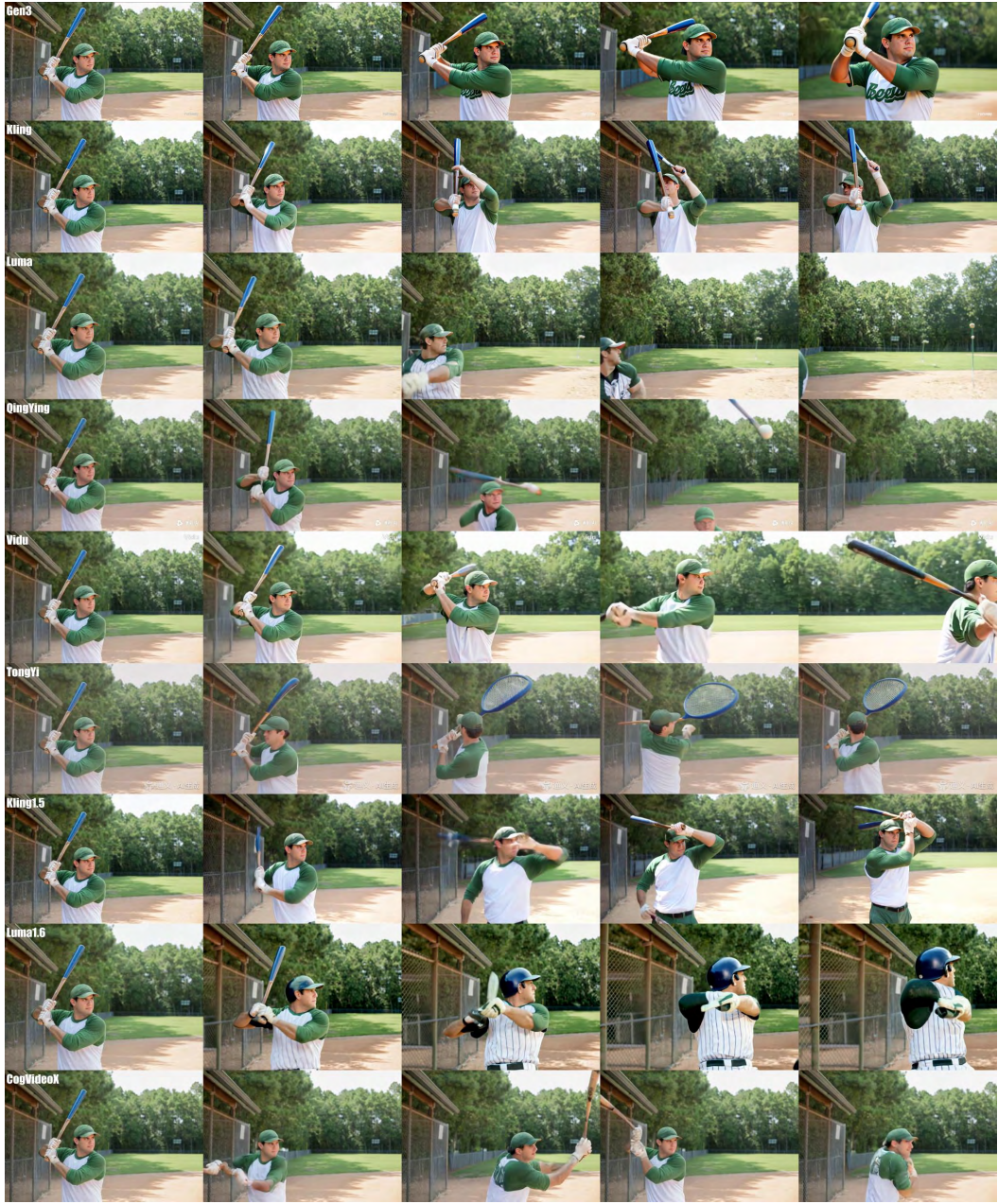

	\centering
	\small
        \begin{overpic}[width=1.\linewidth]{Figs/Sec3/00640.pdf}

	\end{overpic}
        \caption{\emph{Stability, robustness across different text inputs with the same semantic meaning compared to Figure~\ref{Fig:Stability_inputs4}.} Prompt: (I2V-640) "Using both hands, the man holds and swings the bat."}
        \label{Fig:Stability_inputs3}
\end{figure}

\begin{figure}[!ht]
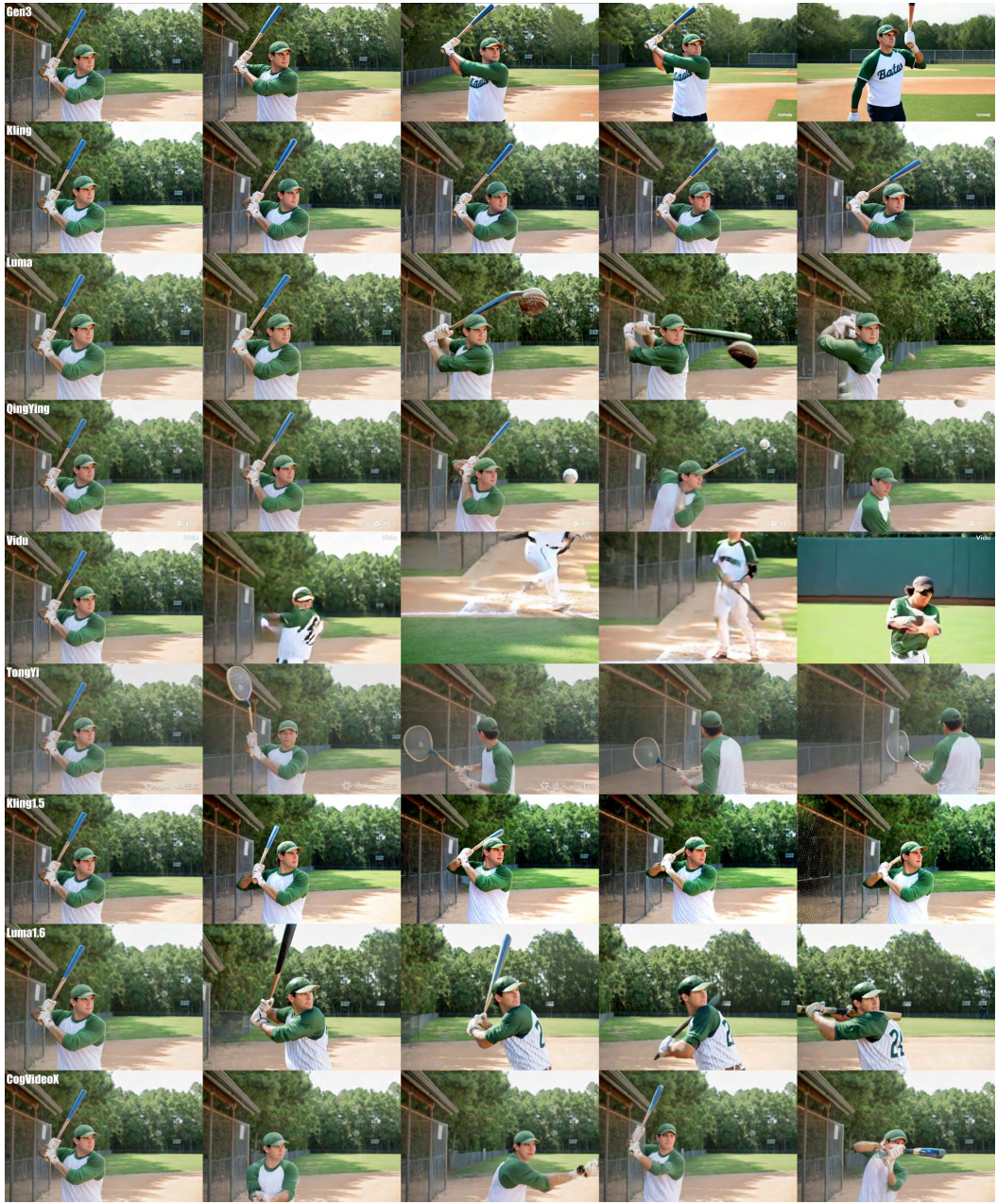

	\centering
	\small
        \begin{overpic}[width=1.\linewidth]{Figs/Sec3/00641.pdf}

	\end{overpic}
        \caption{\emph{Stability, robustness across different text inputs with the same semantic meaning compared to Figure~\ref{Fig:Stability_inputs3}.} Prompt: (I2V-641) "He uses both hands to grip and swing the bat." With different ways of description for the same semantic meaning, the generated motions would change dramatically, indicating the instability of existing models.}
        \label{Fig:Stability_inputs4}
\end{figure}

\begin{figure}[!ht]
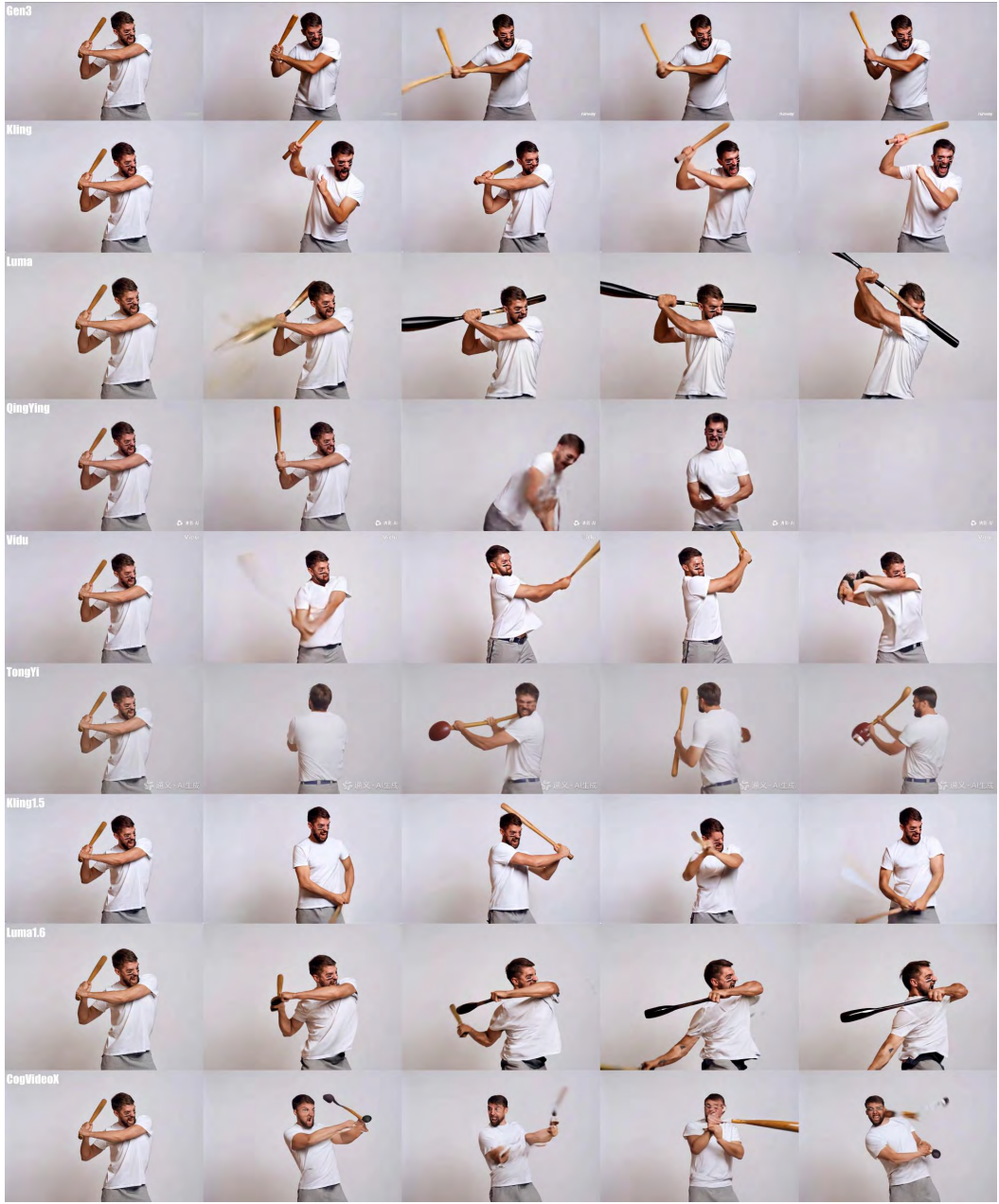

	\centering
	\small
        \begin{overpic}[width=1.\linewidth]{Figs/Sec3/00644.pdf}

	\end{overpic}
        \caption{\emph{Stability, variance introduced by random seeds compared to Figure~\ref{Fig:Stability_seeds2}.} Prompt: (I2V-644) "This person vigorously swings a baseball bat with both hands."}
        \label{Fig:Stability_seeds1}
\end{figure}

\begin{figure}[!ht]
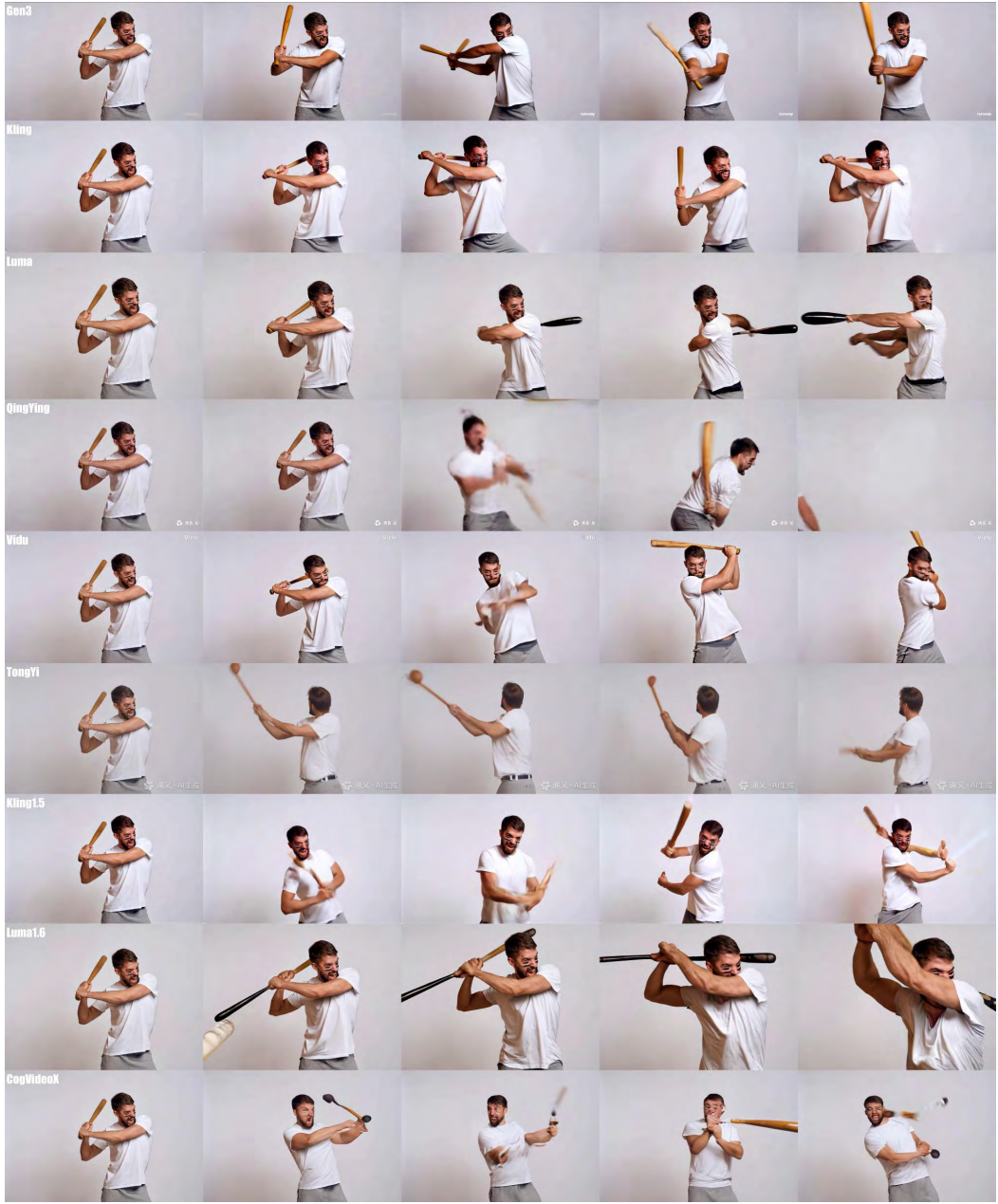

	\centering
	\small
        \begin{overpic}[width=1.\linewidth]{Figs/Sec3/00645.pdf}

	\end{overpic}
        \caption{\emph{Stability, variance introduced by random seeds compared to Figure~\ref{Fig:Stability_seeds1}.} Prompt: (I2V-645) "This person vigorously swings a baseball bat with both hands." Similarly, even the same inputs will generate quite different motions. These results can even range from being effective to completely ineffective animation.}
        \label{Fig:Stability_seeds2}
\end{figure}

Considering real applications, the demand for I2V is stronger than T2V. Since our extensive testing shows that I2V has yet to produce satisfactory results consistently, we outline that stability is quite crucial yet unexplored for I2V at the current stage. Specifically, we outline some key aspects that should be considered to obtain reliable results: \textbf{i)} instance consistency, the model should capture fine-grained features of the input image, \eg, be aware of what instances are in the image, as well as appearances and structures of them. Furthermore, the model should maintain these features during the generation (Figure \ref{Fig:Stability_instance}); \textbf{ii)} robustness across different inputs, for the same input image, handling prompts that express the same content with variations, and also ensuring the same prompt paired with different input images that depict similar contents, could output reliable results (Figure \ref{Fig:Stability_inputs1}, \ref{Fig:Stability_inputs2}, \ref{Fig:Stability_inputs3}, \ref{Fig:Stability_inputs4}); \textbf{iii)} overcoming the variance introduced by random seeds. Random seeds introduce diversity but also contribute to a certain degree of variance. After multiple tests, we observe that current models struggle to consistently generate reliable results, even with identical inputs (Figure \ref{Fig:Stability_seeds1}, \ref{Fig:Stability_seeds2}). In conclusion, existing models still find it hard to generate either stable or diverse videos due to the probabilistic nature of present generative models and the textual annotation variances that may contain various conflict concepts.

\begin{figure}[!ht]
	\centering
	\small
        \begin{overpic}[width=1.\linewidth]{Figs/Sec3/00096.pdf}

	\end{overpic}
        \caption{\emph{Stability, robustness across different input images but the same text compared to Figure~\ref{Fig:Stability_inputs2}.} Prompt: (I2V-96) "The camera remains still, the human is surfing on a wave with the surfboard." Different models generate different camera and subject motions.}
    \label{Fig:Stability_inputs1}
\end{figure}

\begin{figure}[!ht]
	\centering
	\small
        \begin{overpic}[width=1.\linewidth]{Figs/Sec3/00099.pdf}

	\end{overpic}
        \caption{\emph{Stability, robustness across different inputs images but the same text compared to Figure~\ref{Fig:Stability_inputs1}.} Prompt: (I2V-99) "The camera remains still, the human is surfing on a wave with the surfboard." Different models generate different camera and subject motions.}
    \label{Fig:Stability_inputs2}
\end{figure}

\clearpage

\clearpage

\subsection{Motion Diversity}
\begin{figure}[!ht]
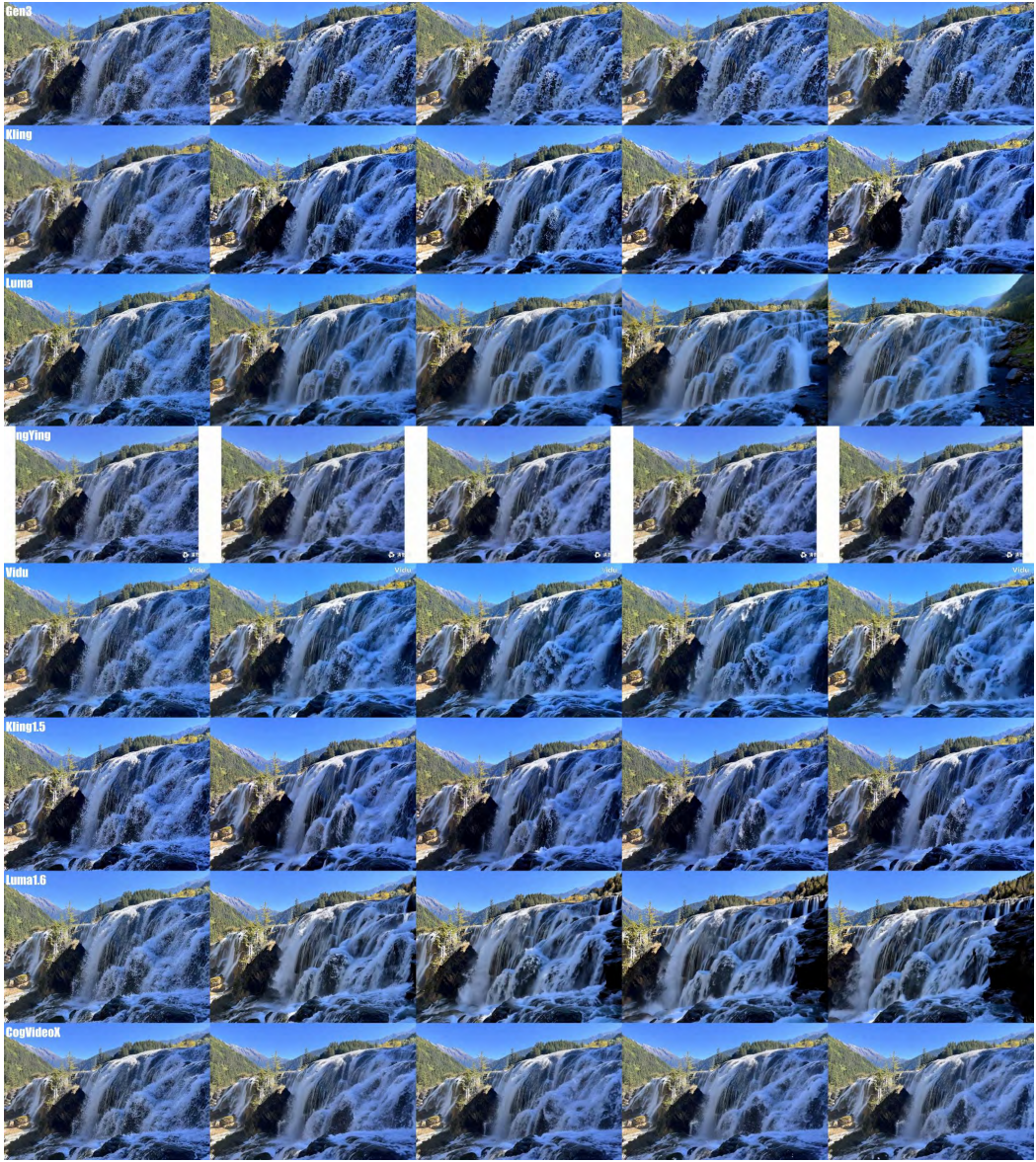

	\centering
	\small
        \begin{overpic}[width=1.\linewidth]{Figs/Sec3/00140.pdf}

	\end{overpic}
        \caption{\emph{Motion diversity, water.} Prompt: (I2V-140) "Static camera, the water from the waterfall hits the rocks, creating splashes of water." Existing models can generate relatively natural and physically plausible motions in landscaping.}
        \label{MotionDiverse_water}
\end{figure}

\begin{figure}[!ht]
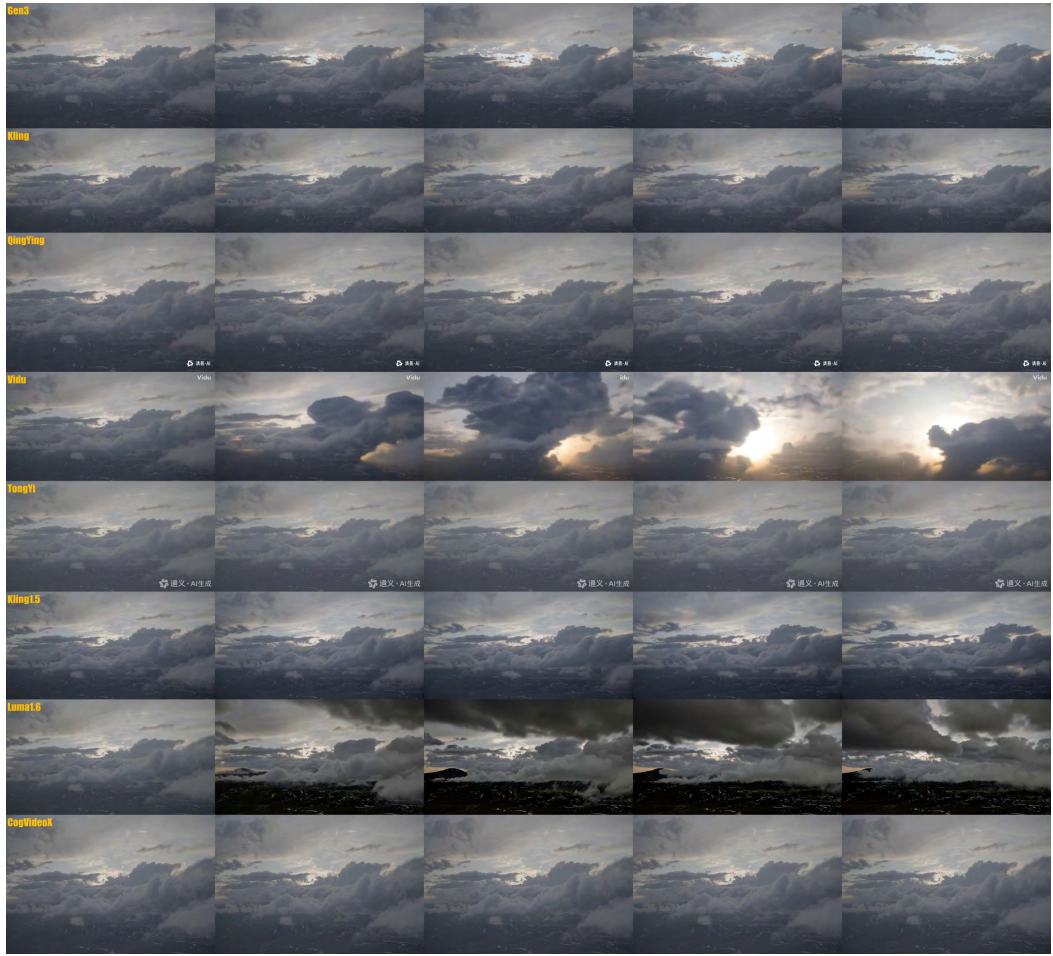

	\centering
	\small
        \begin{overpic}[width=1.\linewidth]{Figs/Sec3/00139.pdf}

	\end{overpic}
        \caption{\emph{Motion diversity, cloud.} Prompt: (I2V-139) "Static camera, the clouds are slowly drifting."}
        \label{MotionDiverse_cloud}
\end{figure}

\begin{figure}[!ht]
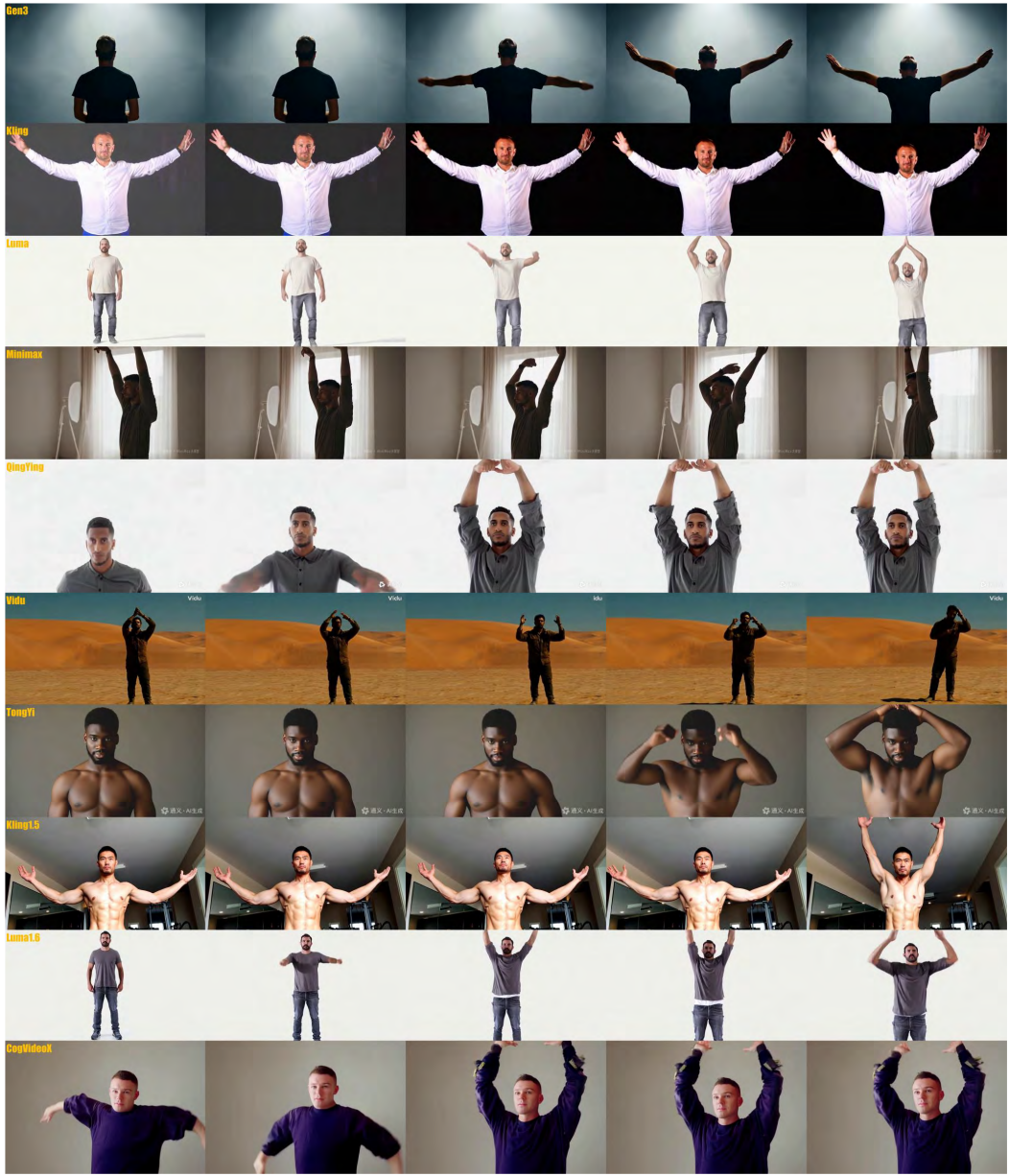

	\centering
	\small
        \begin{overpic}[width=1.\linewidth]{Figs/Sec3/00128.pdf}

	\end{overpic}
        \caption{\emph{Motion diversity, human.} Prompt: (T2V-128) "The camera shots at a man's entire body, and the man raises his hands above his head." The camera and human pose controllability are varied across different models.}
        \label{MotionDiverse_human}
\end{figure}

\begin{figure}[!ht]
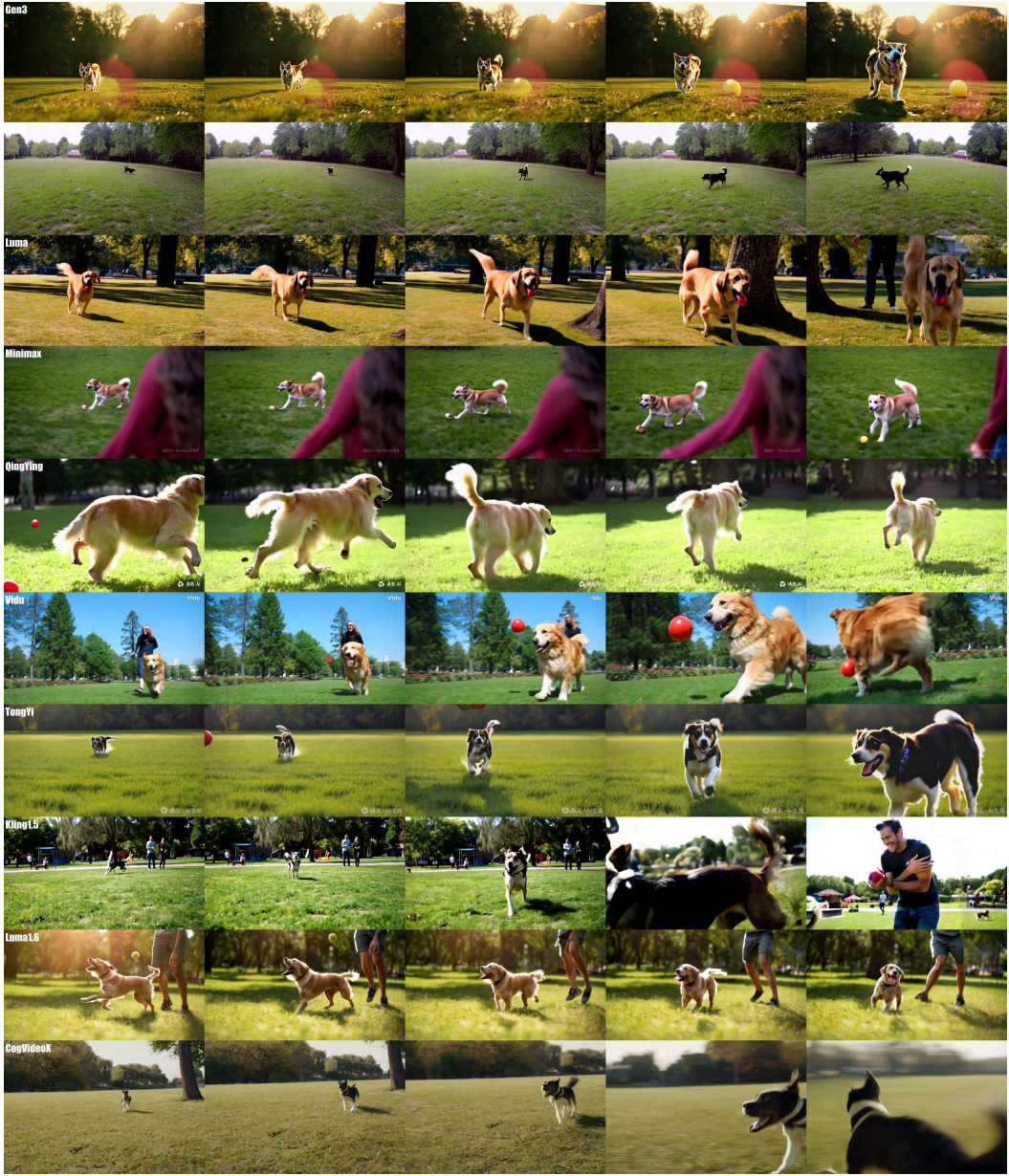

	\centering
	\small
        \begin{overpic}[width=1.\linewidth]{Figs/Sec3/00338.pdf}

	\end{overpic}
        \caption{\emph{Motion diversity, animal \eg, dog.} Prompt: (T2V-338) "A medium shot of a dog running to fetch a ball thrown by its owner in a park. The camera follows the dog's enthusiastic sprint, the wagging of its tail, and the joyful return as it brings the ball back."}
        \label{MotionDiverse_dog}
\end{figure}

\begin{figure}[!ht]
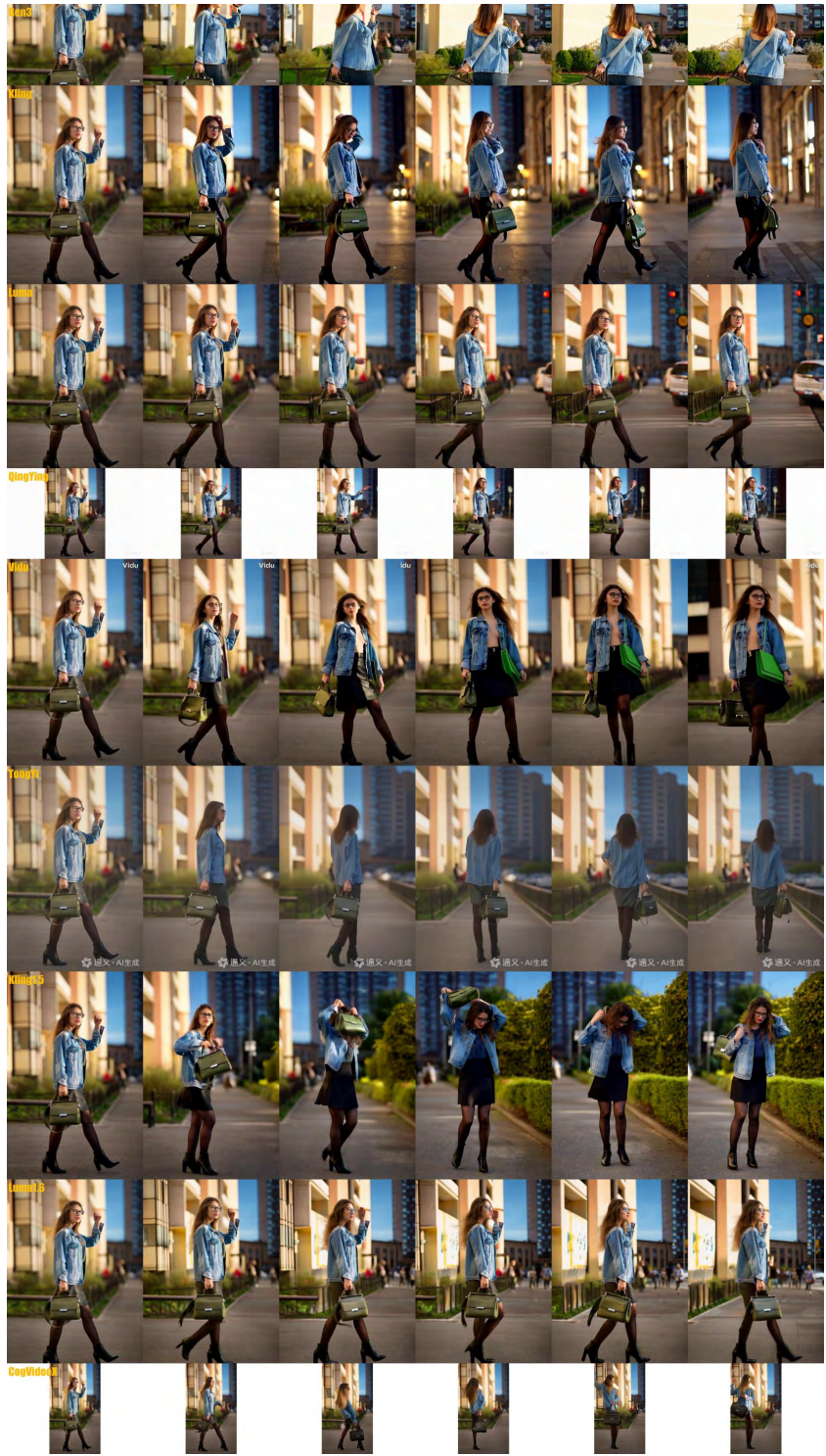

	\centering
	\small
        \begin{overpic}[width=0.8\linewidth]{Figs/Sec3/00046.pdf}

	\end{overpic}
        \caption{\emph{Motion diversity, synergy.} Prompt: (I2V-46) "The camera remains still, the woman is walking while lifting a handbag." Existing image-to-video models often struggle to comprehend input images, especially when dealing with low-resolution objects, and generating motion based on input text. As a result, video generation in a range of interactive motion scenarios frequently exhibits various motion inconsistencies and misunderstandings of physical relationships.}
        \label{MotionDiverse_interaction}
\end{figure}

Generating various motion patterns in nature is the critical and fundamental ability for video generation models. The diverse motions bring challenges, as models need to correspond to specific substances and their motion patterns. Here, we list several types of motion patterns to demonstrate the capabilities of current models, including \textbf{i)} natural elements such as water and clouds (Figure \ref{MotionDiverse_water}, \ref{MotionDiverse_cloud}); \textbf{ii)} biologies such as the human and dog (Figure \ref{MotionDiverse_human}, \ref{MotionDiverse_dog}). In addition to these isolated motion patterns, synergistic motion patterns should also be considered, like \textbf{iii)} interactions, where objects move in sync with human movements (Figure \ref{MotionDiverse_interaction}). Camera motion should also be considered; please refer to Figure \ref{Fig:camera_motion1}, \ref{Fig:camera_motion2}. In addition, we leave detailed motion frequency, velocity, and magnitude results on the website.

\clearpage
\section{Applying SORA-like Models to Ten Real-life Applications}
\label{sec:4}

\label{Sec:application}
From an application-oriented perspective, we first consider which real-world, high-frequency application scenarios could benefit from video generation results, leading us to outline ten existing video application scenarios below. Based on professionally crafted videos from these existing scenarios, we extract the first frame and describe the subsequent motion information over the next few seconds as input to evaluate and compare the performance of these video generation models. Compared to the vertical-domain models discussed in Sec.~\ref{sec:2}, this observation adopts a broader perspective, with inputs drawn from a diverse range of professional or real-life scenario videos.

\begin{figure}[h]
	\centering
	\small
        \begin{overpic}[width=1.\linewidth]{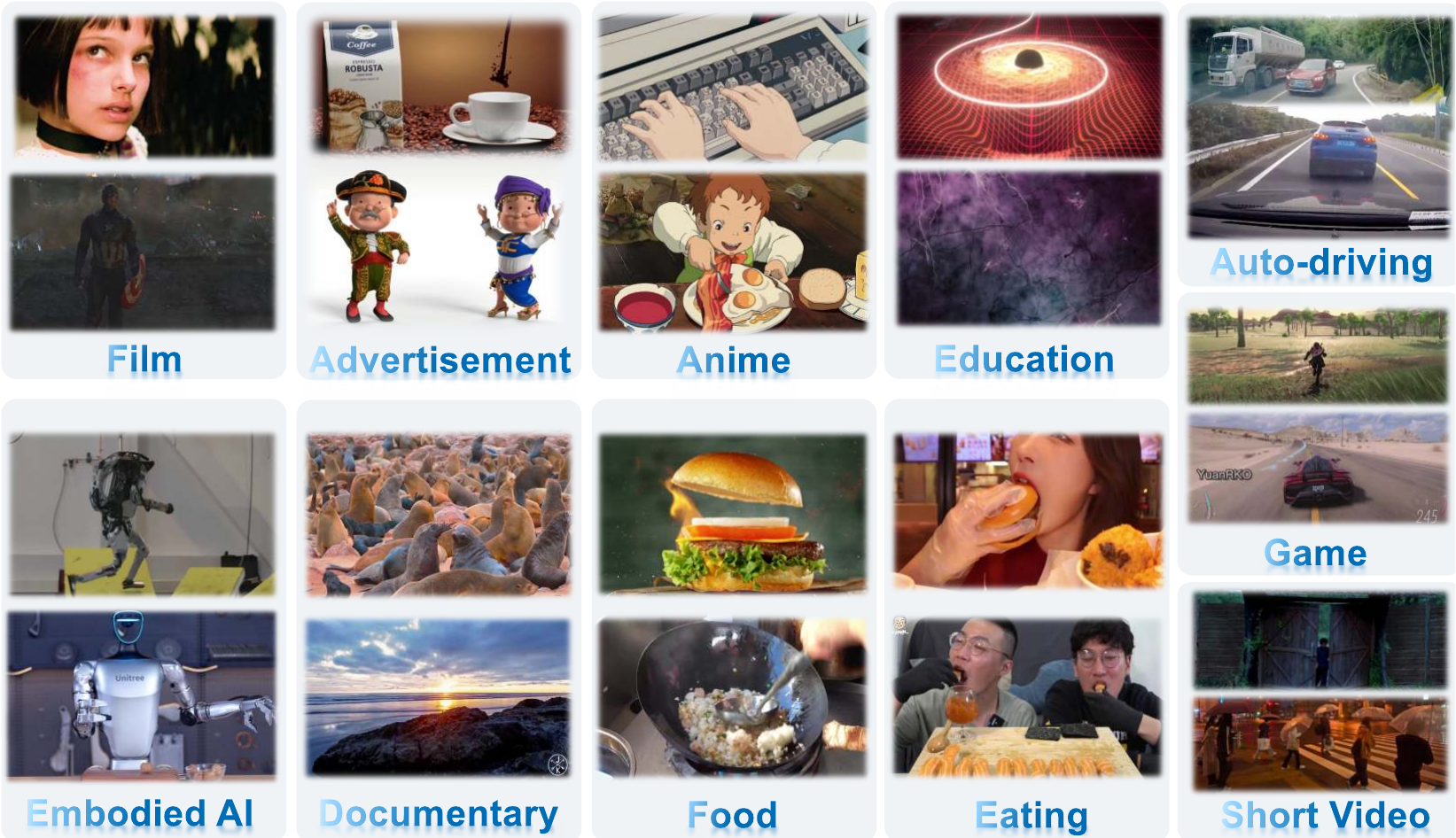}

	\end{overpic}
	\caption{Overview of Section~\ref{sec:5}, we will evaluate the performance of existing SORA-like models in ten key video-applicable scenarios, including filmmaking, advertisement, amine, education, auto-driving, embodied AI, documentary, food and eating show, game, and short videos.}
 \label{Fig:sec4_teaser}
\end{figure}

Specifically, to ensure a comprehensive assessment and construct the application tests, we consider the following evaluation aspects: i) coverage of diverse scenarios, ranging from expert film cuts to specialized contexts, etc. This helps assess the model's capability to adapt to various visual and contextual cues; ii) gradients of generation difficulty, involving variations in the number of subjects, motion complexity, typicality of subjects and actions, and interactions between subjects and their surroundings. This evaluates the model's robustness and generalization capabilities; iii) long shots that provide a continuous view of an event or scene over time, allowing for the assessment of temporal coherence in generated videos, which requires the model to maintain an understanding of the broader context and how the video evolves; iv) video quality, encompassing resolution, frame rate, distortion level, as well as the inclusion of watermarks and subtitles. This is crucial for accurately assessing the visual quality of generated videos. Finally, we summarize all applications in Figure~\ref{Fig:sec4_teaser} and discuss each application as follows.

\begin{figure}[!ht]
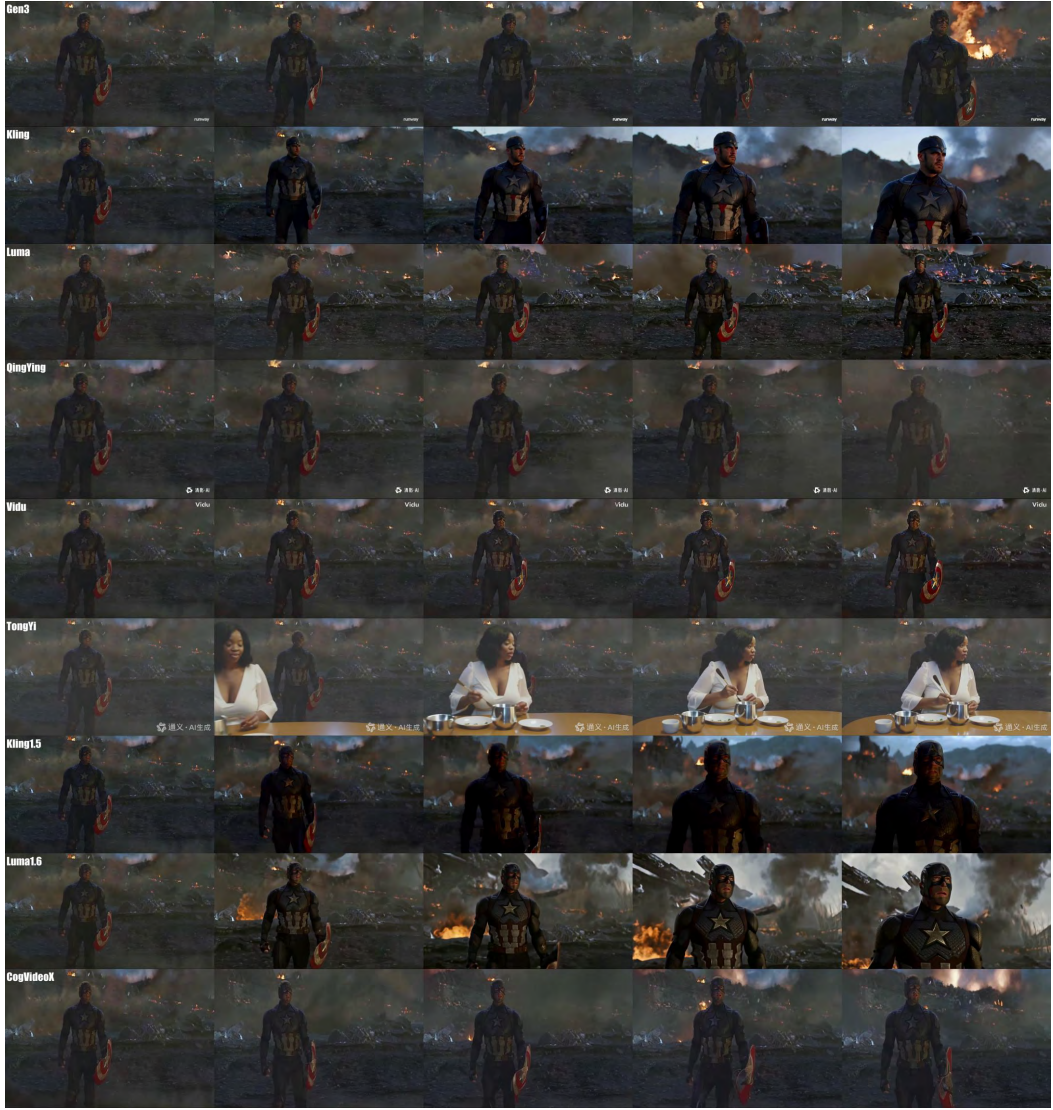

	\centering
	\small
        \begin{overpic}[width=1.\linewidth]{Figs/Sec4/00148.pdf}

	\end{overpic}
	\caption{\emph{For the film-making application.} Prompt: (I2V-148) "The video features a medium shot of a man in a dark superhero costume with a star on his chest, standing in a battle-scarred landscape. The camera remains static throughout the shot." Tongyi exhibit hallucinations by outputting some irrelevant information, while other models perform relatively stably in this scene, with varying abilities in maintaining motion and character consistency.}
 \label{Fig:app_film1}
\end{figure}

\begin{figure}[!ht]
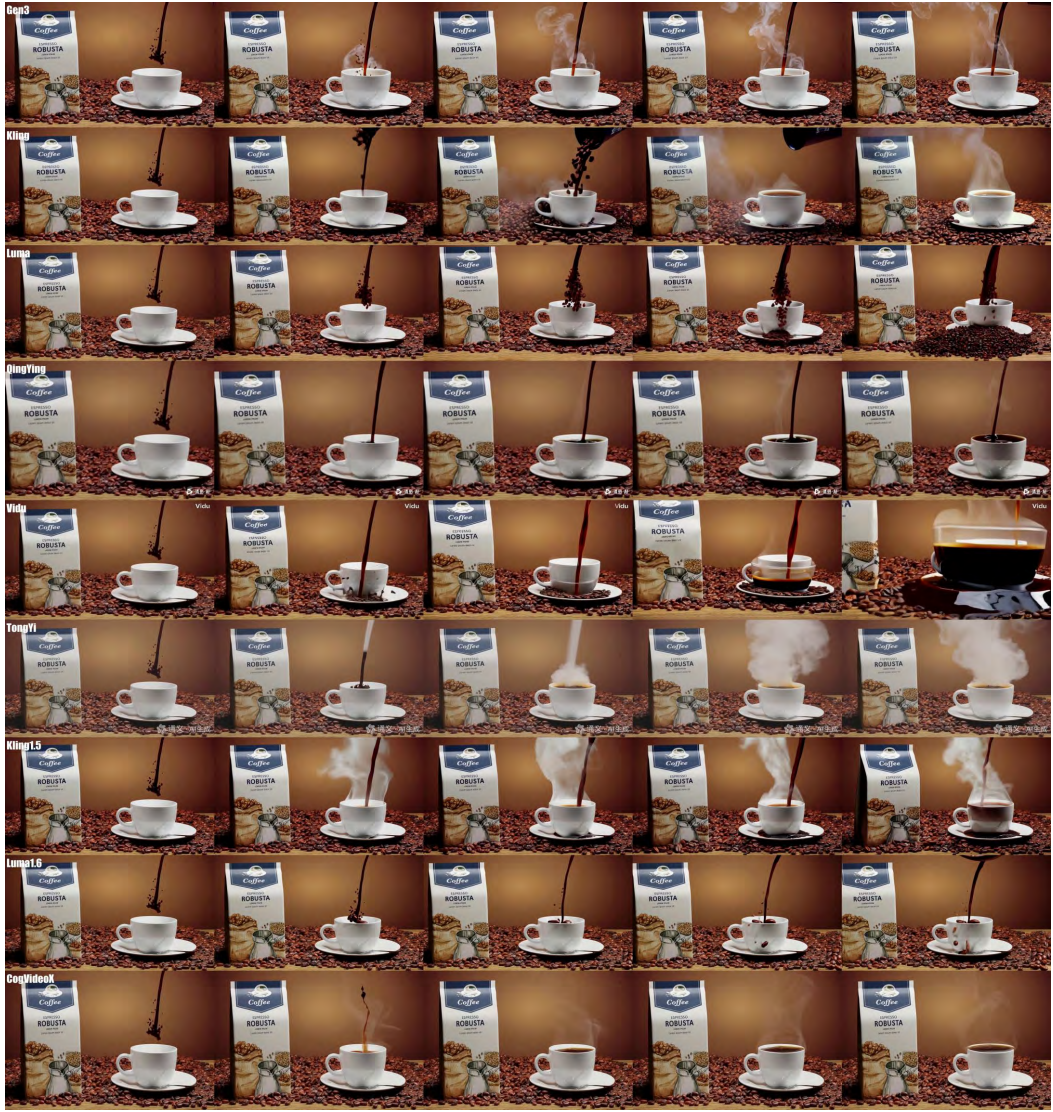

	\centering
	\small
        \begin{overpic}[width=1.\linewidth]{Figs/Sec4/00153.pdf}

	\end{overpic}
	\caption{\emph{For the advertisement application.} Prompt: (I2V-153) "The video is a static, medium shot of a bag of espresso coffee beans and a white coffee cup being filled with coffee. As the coffee fills the cup, steam begins to rise." This scene tests the model's ability to simulate liquid being poured into a cup, followed by physically realistic steam, and the expressiveness of the entire motion; many models struggle to generate the corresponding motion.}
 \label{Fig:app_advertise1}
\end{figure}

\begin{figure}[!ht]
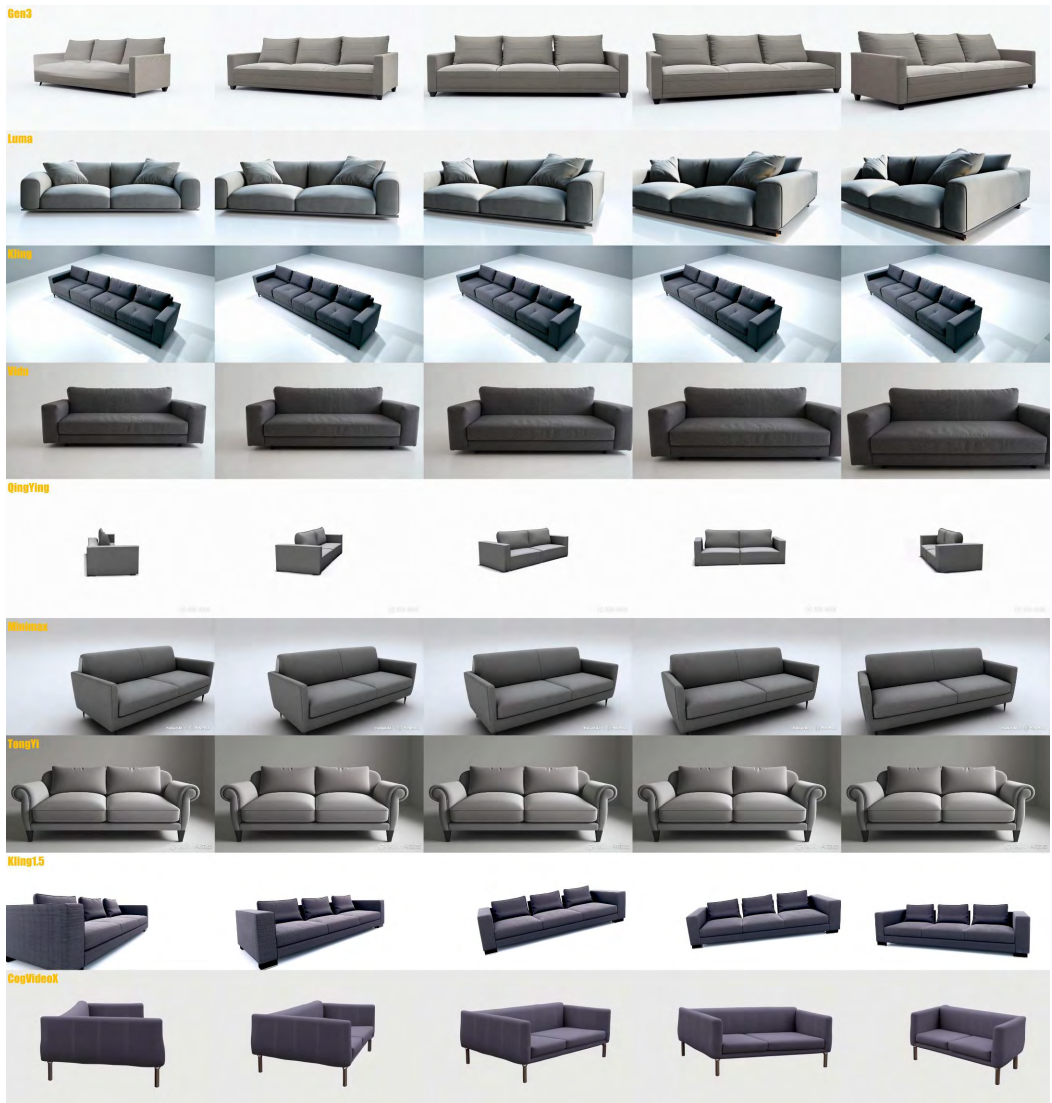

	\centering
	\small
        \begin{overpic}[width=1.\linewidth]{Figs/Sec4/00719.pdf}

	\end{overpic}
	\caption{\emph{For the advertisement application.} Prompt: (T2V-719) "A wide shot high resolution 3D model render of a grey couch against a white background." Most models still struggle to generate a consistent 3D appearance and viewpoints, especially for 360-degree generation.}
  \label{Fig:app_advertise2}
\end{figure}

\begin{figure}[!ht]
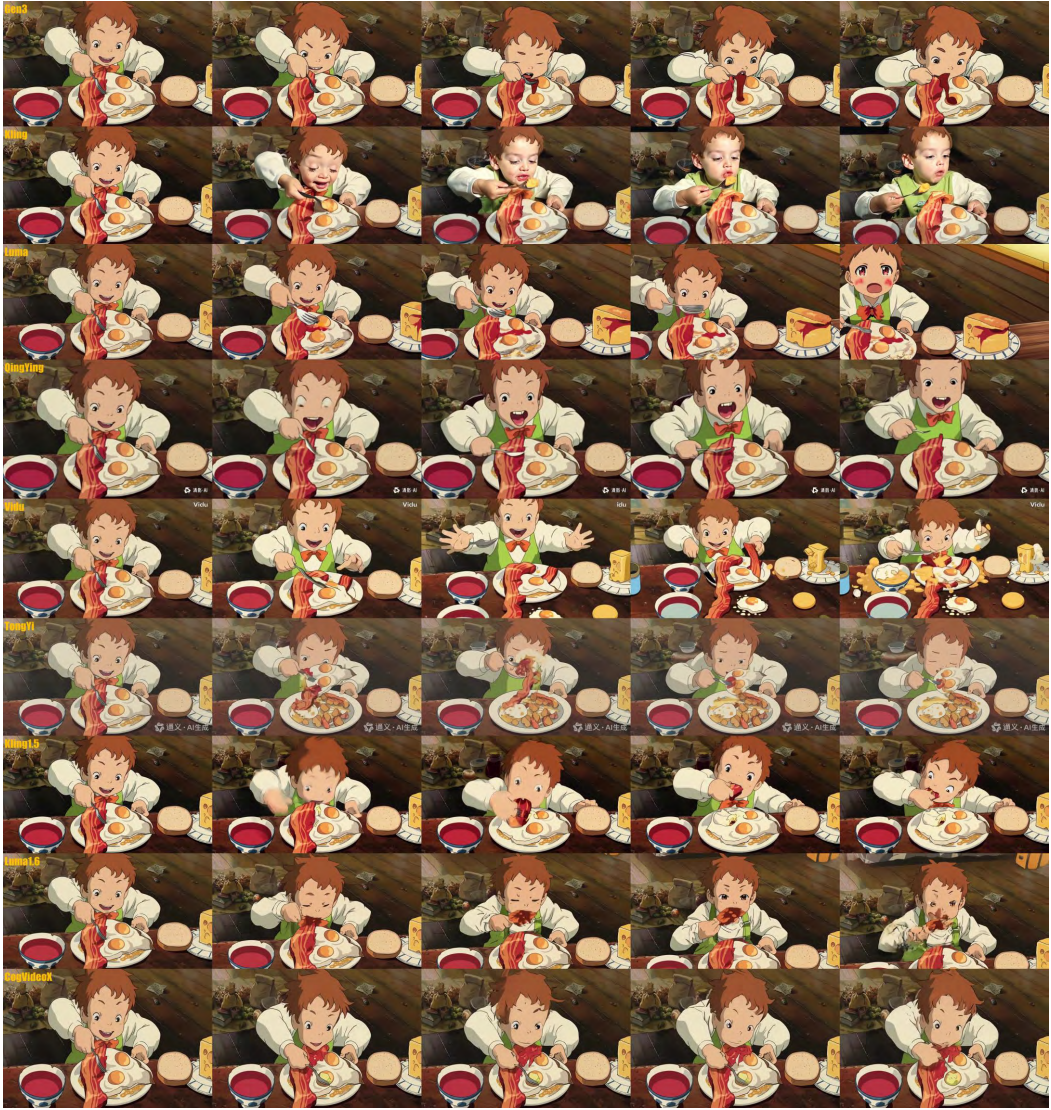

	\centering
	\small
        \begin{overpic}[width=1.\linewidth]{Figs/Sec4/00161.pdf}

	\end{overpic}
	\caption{\emph{For the anime application.} Prompt: (I2V-161) "This is an animated video, with a medium shot, showing a young boy with brown hair hungrily eating eggs and bacon off of a plate. The boy eats quickly and messily, getting food on his face." Maintaining the consistency of an animated character's identity and the completeness of its actions during motion remains quite challenging.}
 \label{Fig:app_anime}
\end{figure}

\begin{figure}[!ht]
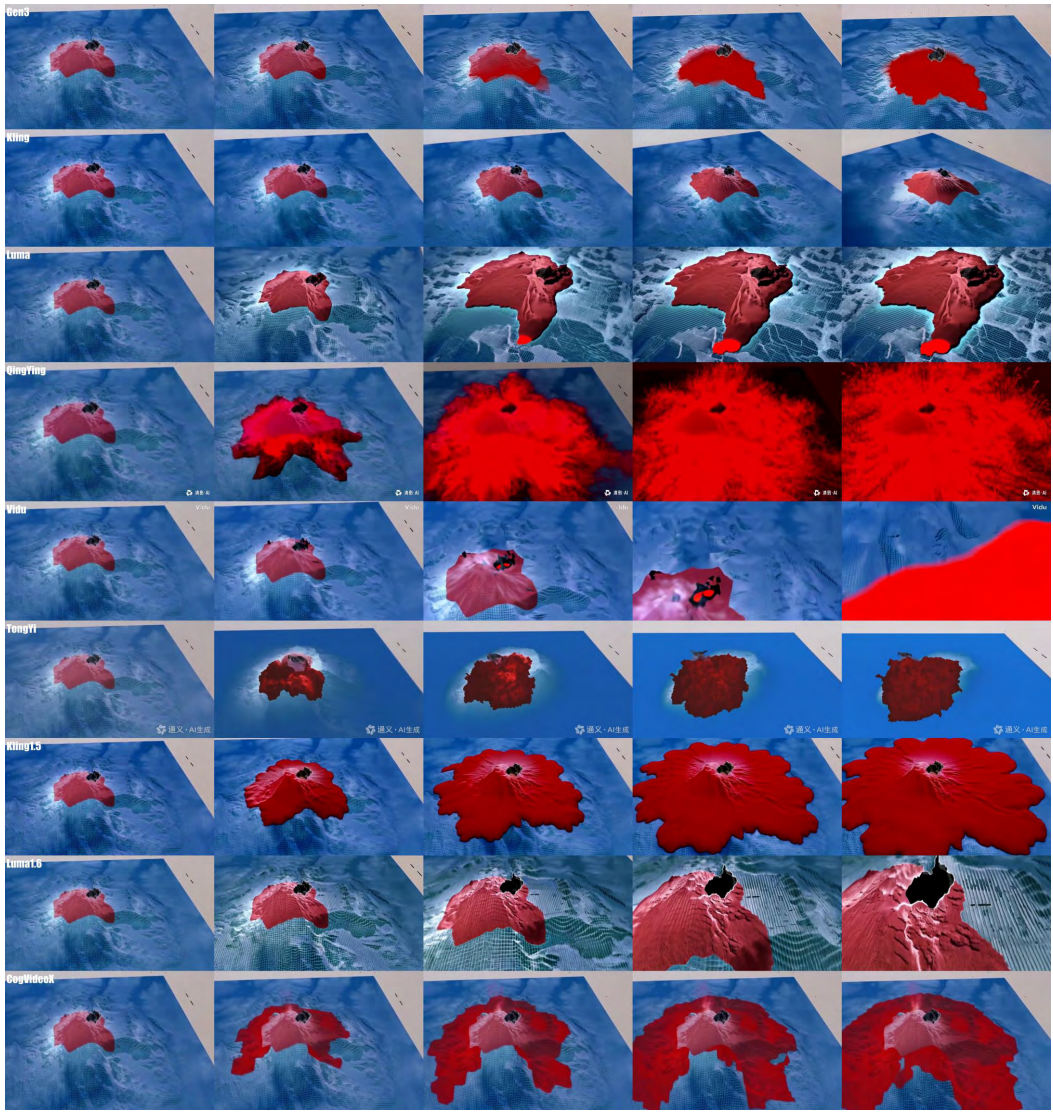

	\centering
	\small
        \begin{overpic}[width=1.\linewidth]{Figs/Sec4/00167.pdf}

	\end{overpic}
	\caption{\emph{For the education application.} Prompt: (I2V-167) "The camera provides an aerial view, zooming in on a 3D model of a volcano as a red area expands outwards from the peak. The video uses animation to depict the potential spread of a volcanic flow." Maintaining the engagement and attention of students requires high-quality, contextually relevant animations, which can be difficult for models to consistently produce across diverse educational topics.}
 \label{Fig:app_education1}
\end{figure}

\begin{figure}[!ht]
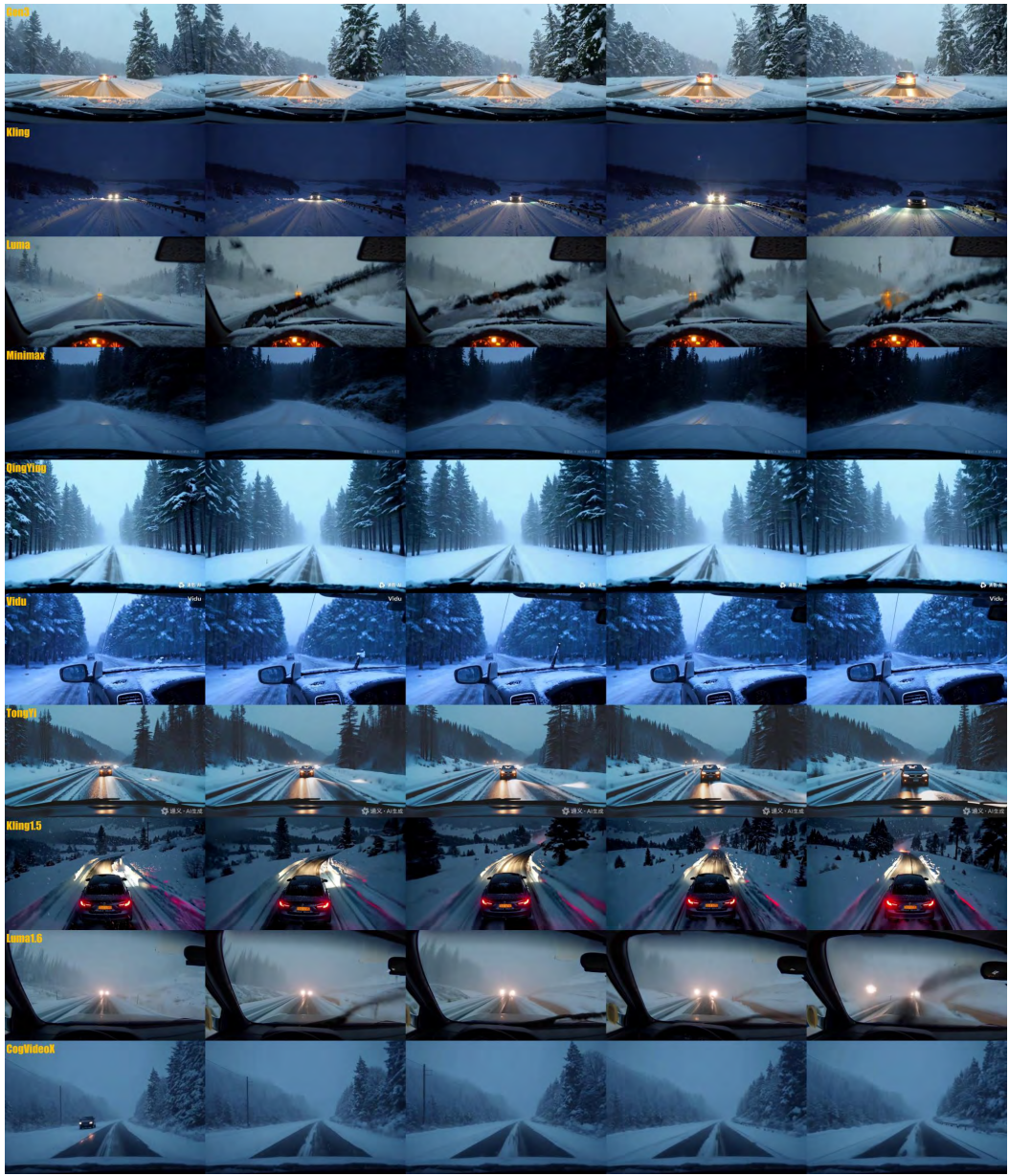

	\centering
	\small
        \begin{overpic}[width=1.\linewidth]{Figs/Sec4/00419.pdf}

	\end{overpic}
	\caption{\emph{For the auto-driving application.} Prompt: (I2V-419) "FPV in the car driving through heavy snowfall. The scene should depict a car traveling on a highway in a rural, wilderness area. Snow should be falling heavily, covering the road and surrounding landscape. The lighting should be dim, creating a serene and somewhat desolate atmosphere. The road should be visible but obscured by the snow, with the headlights of the car illuminating the path ahead." Ensuring temporal consistency and realistic motion behavior of vehicles, pedestrians, and other elements is also difficult, as errors in these aspects could lead to unsafe decision-making in real-world driving scenarios.}
 \label{Fig:app_autodrive}
\end{figure}

\begin{figure}[!ht]
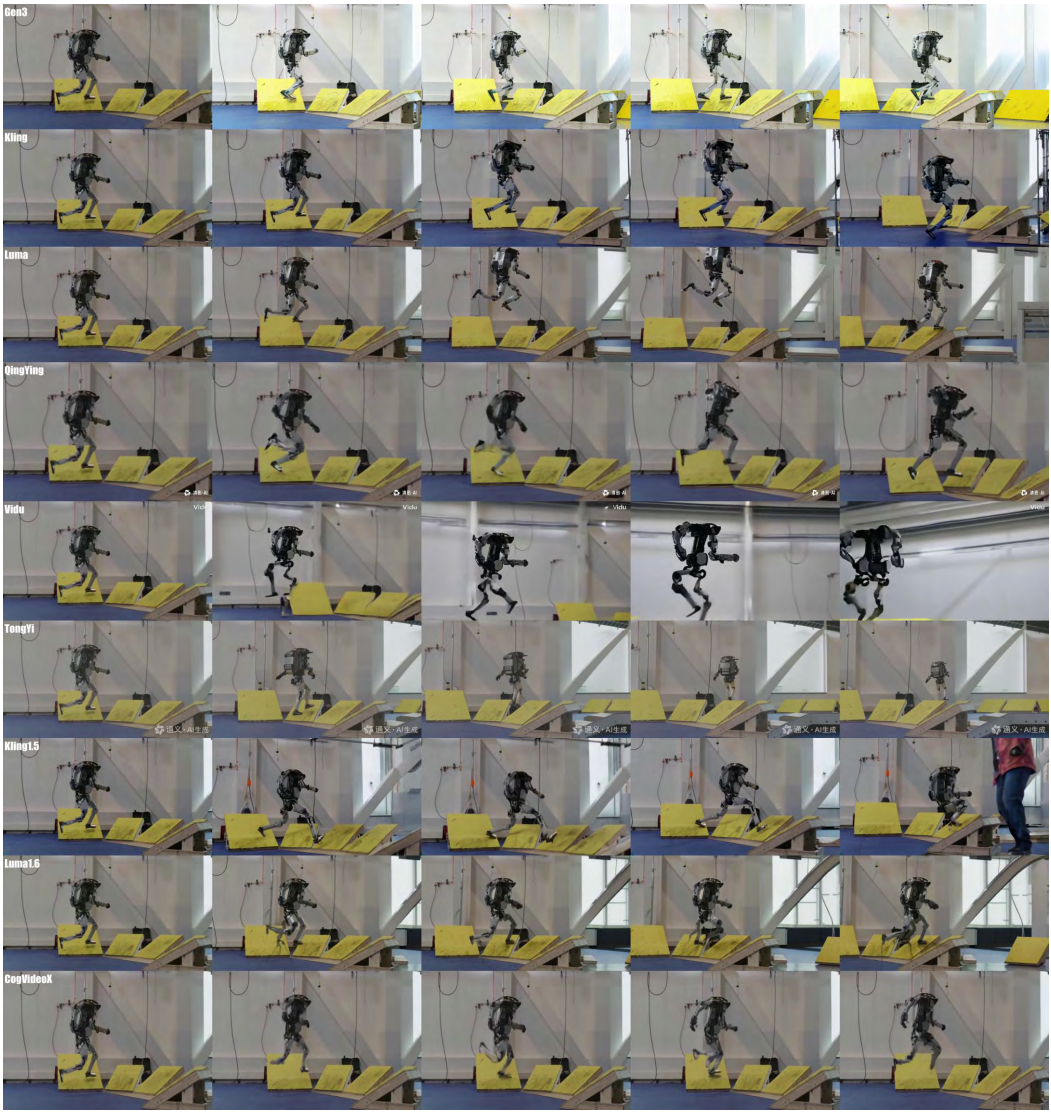

	\centering
	\small
        \begin{overpic}[width=1.\linewidth]{Figs/Sec4/00178.pdf}

	\end{overpic}
	\caption{\emph{For the Embodied AI application.} Prompt: (I2V-178) "The video is a long shot depicting a white humanoid robot running across a series of yellow platforms in a brightly lit indoor space. The camera remains static throughout the video." Due to different motion patterns between humans and human-like robots, generating coherent and contextually appropriate motion sequences across different robots to simulate the movements is still challenging.}
 \label{Fig:app_embodied1}
\end{figure}

\begin{figure}[!ht]
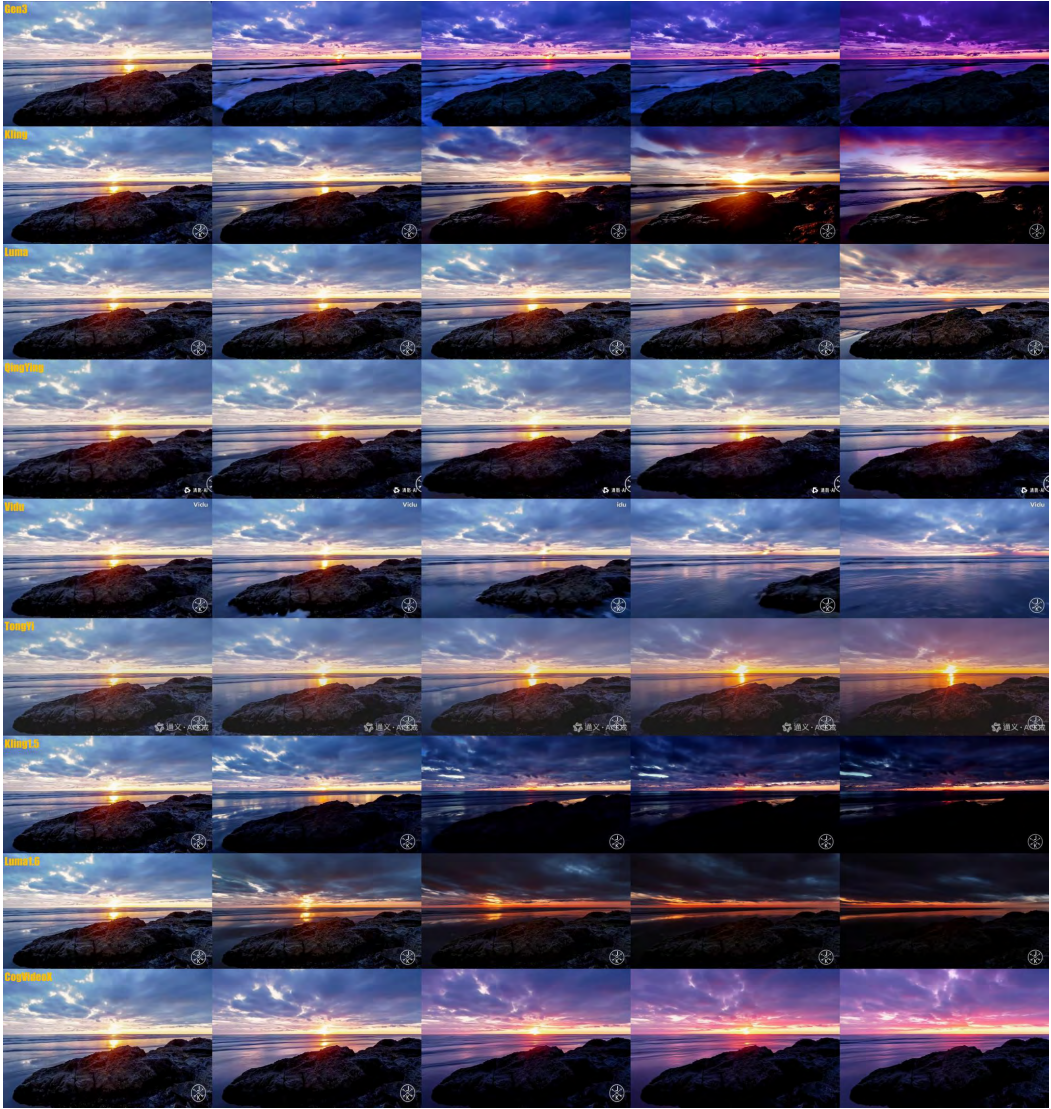

	\centering
	\small
        \begin{overpic}[width=1.\linewidth]{Figs/Sec4/00185.pdf}

	\end{overpic}
	\caption{\emph{For the documentary application.} Prompt: (I2V-185) "The video is a static wide shot of a sunset over the ocean with rocks in the foreground. As the sun sets, the sky changes colors from purple to dark blue." The precise color and state changes for the weather and scenes are relative easy for existing models, especially for Kling.}
 \label{Fig:app_documentary1}
\end{figure}

\begin{figure}[!ht]
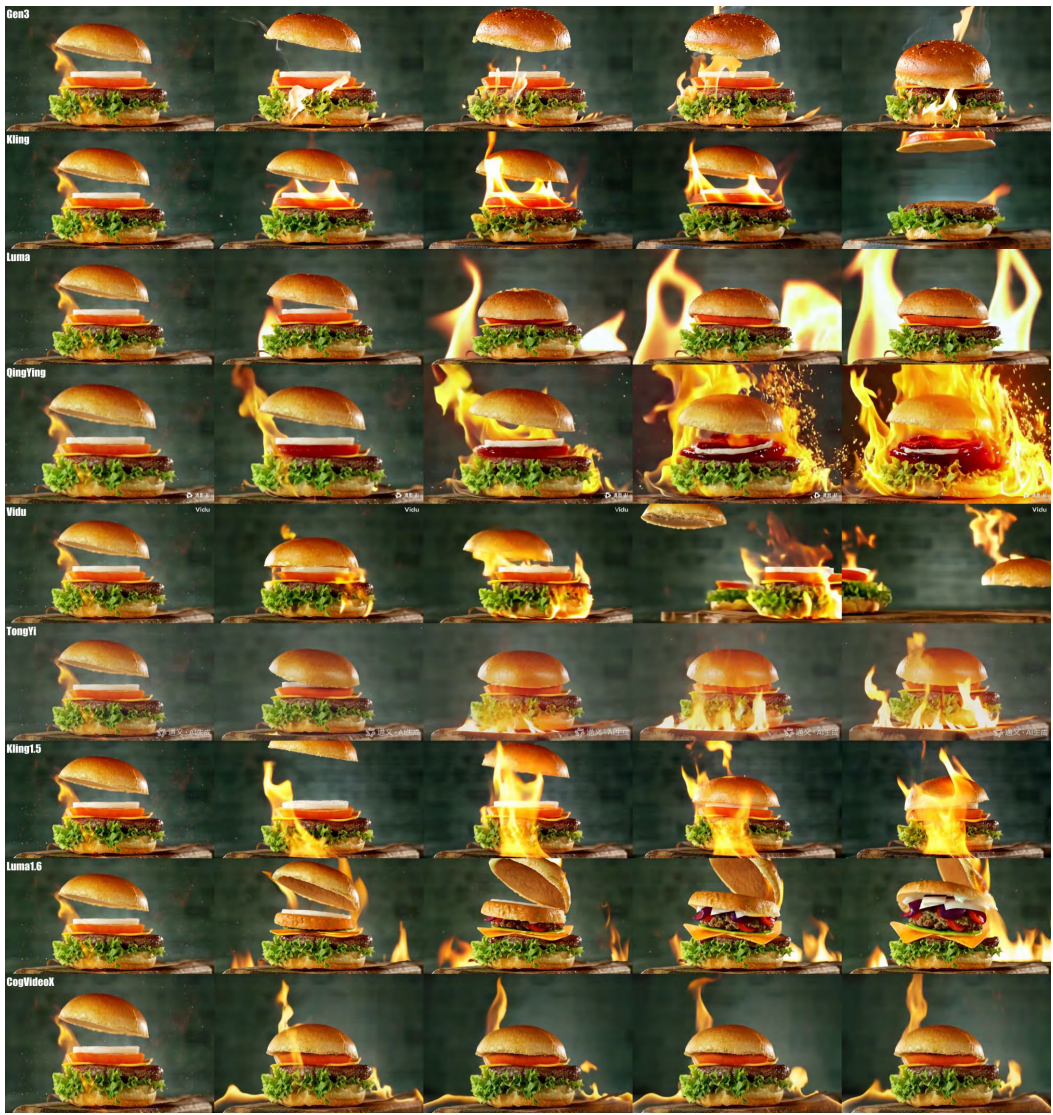

	\centering
	\small
        \begin{overpic}[width=1.\linewidth]{Figs/Sec4/00190.pdf}

	\end{overpic}
	\caption{\emph{For the food content creation application.} Prompt: (I2V-190) "The video features a medium shot of a delicious-looking burger with the top bun levitating above the rest of the ingredients. As the video progresses, flames appear on the wooden board beneath the burger, creating a visually appealing contrast against the dark background." The models perform well overall, but there are some visually implausible floating, \eg, Vidu.}
 \label{Fig:app_food}
\end{figure}

\begin{figure}[!ht]
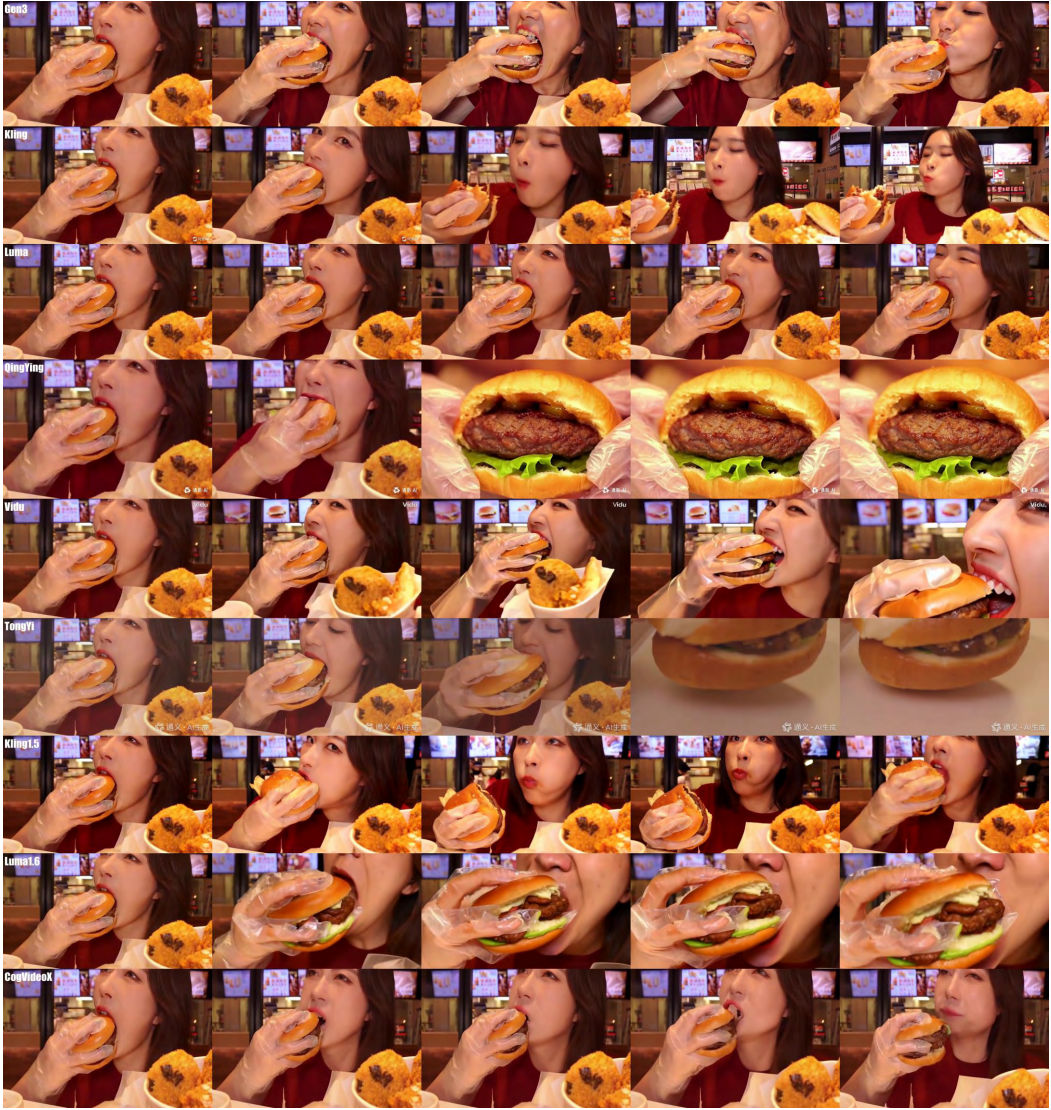

	\centering
	\small
        \begin{overpic}[width=1.\linewidth]{Figs/Sec4/00196.pdf}

	\end{overpic}
	\caption{\emph{For the eating show application.} Prompt: (I2V-196) "The video starts with a close-up shot of a woman taking a bite of a burger. Then the camera zooms in for an extreme close-up shot of the burger as the woman holds it up." Kling 1.0, 1.5, luma and CogVideoX perform well, other models exhibite issues such as abnormal facial distortions and camera moving errors.}
 \label{Fig:app_eating}
\end{figure}

\begin{figure}[!ht]
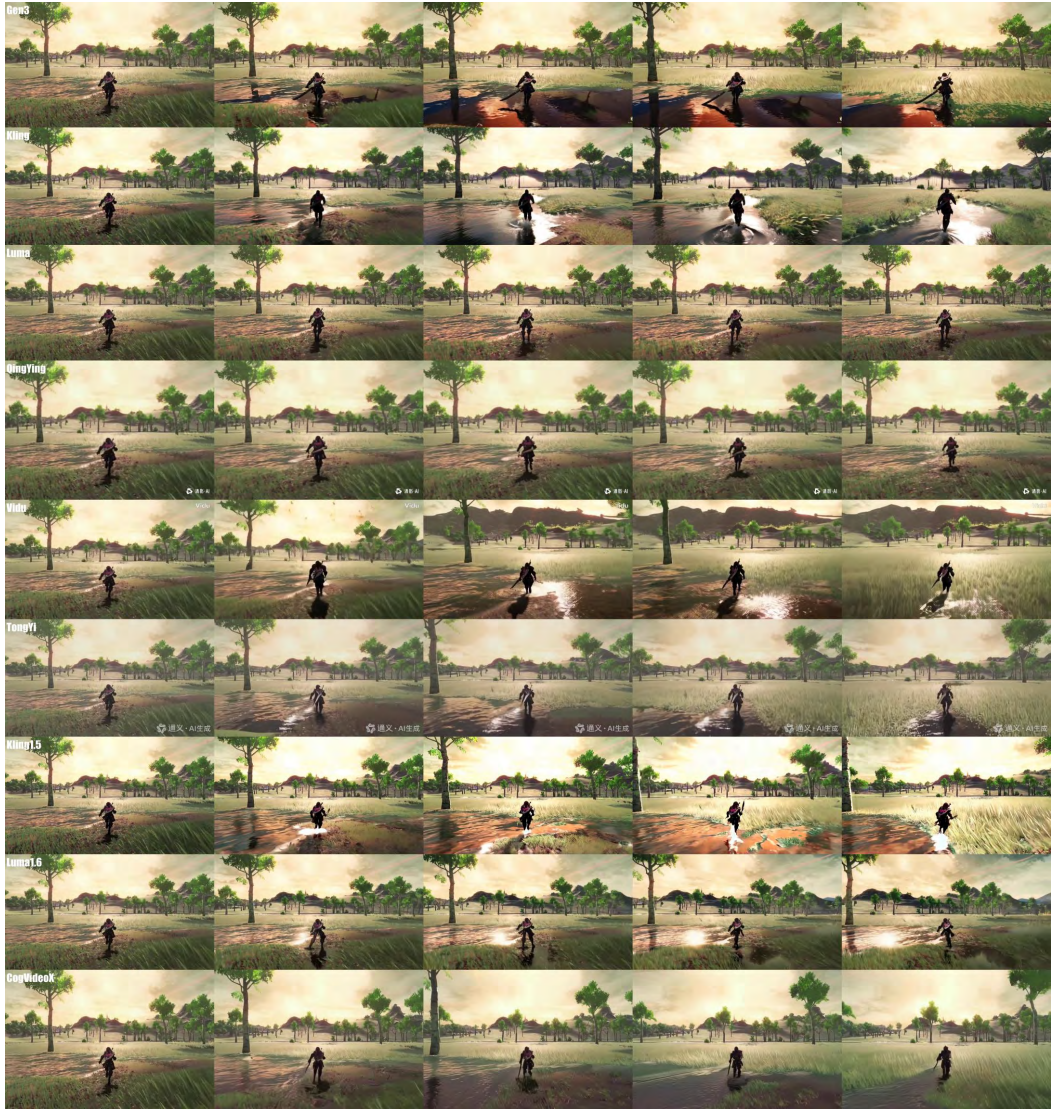

	\centering
	\small
        \begin{overpic}[width=1.\linewidth]{Figs/Sec4/00200.pdf}

	\end{overpic}
	\caption{\emph{For the game application.} Prompt: (I2V-200) "The video features a long shot of a video game character, clad in dark armor and carrying a sword, wading through a shallow stream towards a distant structure. The camera remains static throughout the shot, capturing the character's progress through the grassy plain." From the third-person perspective, the models basically maintain the appearance of the character and the scene, while also capturing the character’s movement.}
 \label{Fig:app_game1}
\end{figure}

\begin{figure}[!ht]
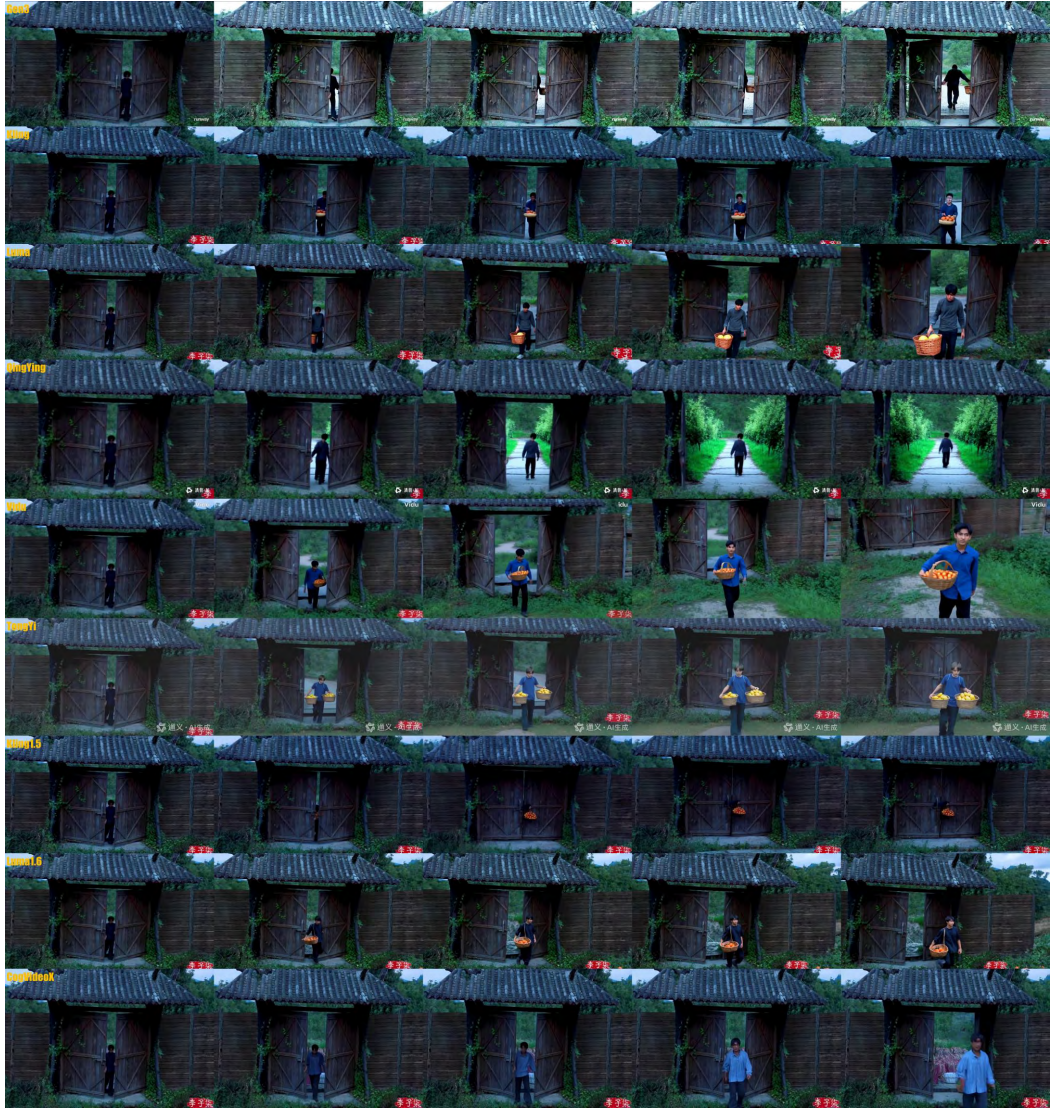

	\centering
	\small
        \begin{overpic}[width=1.\linewidth]{Figs/Sec4/00205.pdf}

	\end{overpic}
	\caption{\emph{For the short video creation application.} Prompt: (I2V-205) "The video uses a static, medium-long shot to show a young man carrying a basket of persimmons, opening a set of wooden doors, and walking through them. The video has a realistic style." Since the character information in the first frame is rather vague, higher demands are placed on the model to generate a reasonable appearance and motion.}
 \label{Fig:app_shortvideo}
\end{figure}

\begin{itemize}

\item \textbf{Film-making}. Video generation models can assist in creating visual effects, close-up shots, dynamic camera movements, etc., which can significantly reduce the costs associated with film shooting and production (Figure \ref{Fig:app_film1}). However, for practical applications, ensuring the consistency of characters during multi-shot transitions or significant camera movements is essential. Characters' emotional expressions and storytelling capabilities would be enhanced.

\item \textbf{Advertisement}. Generating commercial advertisement videos is directly aimed at business applications. However, showcasing products comprehensively in advertisements typically involves 360-degree product rotation display, various physical phenomena, and interactions (Figure \ref{Fig:app_advertise1}, \ref{Fig:app_advertise2}), which remain highly challenging for current video generation models.

\item \textbf{Anime}. This production remains labor-intensive due to its frame-by-frame manual drawing nature. Cartoon frames often exhibit temporal sparsity and large motions due to the high cost of drawing, resulting in more frequent occurrences of textureless color regions compared to live-action videos~\cite{xing2024tooncrafter}. 
Based on the current capabilities demonstrated by the video generation model, they hold the potential to streamline the production of animated content by automatically generating character animations (Figure \ref{Fig:app_anime}). This can significantly reduce the time-consuming manual work required in traditional animation processes. However, replicating the unique art styles and emotional expressions of anime characters is still a major challenge, especially when dealing with complex emotional shifts.

\item \textbf{Education}. Video generation models enhance educational content creation by generating animated tutorials, interactive visualizations, and scenario-based learning videos (Figure \ref{Fig:app_education1}). These tools allow for personalized learning experiences and help explain complex concepts through dynamic illustrations. However, generating accurate educational content that maintains pedagogical clarity and ensures the fidelity of scientific or technical representations is still hard for current models, particularly when adapting to diverse subject matters or instructional styles. 

\item \textbf{Auto-driving} Regarding auto-driving, video generation models can play a critical role in simulating real-world driving scenarios for autonomous vehicle training, including various weather conditions, traffic situations, and pedestrian behaviors (Figure \ref{Fig:app_autodrive}). These simulations can help reduce the reliance on costly and time-consuming real-world data collection. However, accurately generating dynamic environments with complex interactions between vehicles, pedestrians, and road conditions poses significant challenges. Meanwhile, the conditions could be considered for injection into existing models.

\item \textbf{Embodied AI}. Through video generation, there is potential to generate motion trajectories of certain configurations (Figure \ref{Fig:app_embodied1}), which can promote the generation of training data and planning. However, it is also not easy to understand the motion patterns of various mechanical structures and adapt to their motion speed.

\item \textbf{Documentary}. Video generation models assist in creating dynamic visualizations for documentaries, such as historical reenactments, time-lapse sequences, aerial shots, etc. (Figure \ref{Fig:app_documentary1}). This can reduce the need for on-location shooting and extensive post-production work, allowing for cost-effective content creation. However, maintaining historical accuracy, storytelling abilities, and ensuring the authenticity of real-world events and environments remain significant challenges for current models, especially when dealing with intricate details or culturally significant scenes. Much of the footage in documentaries still comes from live-action filming, making it crucial to ensure continuity and stylistic consistency between generated content and real footage.

\item \textbf{Food and Eating}. The creation of food-related content may be revolutionized by video generation models, from recipe demonstrations to visually appealing food commercials (Figure \ref{Fig:app_food}). These models can generate cooking sequences, ingredient transformations, and various preparation techniques, reducing the need for repeated filming of complex cooking processes. Capturing the textures, colors, and nuanced presentation of food in a way that evokes appetite and realism remains a considerable challenge. Especially when showcasing subtle details such as steam, melting, or crispiness. Additionally, based on food content generation, the increasingly popular eating shows (Figure \ref{Fig:app_eating}) may also be revolutionized. This scenario involves complex interactions, physical laws, and causal phenomena, making it highly challenging to generate convincingly through current video generation models.

\item \textbf{Game}. For the gaming industry, taking video generation models to create content (Figure \ref{Fig:app_game1}) is not enough to address the core challenges. In gaming, interactive and precise controls are far more critical than merely maintaining visual and motion consistency. Currently, closed-source models struggle to achieve the level of precision required for these elements.

\item \textbf{Short Video}. Short videos are suitable for current video generation models as they do not rely on quite long-term content. The generation models can streamline the production of short-form content for social media, such as dynamic transitions, animated text effects, and engaging visual stories (Figure \ref{Fig:app_shortvideo}).
\end{itemize}

In summary, SORA-like video generation models have demonstrated promising capabilities across various applications, such as advertisement, education, and animation, where they contribute to creative visual content and special effects. However, they still face challenges in capturing complex motions, making the generated videos easy to enhance and improve for specific requirements, maintaining appearance and motion consistency, high aesthetics, and understanding intricate interactions and physical logic, which limits their ability to fully meet the demands of more dynamic and detailed real-world scenarios. Further advancements are needed to enhance their controllability, adaptability, accuracy, and realism in diverse applications.

\clearpage
\section{Exploration of Usage Scenarios and Tasks}
\label{sec:5}

\begin{figure}[h]
	\centering
	\small
        \begin{overpic}[width=0.95\linewidth]{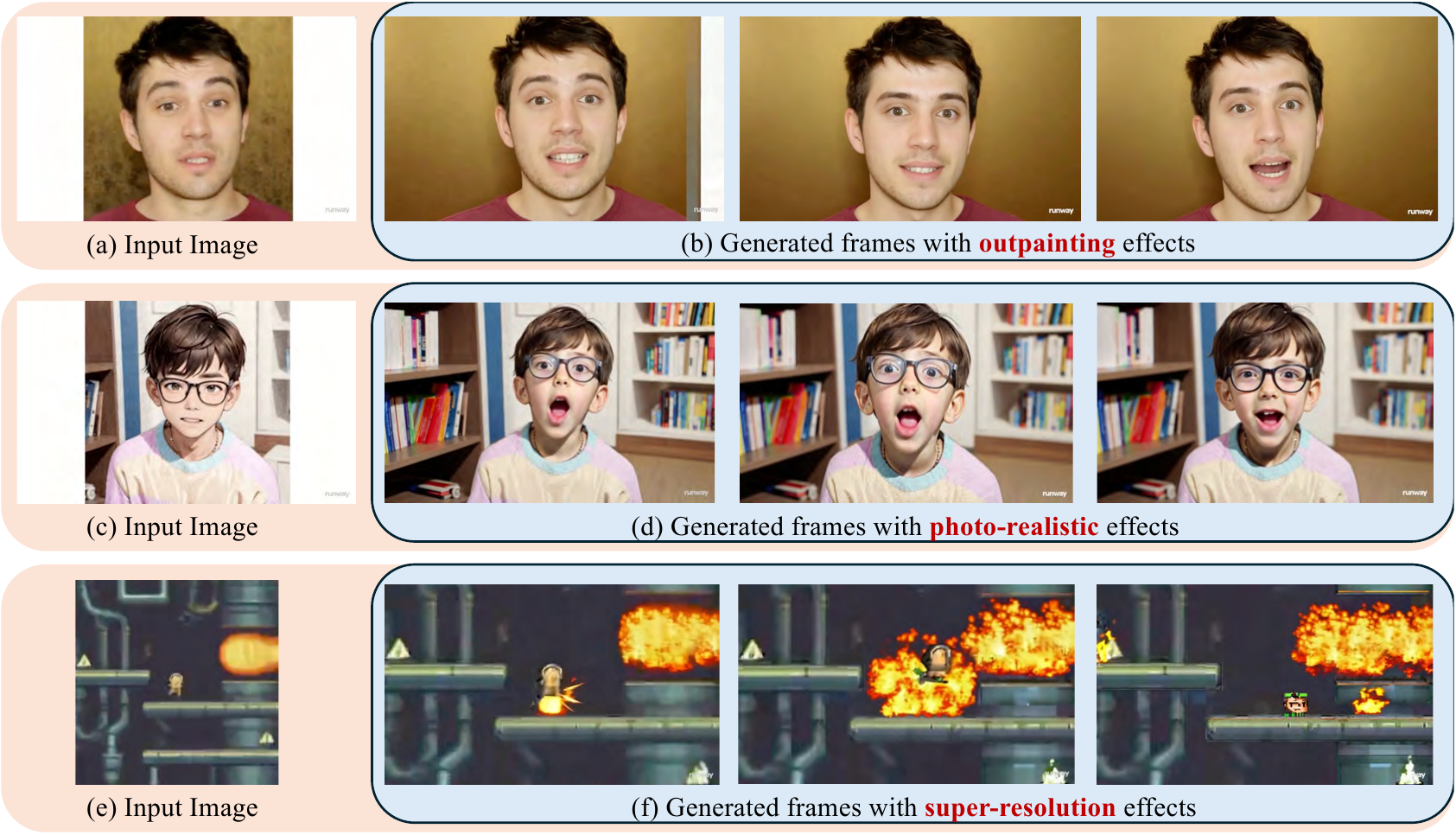}

	\end{overpic}
        \caption{The ability of outpainting, transferring to photo-realistic styles and super-resolution from existing image-to-video generation models, \eg, Gen-3.}
        \label{Fig:Exploration1}
\end{figure}

This section explores the diverse usage scenarios and tasks that advanced video generation models can achieve. From video outpainting and super-resolution functions shown in Figure \ref{Fig:Exploration1} to texture generation and style transfer tasks, these models demonstrate their potential to significantly enhance and expand the applications of video editing and production.

\begin{itemize}

\item \emph{Video Outpainting}: Given an image surrounded by an arbitrary white frame intended for outward expansion, Gen-3's image-to-video (I2V) generation capability can generate a video while simultaneously extending the scene with additional context while maintaining coherence in style and motion. This process leverages the inherent powerful content generation ability of the foundation model.
\item \emph{Video Super-Resolution}: Given a low-resolution image, existing I2V models can enhance the generated video resolution to achieve higher clarity and more details, improving the overall visual quality of the video.
\item \emph{Texture Generation in Videos}: Benefit from the video-to-video function from Gen-3, given a rough draft video or a video of geometric objects following a traditional graphics pipeline, it is possible to generate a video with natural and high-quality textures by providing only the geometric outlines, basic motion information, and the desired texture description (refer to Figure~\ref{Fig:Stylization_v2v1}, \ref{Fig:Stylization_v2v2}). It is generic and can be applied to various objects within a video, contributing to more immersive and lifelike visual effects.

\item \emph{Style Transfer to Photo-Realistic Videos}: Converting arbitrary visual styles (\emph{e.g.}, cartoon, sketch, and 3D model) into photorealistic representations enhances the versatility of video models, allowing them to adapt to different stylistic requirements.

\item \emph{Arbitrary Video Editing via Text Instructions}: Given a (generated) video, this task targets to modify any local or global video components, allowing users to add, delete, change, enhance, or transform video contents by simply providing descriptive text input. It enables a broad range of video editing tasks, including altering scenes, modifying objects, changing colors, adding effects, or even adjusting video timing, without requiring manual intervention through traditional video editing software. 
\end{itemize}

\begin{figure}[!ht]
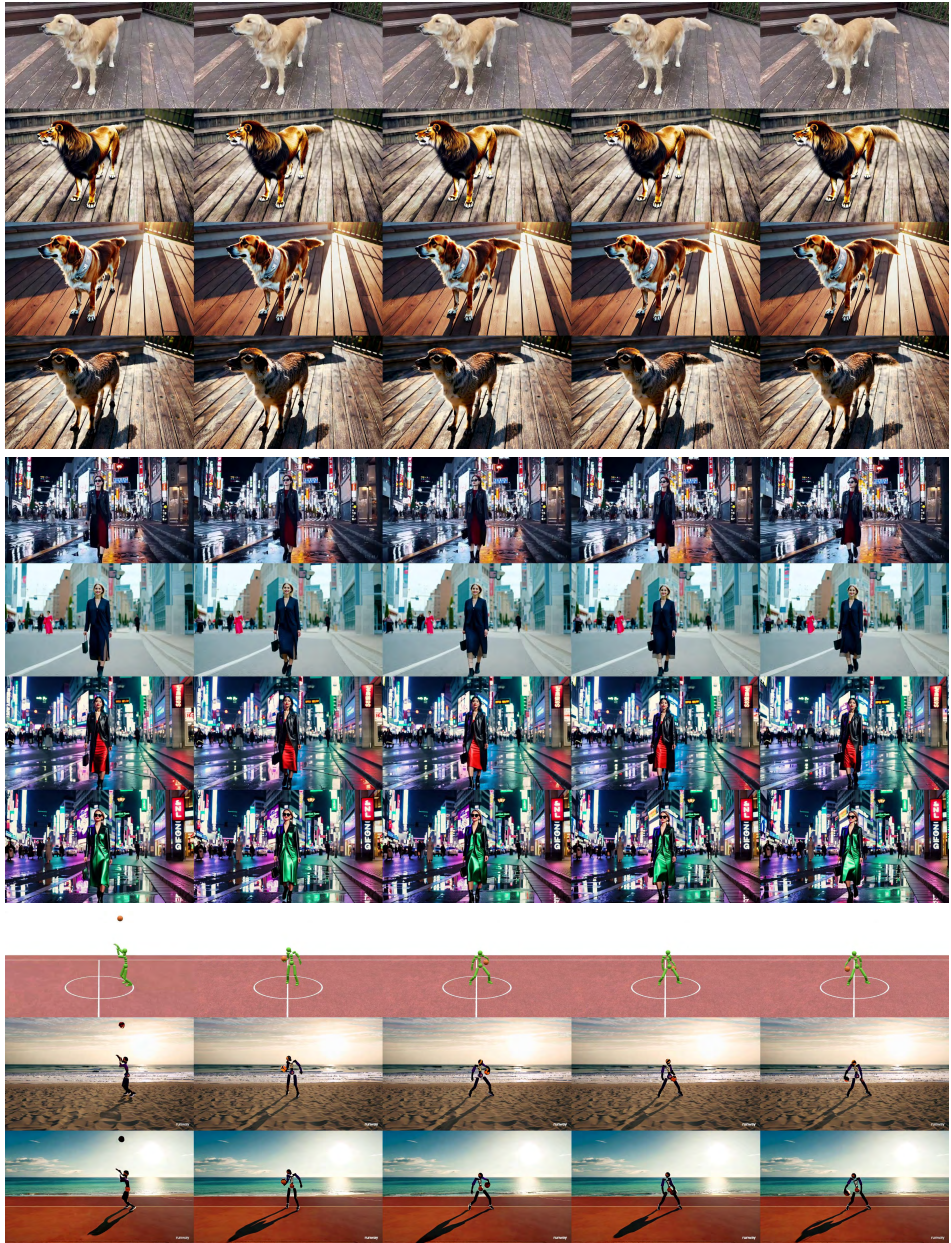

\vspace{-0.3cm}
	\centering
	\small
        \begin{overpic}[width=0.9\linewidth]{Figs/Sec5/00669_00712_00715.pdf}
	\end{overpic}
    \vspace{-0.3cm}
        \caption{\emph{Evaluation on video editing.} Prompt: (V2V) For each row, 1) the input video, 2) Prompt: "change this dog into a lion" [success], 3) Prompt: "add a sunglass on this dog" [fail], 4) Prompt: "remove this dog" [fail], 5) the input video, 6) Prompt: "add a man meets with the woman in this video, and then hug together" [fail], 7) Prompt: "change a camera view to see this video" [fail], 8) Prompt: "change the cloth of the person" [success but the background is changed], 9) the input video, 10) Prompt: "Kobe is playing basketball at the beach wearing the number 26 jersey t-shirt", 11) Prompt: "Kobe is playing basketball at the long beach wearing the number 18 jersey t-shirt."}
        \label{Fig:Exploration2}
        \vspace{-0.3cm}
\end{figure}

This exploration highlights the potential for these models to expand beyond conventional applications, proving their adaptability and effectiveness in more advanced video editing and enhancement tasks. 
However, as shown in Figure \ref{Fig:Exploration2}, arbitrary text-guided video editing still faces limitations. The current V2V models can only perform stylization based on injected text, influenced by the text injection method and how this functionality is trained (\emph{e.g.}, possibly conditioned on the depth maps). 

Recently, Movie Gen~\cite{meta2024moviegen} has shown its ability for precise instruction-based video editing based on the pre-trained foundation text-to-video generation model. It can except that as text control capabilities become more refined, the text will enable arbitrary local to global video editing in the near future, such as adding, removing, or replacing objects within a video using text solely. Additionally, the scope would be expanded to video perception tasks, such as video depth estimation, pose estimation, segmentation, and tracking in videos.

\clearpage

\section{Open-source v.s Closed-source SORA-like models}
\label{sec:opensource}

\begin{figure}[!ht]
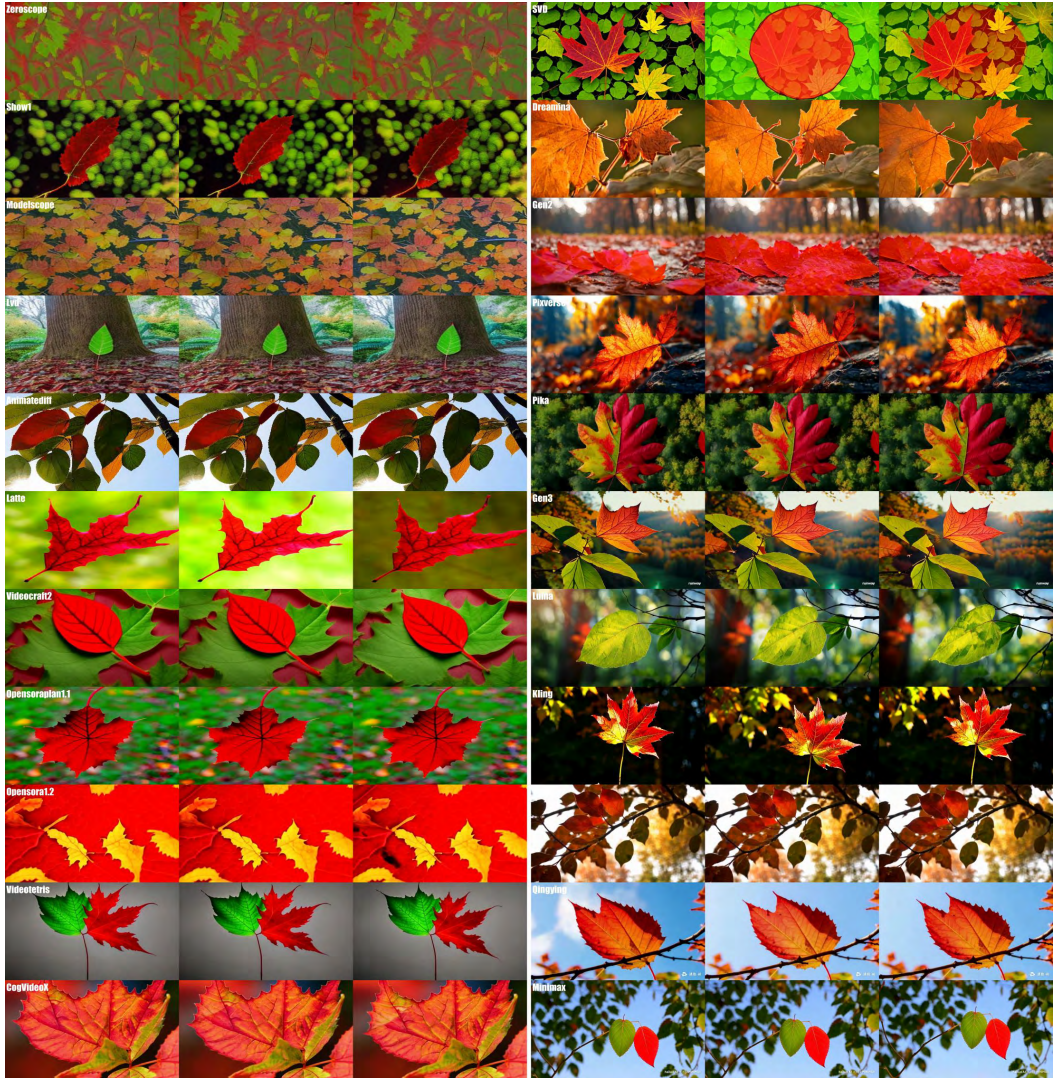

	\centering
	\small
        \begin{overpic}[width=1.\linewidth]{Figs/Sec6/00659.pdf}

	\end{overpic}
        \caption{Comparisons of representative open-source models with closed-source models on the \emph{dynamic attribute binding problem where the attributes change with time}. Prompt: (T2V-659) "A timelapse of a leaf transitioning from green to bright red as autumn progresses." We have the following observations: i). All models still have deficiencies in text semantic comprehension and temporal dynamics change, where they do not change the process from a green to red leaf; some models are spatially showing both green and red leaves (\emph{e.g.}, Videotetris, Gen-3, and MiniMax); ii). Closed-source models are generally better than open-source models regarding visual quality and aesthetics; iii). Not all models have a large change in motion without "time-lapse" photography.}
        \label{Fig:openclose_1}
\end{figure}

\begin{figure}[!ht]
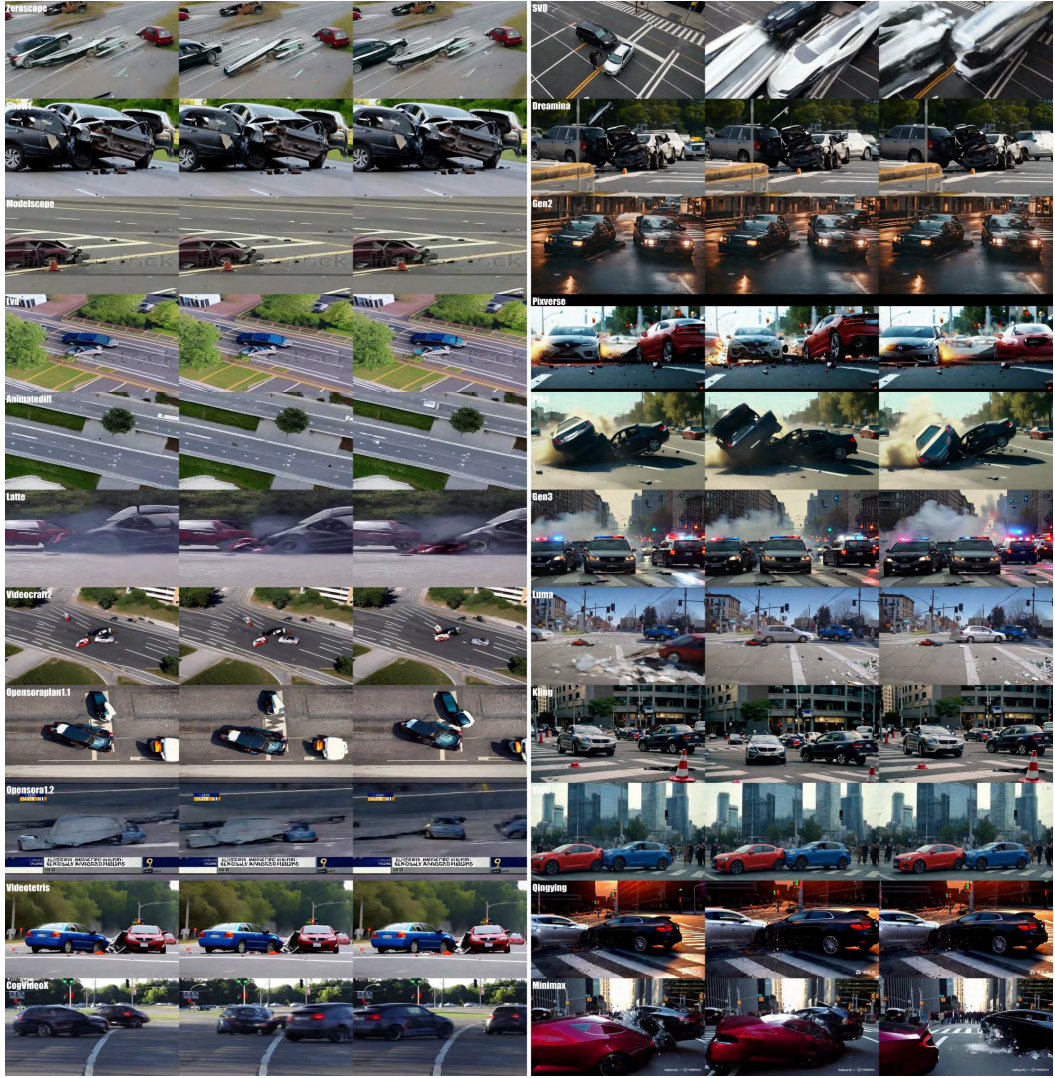

	\centering
	\small
        \begin{overpic}[width=1.\linewidth]{Figs/Sec6/00661.pdf}

	\end{overpic}
        \caption{Comparisons of representative open-source models with closed-source models on the \emph{causal phenomenon}. Prompt: (T2V-661) "Two cars collide at an intersection." Understanding and generating dynamic interactions with reasonable physical and social interactions are important. We can observe: i). the ability to generate spatial interaction relationships is enhanced; ii). the occurrence of motion collapse (\emph{e.g.}, from SVD) in parts with large motion is alleviated; iii). However, there still exists a difficulty in simulating realistic, physically compliant collision visualization when collision motion occurs in the interaction.}
        \label{Fig:openclose_2}
\end{figure}

\begin{figure}[!ht]
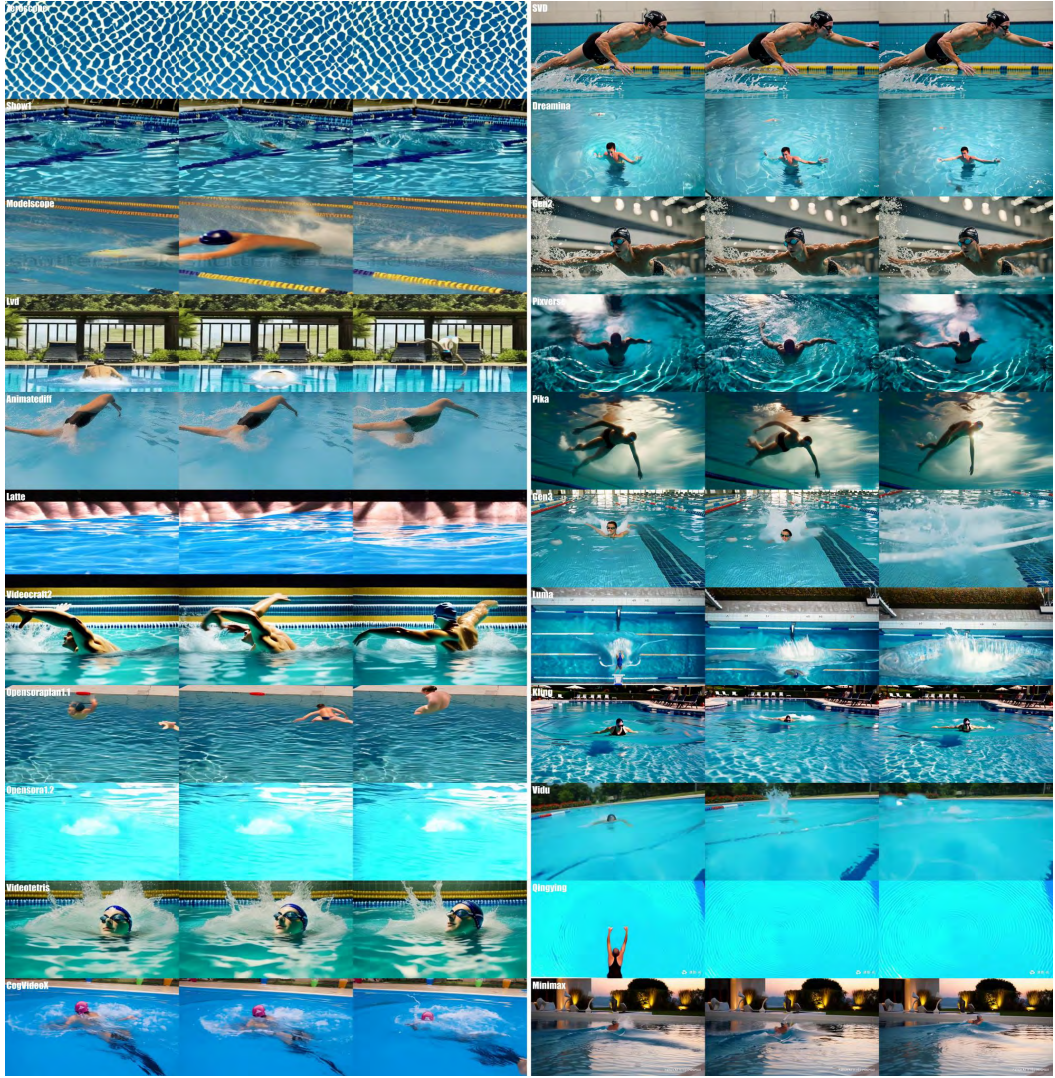

	\centering
	\small
        \begin{overpic}[width=1.\linewidth]{Figs/Sec6/00662.pdf}

	\end{overpic}
        \caption{Comparisons of representative open-source models with closed-source models on the \emph{human-scene interaction and complex human motion problem}. Prompt: (T2V-662) "A swimmer dives into a pool, creating ripples." Due to the visibility of people underwater being obscured by water, the generation of swimming movement tends to be more complex. i). With the enhancement of modeling capabilities, the appearance, aesthetics, and biology of complex motions are enhanced; ii). At the same time, the swimming process of the physical simulation of the splash generated by the movement of the limbs has been strengthened; iii). However, the complex motion generation(\emph{e.g.}, diving) is still challenging to all models.}
        \label{Fig:openclose_3}
\end{figure}

\begin{figure}[!ht]
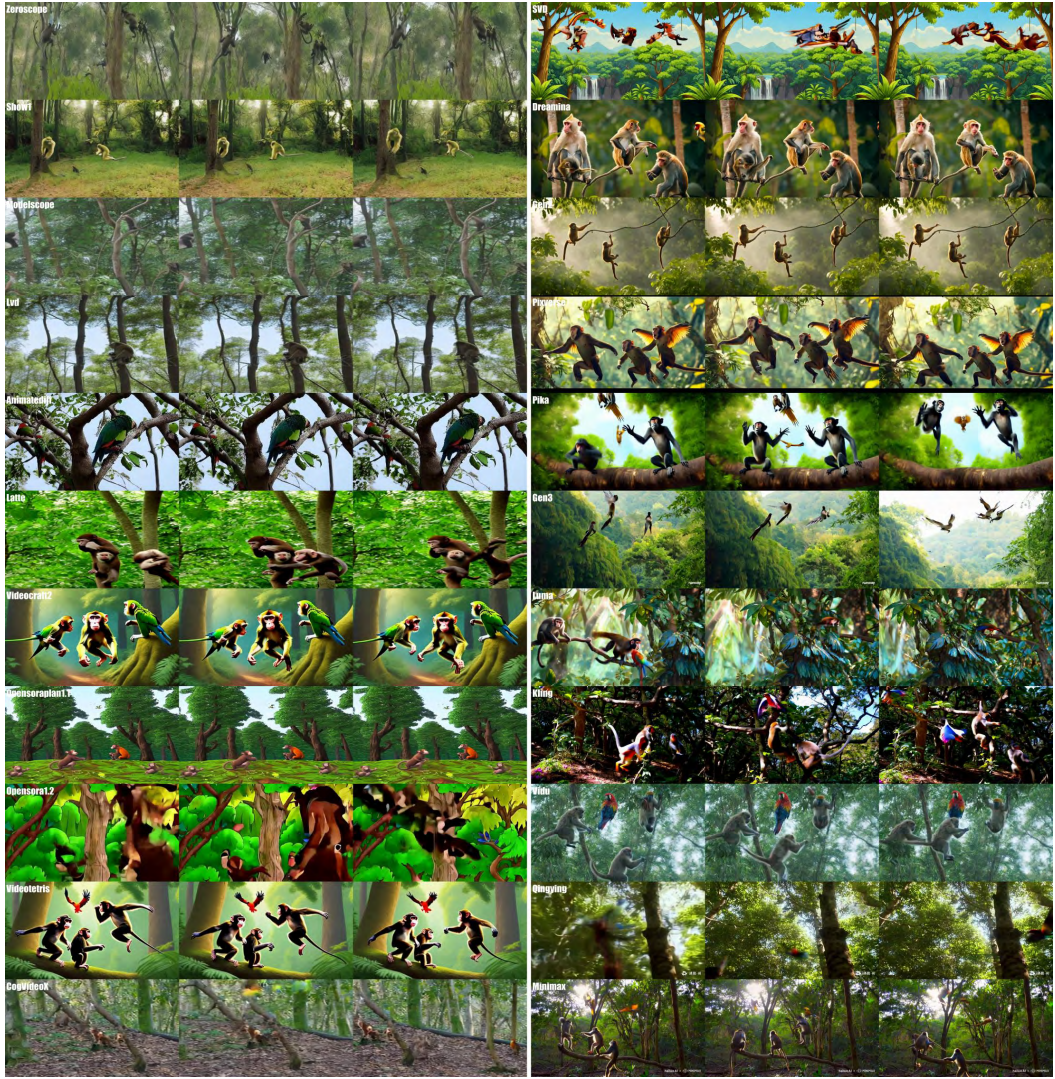

	\centering
	\small
        \begin{overpic}[width=1.\linewidth]{Figs/Sec6/00664.pdf}

	\end{overpic}
        \caption{Comparisons of representative open-source models with closed-source models on the \emph{multi-class multi-object numeracy problem}. Prompt: (T2V-664) "Three monkeys are jumping in the forest, while two parrots are flying among the trees." MiniMax generates quantities, objects, and motions with an accuracy closest to the input's textual description. Other models suffer from conceptual confusion of spatial objects and insensitivity to quantity words.}
        \label{Fig:openclose_4}
\end{figure}

As shown in Figure \ref{Fig:openclose_1}, \ref{Fig:openclose_2}, \ref{Fig:openclose_3}, \ref{Fig:openclose_4}, we provide compared visualization among numerous open-source and closed-source models for intuitive observation. The compared models include a series of UNet-based and DiT-based models (Zeroscope~\cite{khachatryan2023text2video_zero}, Show1~\cite{zhang2023show1}, Modelscope~\cite{wang2023modelscope}, LVD~\cite{lian2023llm}, Animatediff~\cite{guo2023animatediff}, Latte~\cite{ma2024latte_videogen}, VideoCraft2~\cite{chen2024videocrafter2}, Open-Sora-Plan 1.2~\cite{yuan_opensora}, Videotetris~\cite{tian2024videotetris}, CogVideoX~\cite{hong2022cogvideo,yang2024cogvideox}, SVD~\cite{blattmann2023svd}), as well as closed-source video generation products (Dreamina~\cite{dreamina}, Gen-2~\cite{runway2024gen2}, Pixverse~\cite{pixverse}, Pika~\cite{pika2024pika}, Gen-3~\cite{runway2024gen3}, Luma~\cite{luma2024dm}, Kling~\cite{kuaishou2024kling}, Vidu~\cite{bao2024vidu}, Qingying~\cite{zhipu2024qingying}, and MiniMax~\cite{minimax2024hailuo}.
These prompts and open-source videos are from T2V-CompBench~\cite{sun2024t2v_bench}. We select several representative prompts from the dynamic attribute binding, object interactions, and numeracy for multi-class multi-object generation. 

Closed-source models generally outperform open-source models in several key aspects due to the larger model size, more diverse and higher-quality training data, and advanced fine-tuning techniques. They often exhibit superior video resolution, visual quality, smooth and dynamic motion, and more accurate visual-text alignment, enabling more coherent and realistic video generation across various scenarios. Furthermore, closed-source models demonstrate enhanced generalization capabilities, handling complex prompts and nuanced details with greater precision. However, all models still face many common challenges, such as accurately generating dynamic attribute binding content (Figure~\ref{Fig:openclose_1}), handling complex motions with corresponding physical reactions (Figure~\ref{Fig:openclose_2}, \ref{Fig:openclose_3}), and generating multi-class, multi-object counting scenarios (Figure~\ref{Fig:openclose_4}).

\clearpage

\section{Challenges and Future Work}

\label{sec:challenge}

\begin{figure}[h]
	\centering
	\small
        \begin{overpic}[width=1.\linewidth]{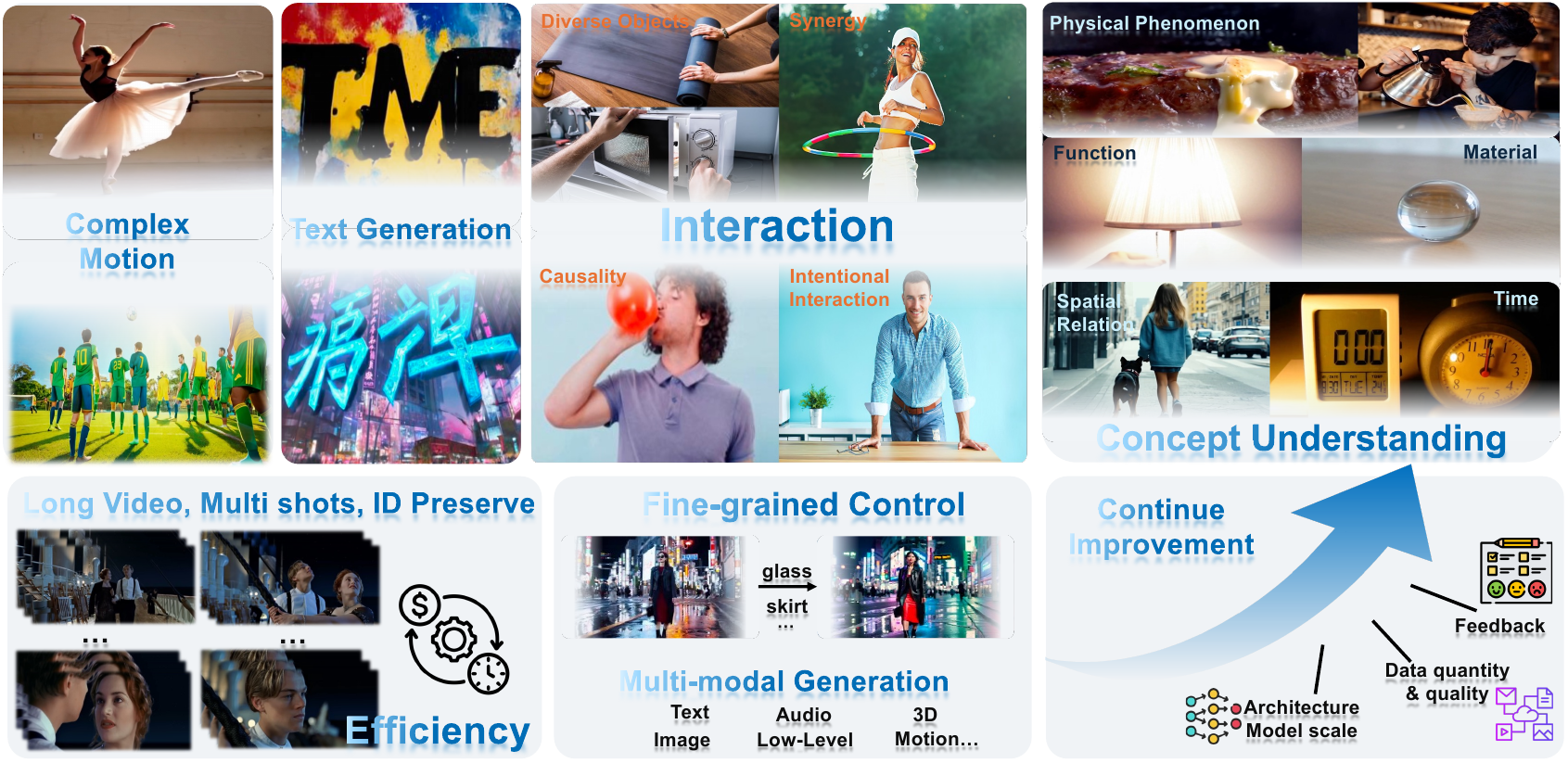}

	\end{overpic}
        \caption{Overview of Section~\ref{sec:challenge}. We discuss existing pertinent challenges in the current SORA-like models, as well as future directions.}
        \label{Fig:Challenge_teaser}
\end{figure}

As shown in Figure \ref{Fig:Challenge_teaser}, we summarize the challenges that existing models still face and analyze the underlying problems that need to be addressed, inspiring future research. 

\subsection{Complex Motion}
\begin{figure}[!ht]
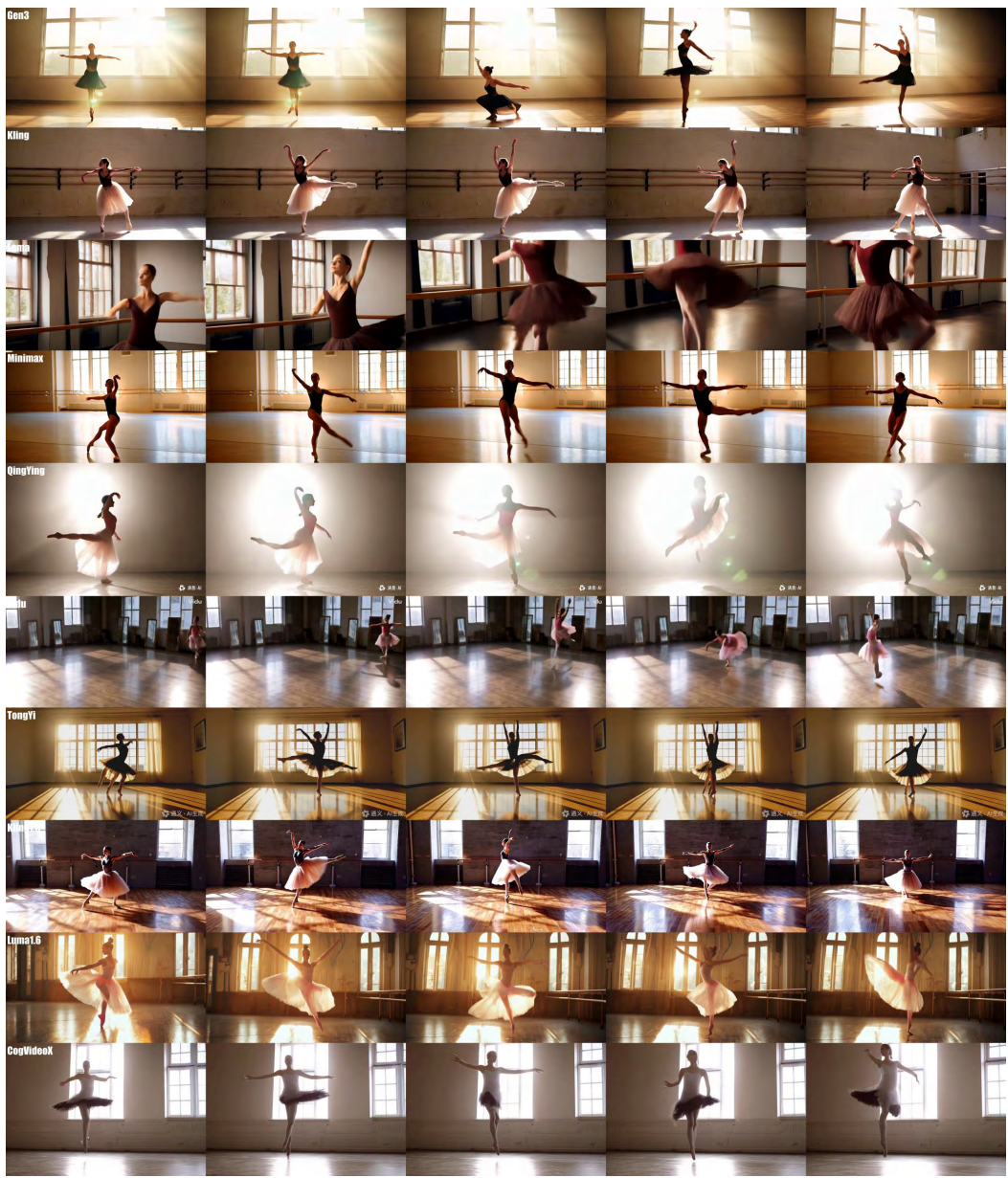

	\centering
	\small
        \begin{overpic}[width=1.\linewidth]{Figs/Sec7/00261.pdf}

	\end{overpic}
        \caption{\emph{Complex motion, \eg , ballet.} Prompt: (T2V-261) "A professional ballet dancer performs a grand jeté across a sunlit studio, with the camera capturing the grace and fluidity of her movements in mid-air." Most models inevitably generated unnatural motions and incorrect body topologies (such as an extra leg). Kling 1.5 and MiniMax perform better.}
        \label{Fig:ComplexMotion_1}
\end{figure}

\begin{figure}[!ht]
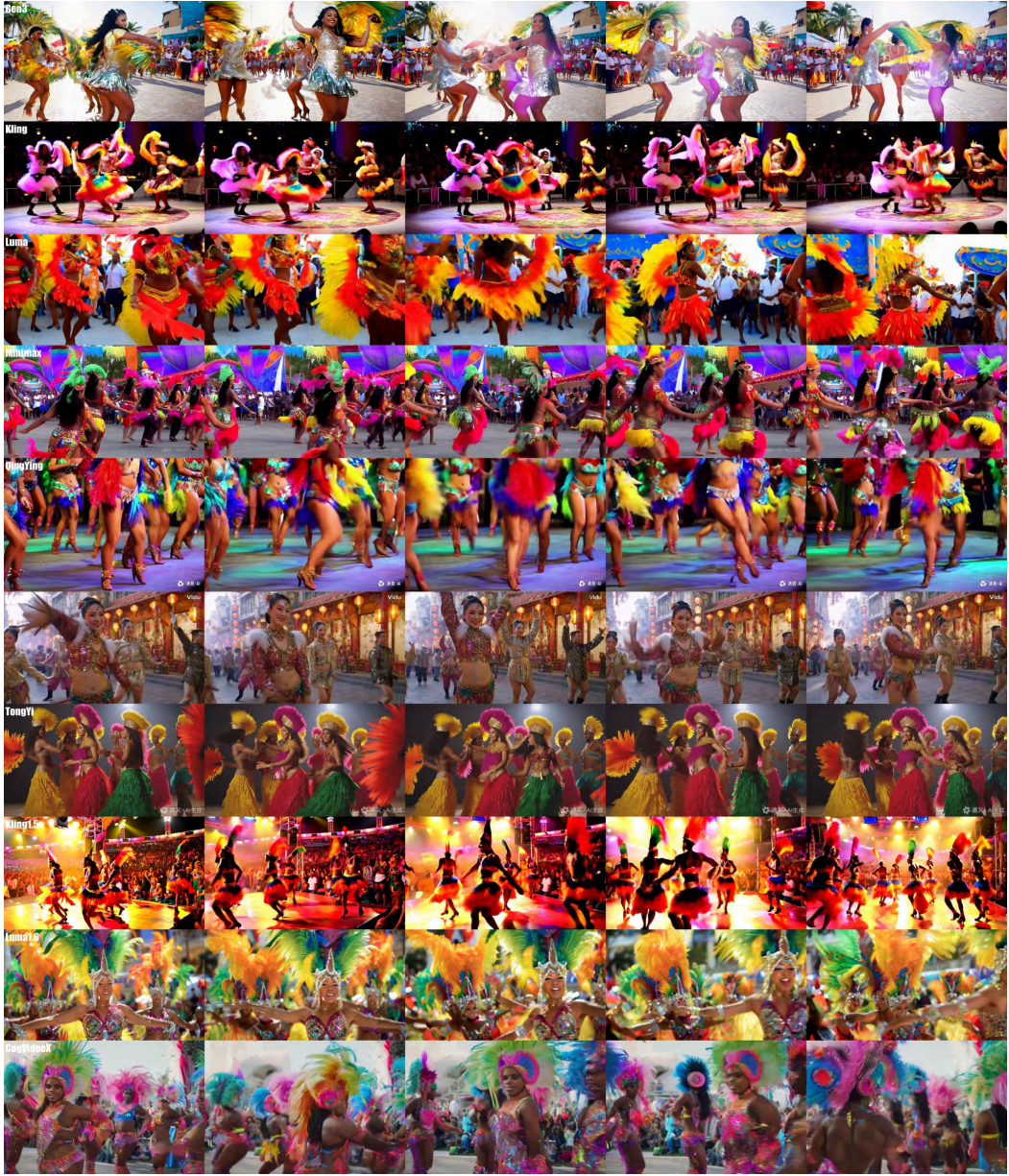

	\centering
	\small
        \begin{overpic}[width=1.\linewidth]{Figs/Sec7/00218.pdf}

	\end{overpic}
        \caption{\emph{Complex motion, multi-instances.} Prompt: (T2V-218) "A group of Caribbean dancers performing at a carnival, wearing colorful costumes with feathers and sequins. The camera follows their energetic dance moves and the lively music." The generation of complex motions faces greater challenges due to issues such as occlusion with multi-instance scenarios.}
        \label{Fig:ComplexMotion_2}
\end{figure}

\textbf{Challenges.} Existing models still face challenges when it comes to generating complex motions, especially in complex or rare human motions with high movements, heavy occlusions, and complex interactions, like ballet (Figure \ref{Fig:ComplexMotion_1}) or diving, and multi-instance activities (Figure \ref{Fig:ComplexMotion_2}). On the one hand, these motions involve joint rotations and interactions that significantly deviate from common human motions. The long-tail distribution may exist in training data, even though the post-training strategies with high-quality data, making hard and rare motions still challenging. On the other hand, the challenge may lie in using the text to control the fine-grained generation of such complex motions, which inherently leads to ambiguities from either spatial or temporal dimensions, \eg, it is hard to concisely describe some continuous and complex motions through text solely.

\textbf{Future Work.} Leveraging low-level signals provided by perception models or user inputs, such as human poses, masks, and trajectories, and audio as additional control conditions, might be a solution, which has already shown initial effects in certain methods \cite{mou2024revideo, hu2024animate_anyone,ma2024emoji,tian2024emo}. However, these pixel-level signals still face projection ambiguity, \eg, due to the lack of a dimension, pixel-level signals are hard to explicitly represent rotations in 3D space. Exploring the type of signals (including 3D) that drive motion generation and how they are integrated remains worthy of further investigation. In addition, specific post-training strategies and motion-aware personalized modules would be introduced to enhance the motion quality, stylization, customization, and expressiveness. Automatic motion adjustment and distillation in video data selection and training could make the training process more efficient.

\clearpage

\subsection{Concept Understanding}
\begin{figure}[!ht]
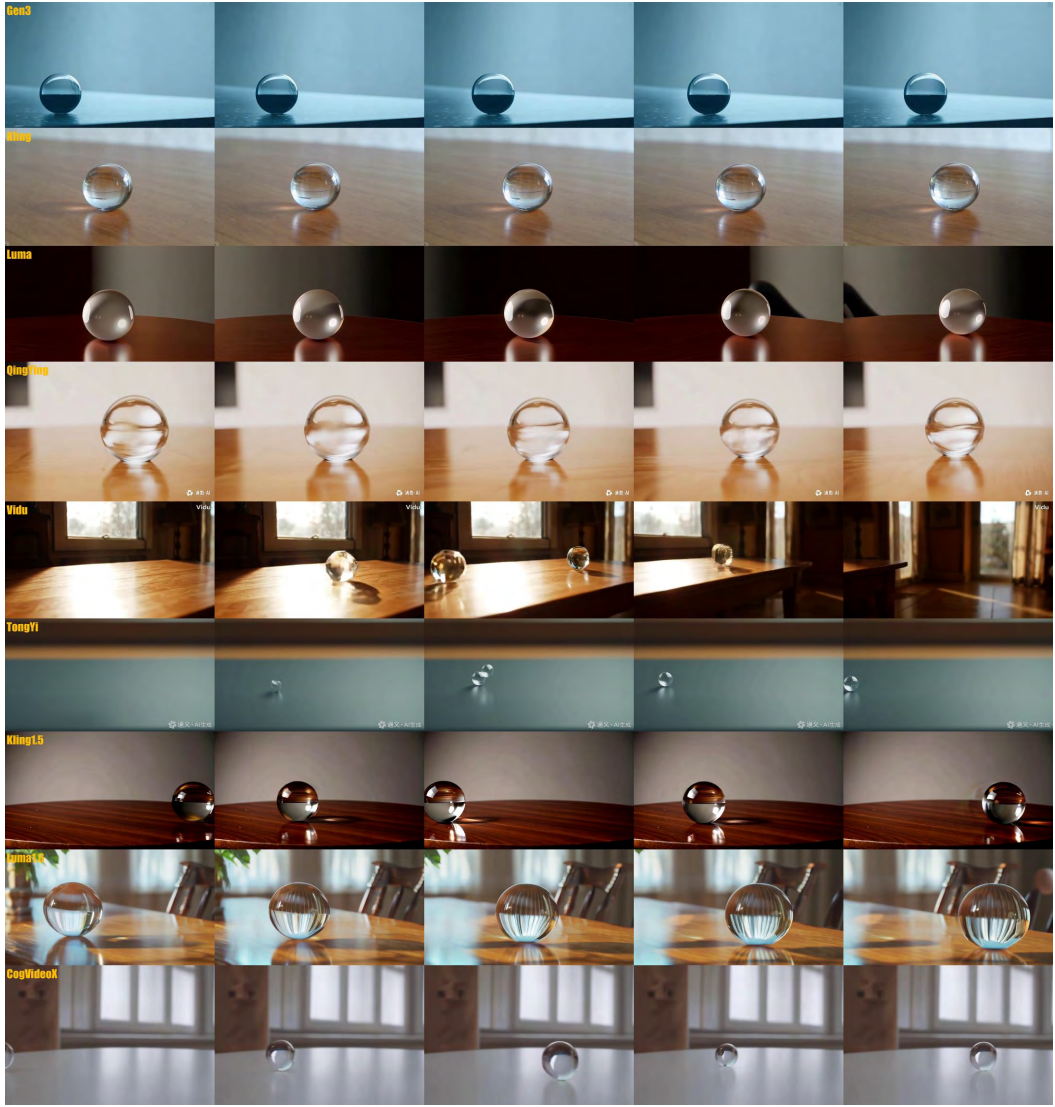

	\centering
	\small
        \begin{overpic}[width=1.\linewidth]{Figs/Sec7/00102.pdf}

	\end{overpic}
        \caption{\emph{Object attributes, the material.} Prompt: (T2V-102) "Static camera, a glass ball rolls on a smooth tabletop." Currently, all models struggle to generate motions that comply with the physical laws corresponding to the given material.}
        \label{Fig:material_1}
\end{figure}

\begin{figure}[!ht]
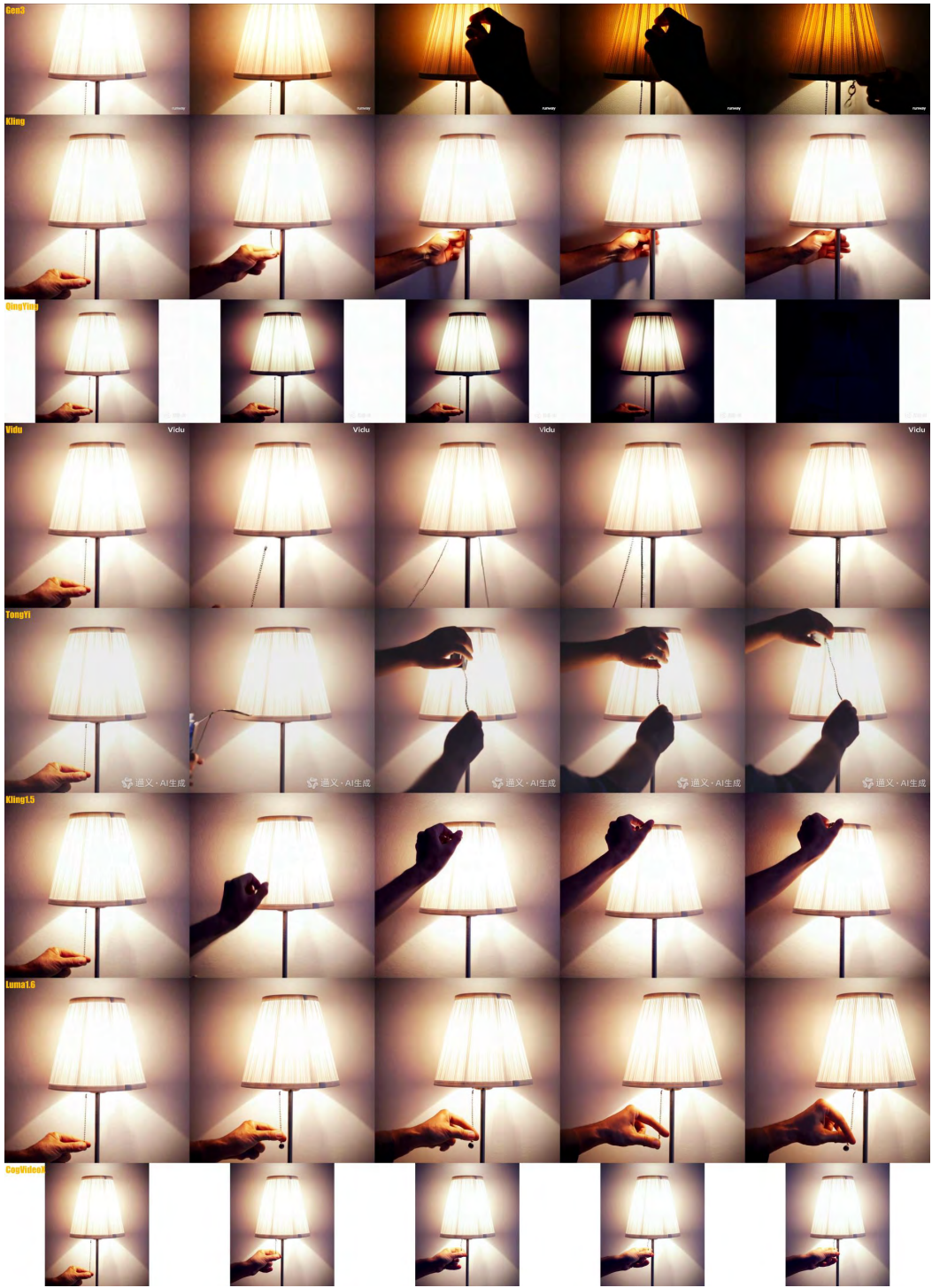

	\centering
	\small
        \begin{overpic}[width=1.\linewidth]{Figs/Sec7/00005.pdf}

	\end{overpic}
        \caption{\emph{Object attributes, the function and reasoning abilities.} Prompt: (I2V-5) "The camera remains still, a hand reaching out to pull the chain of a lampshade, turning the lamp off." Except for Qingying, none of the other models generat the correct action of turning off the lamp or the lighting changes caused by the action, indicating the misunderstanding of the functions of objects.}
        \label{Fig:function}
\end{figure}

\begin{figure}[!ht]
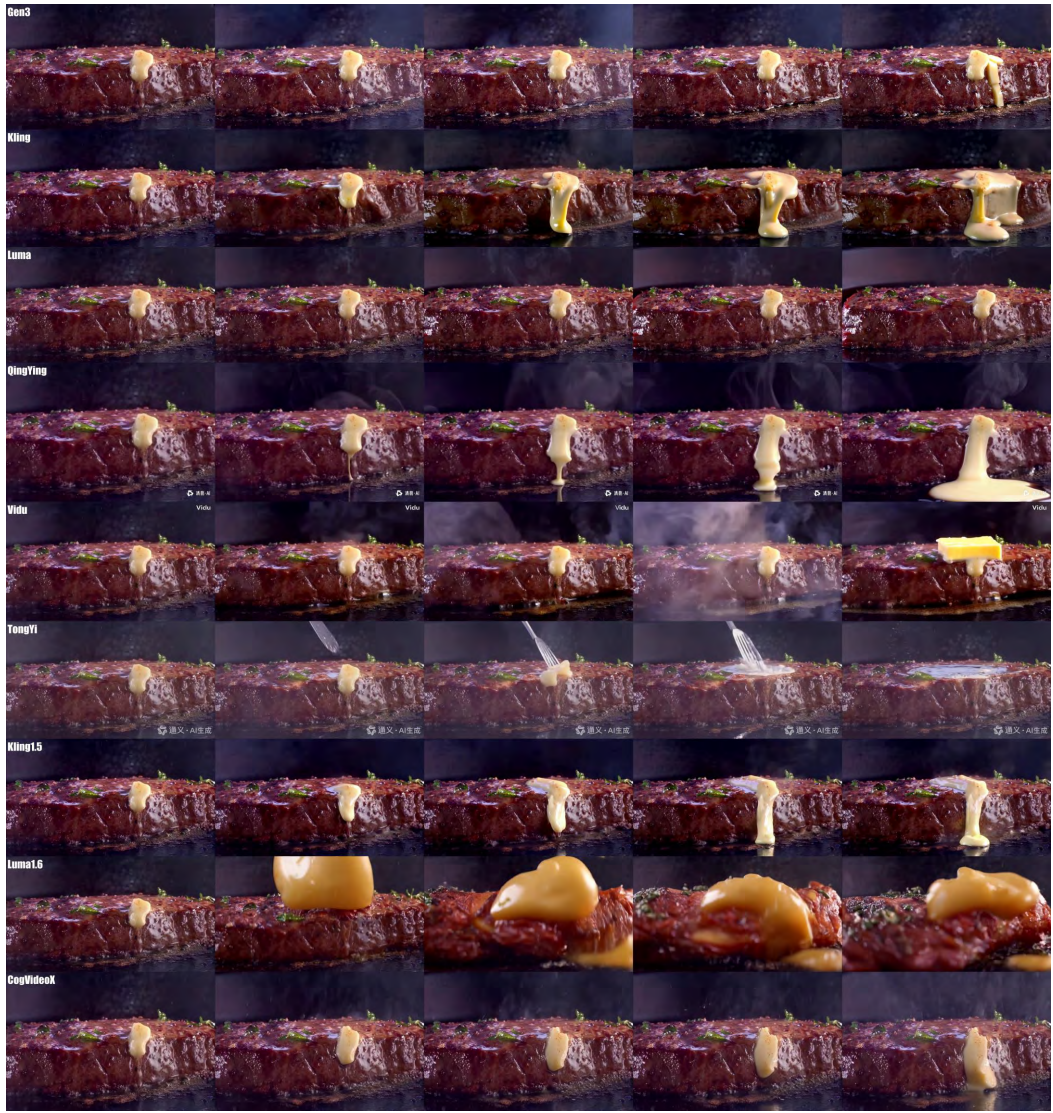

	\centering
	\small
        \begin{overpic}[width=1.\linewidth]{Figs/Sec7/00188.pdf}

	\end{overpic}
        \caption{\emph{Physical phenomenon, heat.} Prompt: (I2V-188) "A close-up static shot shows a steak being cooked on a hot surface as a pat of butter melts on top of it. The butter melts as the steak continues to cook." Some models can visually simulate the melting of butter caused by heating, but the physical rules in the scene are unreasonable (the volume of the butter does not change after melting, or even increases).}
        \label{Fig:heating}
\end{figure}

\begin{figure}[!ht]
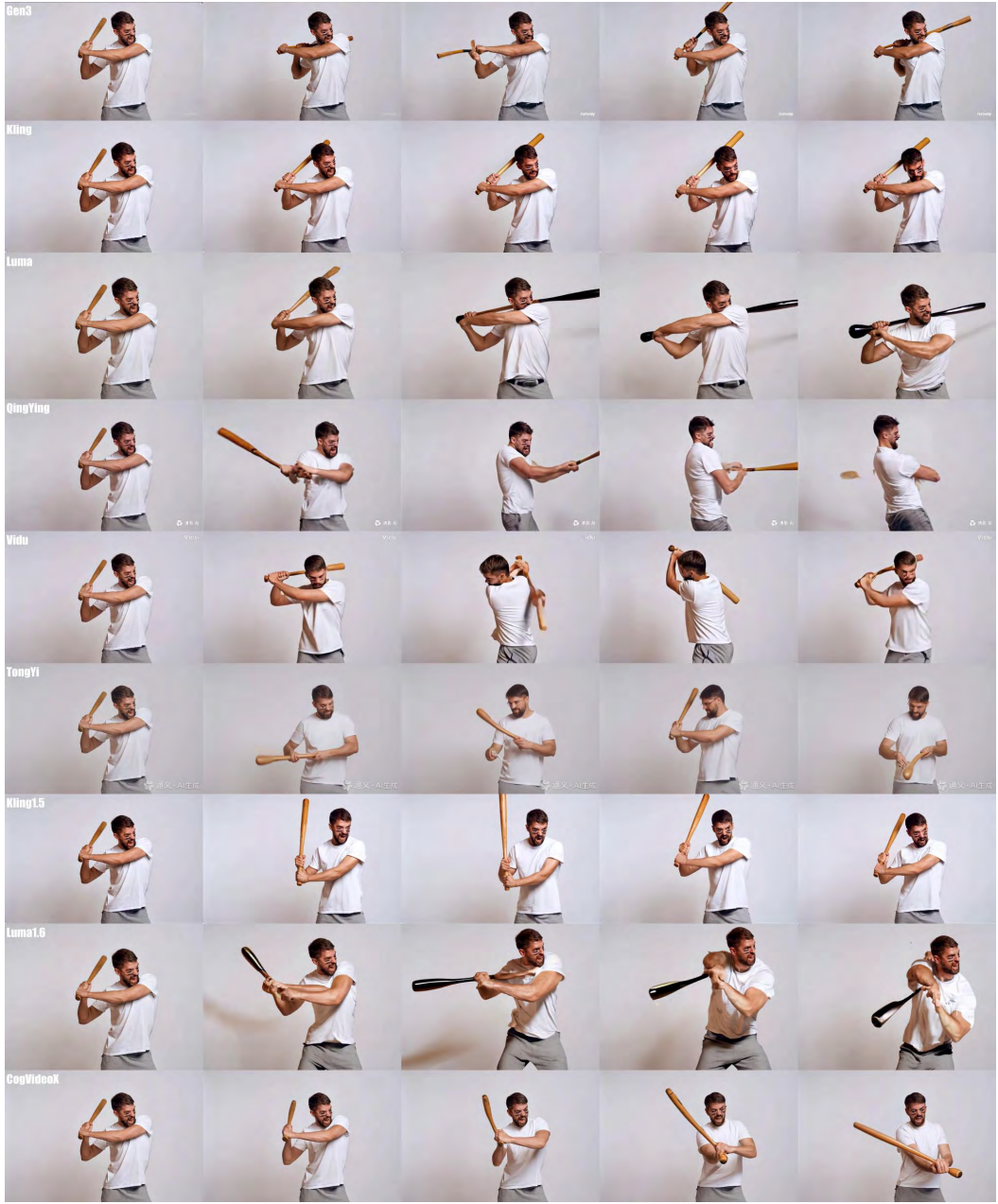

	\centering
	\small
        \begin{overpic}[width=1.\linewidth]{Figs/Sec7/00637.pdf}

	\end{overpic}
        \caption{\emph{Dynamics, slow motion.} Prompt: (I2V-637) "This person swings the baseball bat \textbf{slowly} with both hands." Compared with Figure \ref{Fig:fast_motion}, the goal is to verify whether the model can handle the root cause of motion changes, such as the degree to which the person is exerting force in this case. We recommend watching videos on the website to observe the results.}
        \label{Fig:slow_motion}
\end{figure}

\begin{figure}[!ht]
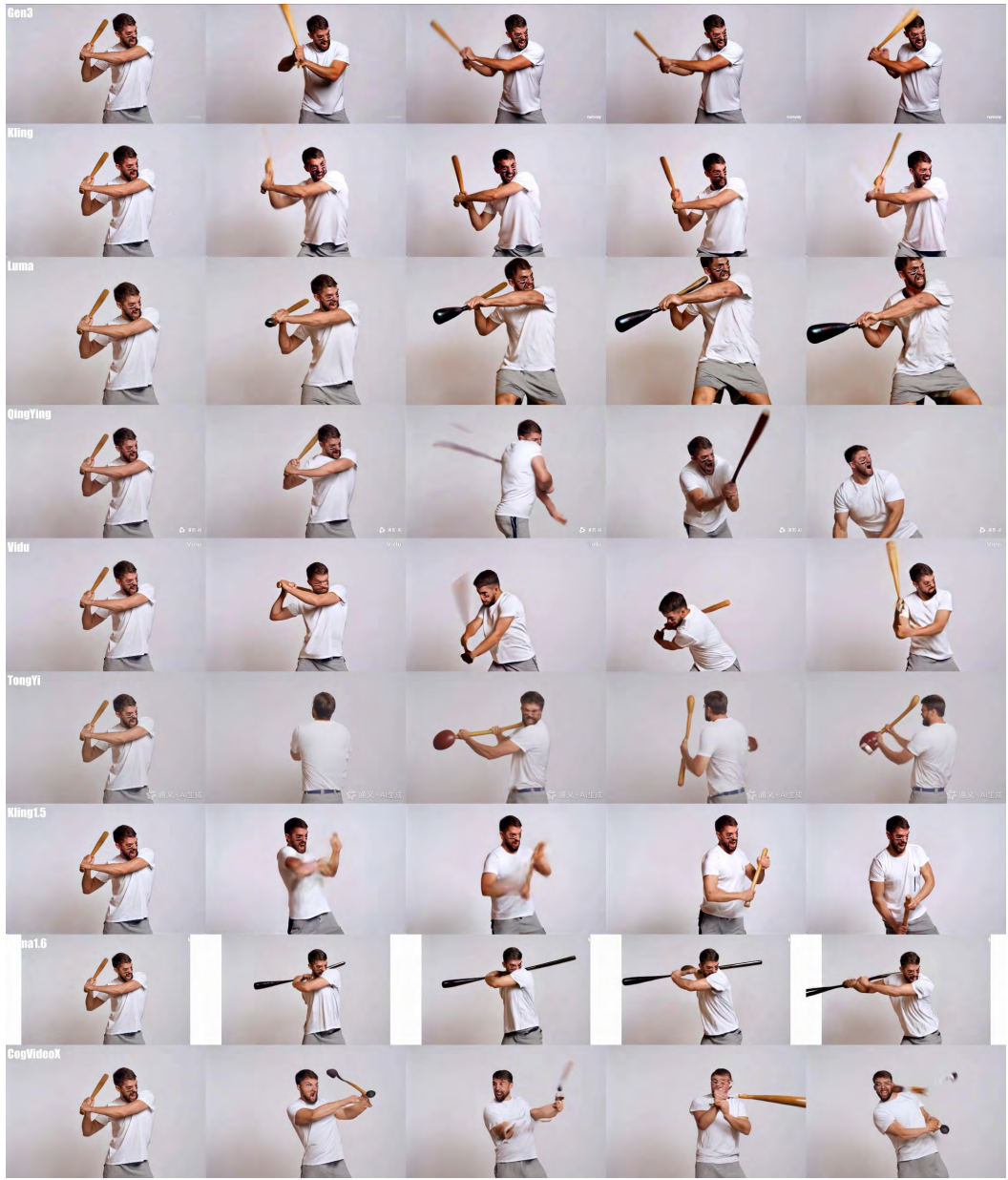

	\centering
	\small
        \begin{overpic}[width=1.\linewidth]{Figs/Sec7/00638.pdf}

	\end{overpic}
        \caption{\emph{Dynamics, fast motion.} Prompt: (I2V-638) "This person \textbf{vigorously} swings a baseball bat with both hands." Compared with Figure \ref{Fig:slow_motion}. Existing models are hard to control the degree of motions and will fail at maintaining the integrity of the character's appearance. }
        \label{Fig:fast_motion}
\end{figure}

\begin{figure}[!ht]
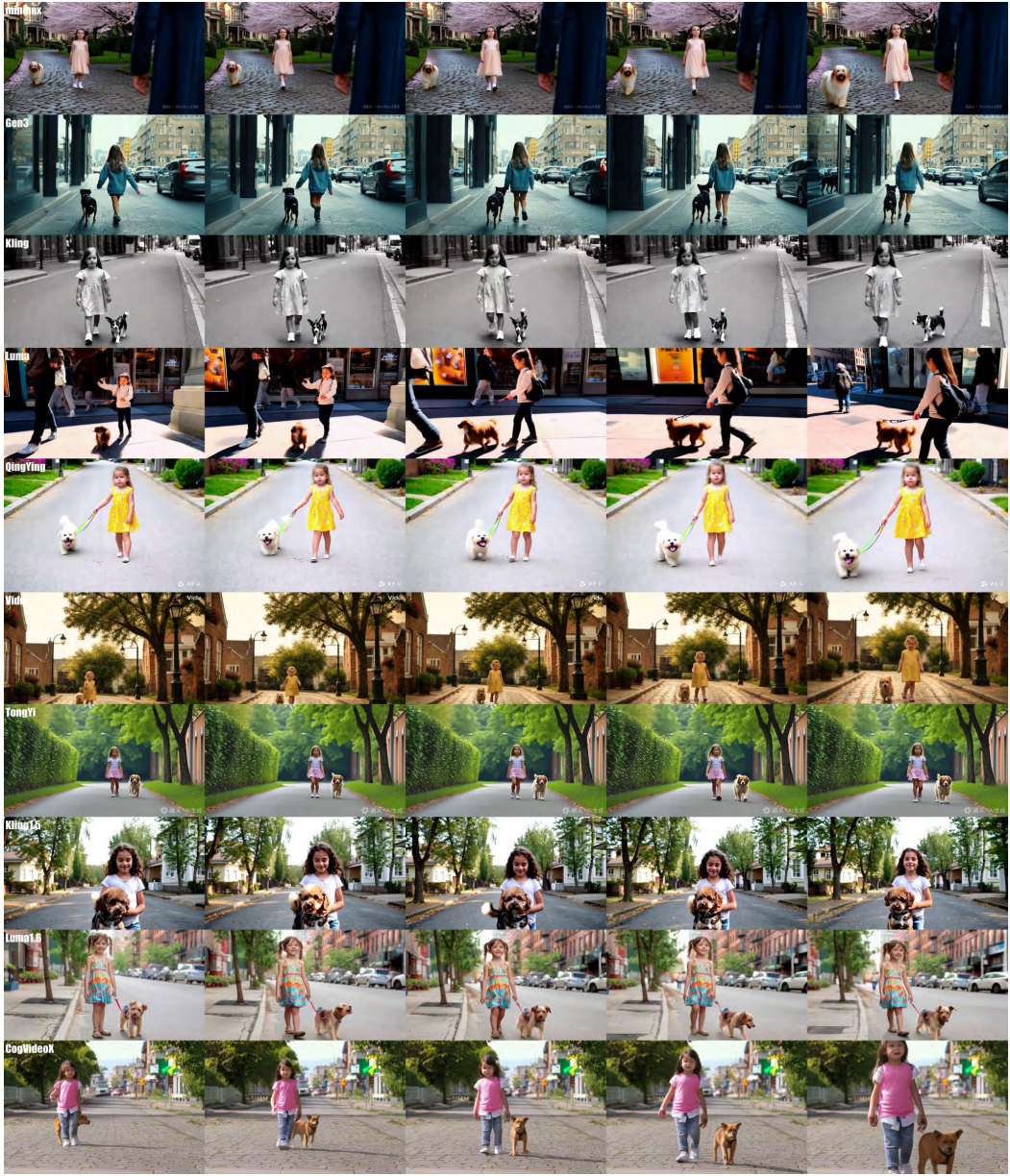

	\centering
	\small
        \begin{overpic}[width=1.\linewidth]{Figs/Sec7/00131.pdf}

	\end{overpic}
        \caption{\emph{Concept, spatial relations.} Prompt: (T2V-131) "Static camera, a little girl is walking on the street with a small dog on her left." Due to the inaccuracy of large vision-language model in describing spatial relationships in videos, current generative models also have a high likelihood of errors in generating spatial relationships.}
        \label{Fig:Concept_spatial}
\end{figure}

\begin{figure}[!ht]
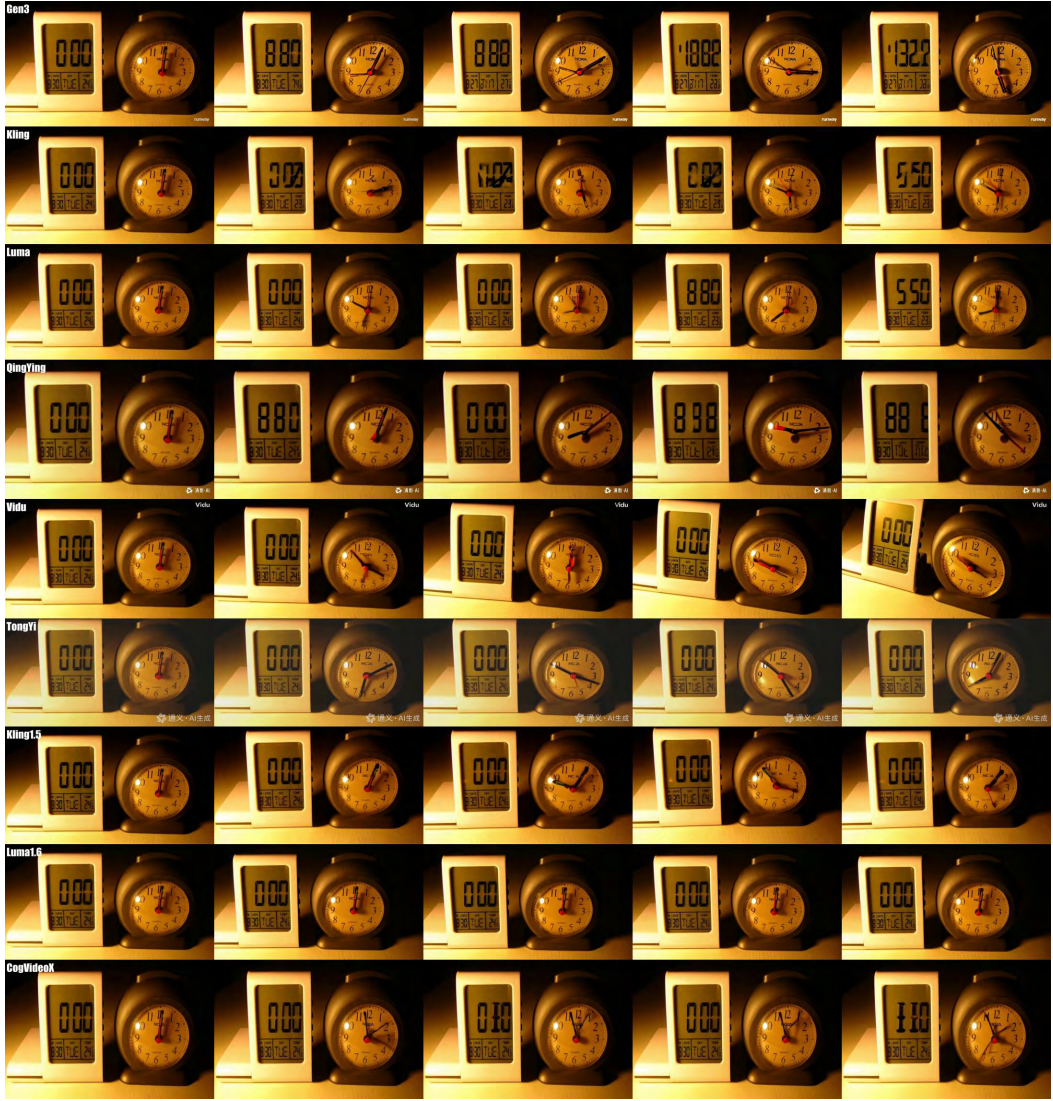

	\centering
	\small
        \begin{overpic}[width=1.\linewidth]{Figs/Sec7/00451.pdf}

	\end{overpic}
        \caption{\emph{Concept, time awareness.} Prompt: (I2V-451) "The video shows a static camera with a digital alarm clock and an analog clock placed side by side. The time on the digital clock moves rapidly, while the time on the analog clock moves synchronously. The red second hand rotates counterclockwise." Perceiving time is crucial for many scenarios, but due to incomplete temporal annotations, all current models struggle to perceive and generate time accurately.}
        \label{Fig:Concept_time}
\end{figure}

\textbf{Challenges.} when treating video generation models as world simulators, whether they simulate real-world concepts is an essential factor to consider, as it is a key driver of state changes. Mastering various concepts is non-trivial, here, we outline several points to concretize these challenging concepts:

\begin{itemize}[leftmargin=*]

\item  \textbf{Object Attributes}: in generated videos, objects are mainly manifested as two aspects, appearance and state change, which are mainly determined by material and their functions, respectively. The combination of an object’s material and ambient lighting largely determines its visual appearance. Additionally, the material also affects the object’s motion, \eg, the friction of a glass ball and a wooden ball differs, leading to different movements (Figure \ref{Fig:material_1}, \ref{Fig:material_2}) on the same plane. Plus, prior knowledge of objects' functional properties, like affordance \cite{gibson2014ecological, Yang_2023_ICCV}, inherently causes state changes, including their movements and alterations to the environment. For instance, the function of a desk lamp is to provide illumination, and it changes the environmental lighting (Figure \ref{Fig:function}).

\item  \textbf{Physical Phenomenon}: the physical phenomena manifested in video generation models can be summarized into the following aspects: \textbf{i) Physical Property}, which includes temperature, mass, volume, fluid, and so on. These intrinsic attributes partly determine the states in temporal, such as melting due to heating (Figure \ref{Fig:heating}), water overflowing when it exceeds the cup’s capacity, etc; \textbf{ii) Dynamics}, actually, existing methods predominantly focus on kinematics, that is, whether position and velocity appear visually plausible. However, tracing back to what causes or modifies the motion is crucial for capturing dynamics. For instance, when swinging a baseball bat forcefully or gently, the motion will differ, as will the speed of the baseball being hit (Figure \ref{Fig:slow_motion}, \ref{Fig:fast_motion}).

\item  \textbf{Spatial \& Time Concepts}: spatio-temporal characteristics are the nature of video data. Regarding spatial concepts, the simplest one is orientation. Beyond that, models need to handle the boundaries of the scene and complex spatial relations between objects. However, current methods still exhibit confusion with even the basic orientations, \eg, the left or right (Figure \ref{Fig:Concept_spatial}). 
The concept of time can be divided into two levels: \textbf{i)} the generated content aligns with the chronological order, \eg, following the temporal prepositions in instructions (please refer to Figure \ref{Fig:world1}, \ref{Fig:world2}); \textbf{ii)} generating contents according to unit-based time with a given fps (Figure \ref{Fig:Concept_time}), which is particularly important for content with time constraints, such as advertisements.
\end{itemize}

\textbf{Future Work}. To generate content of complex concepts, a perception model that understands real-world laws is a prerequisite. Decoupling the understanding and reasoning of concepts from content generation, the perception model is responsible for understanding and reasoning about concepts, then serves as a condition to guide the generative model for content-aligned generation. Furthermore, regarding the physical phenomena,  whether to use only video data for supervised training in pixel space or to further utilize physical simulation \cite{liu2024physgen} to generate these phenomena remains worth exploring. Plus, generating accurate temporal and spatial concepts requires more fine-grained data annotation, such as temporal localization and spatial relation labeling of the training video data.

\begin{figure}[!ht]
	\centering
	\small
        \begin{overpic}[width=1.\linewidth]{Figs/Sec7/00103.pdf}

	\end{overpic}
        \caption{\emph{Object attributes, the material.} Prompt: (T2V-103) "Static camera, a metal ball rolls on a smooth tabletop." Currently, all models struggle to generate motions that comply with the physical laws corresponding to the given material.}
        \label{Fig:material_2}
\end{figure}

\clearpage

\subsection{Interaction}
\begin{figure}[!ht]
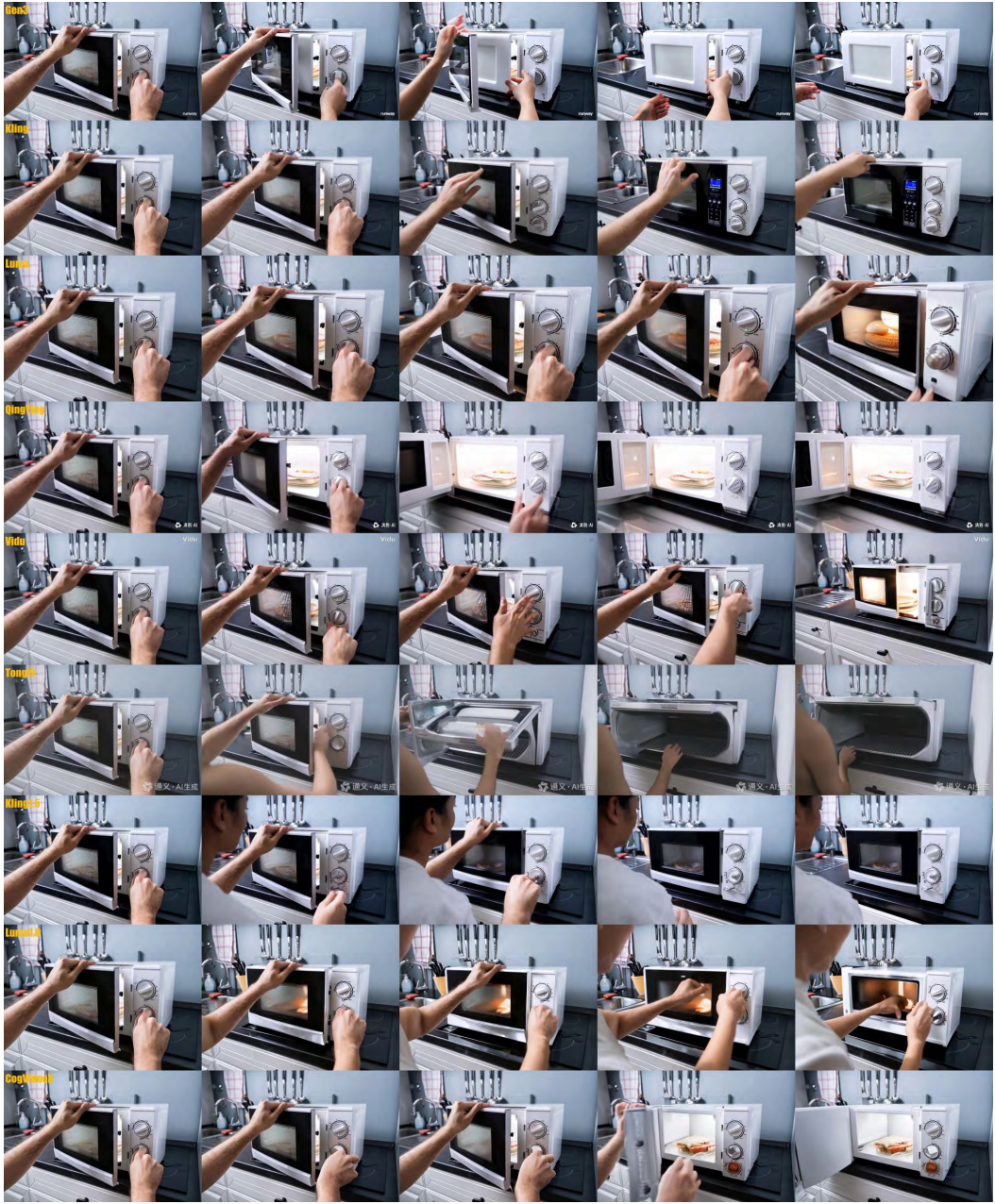

	\centering
	\small
        \begin{overpic}[width=1.\linewidth]{Figs/Sec7/00068.pdf}

	\end{overpic}
        \caption{\emph{Diverse Objects, articulated interaction.} Prompt: (I2V-68) "The camera remains still, the person is opening the door of a microwave." Based on the image, most models struggle to correctly understand and generate motions that align with the functional attributes of articulated interaction.}
        \label{Fig:articulated}
\end{figure}

\begin{figure}[!ht]
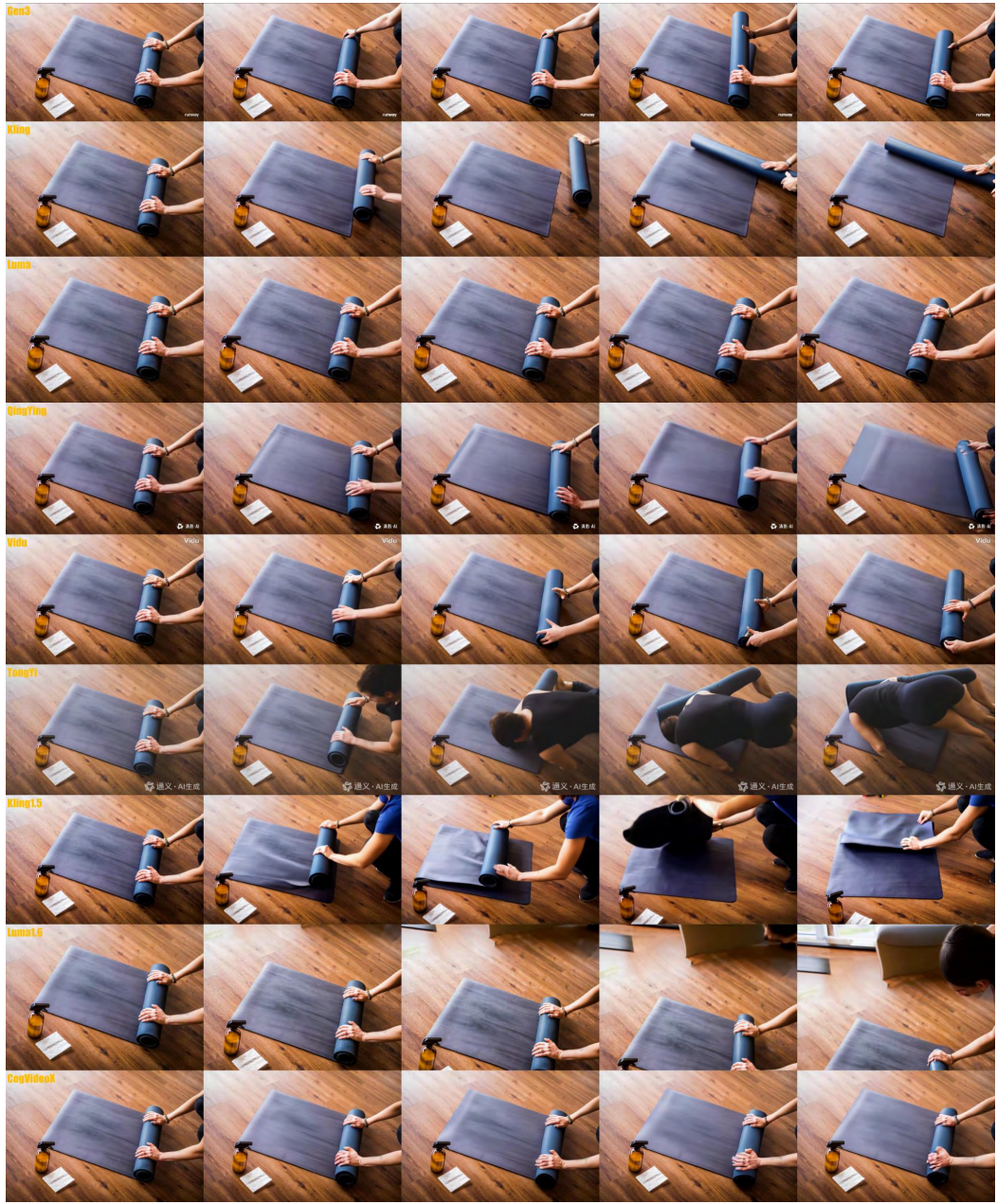

	\centering
	\small
        \begin{overpic}[width=1.\linewidth]{Figs/Sec7/00073.pdf}

	\end{overpic}
        \caption{\emph{Diverse Objects, deformation.} Prompt: (I2V-73) "The camera remains still, the person is rolling up a yoga mat." Based on the image, most models struggle to correctly understand and generate motions that align with the functional attributes of deformation objects.}
        \label{Fig:deformable}
\end{figure}

\begin{figure}[!ht]
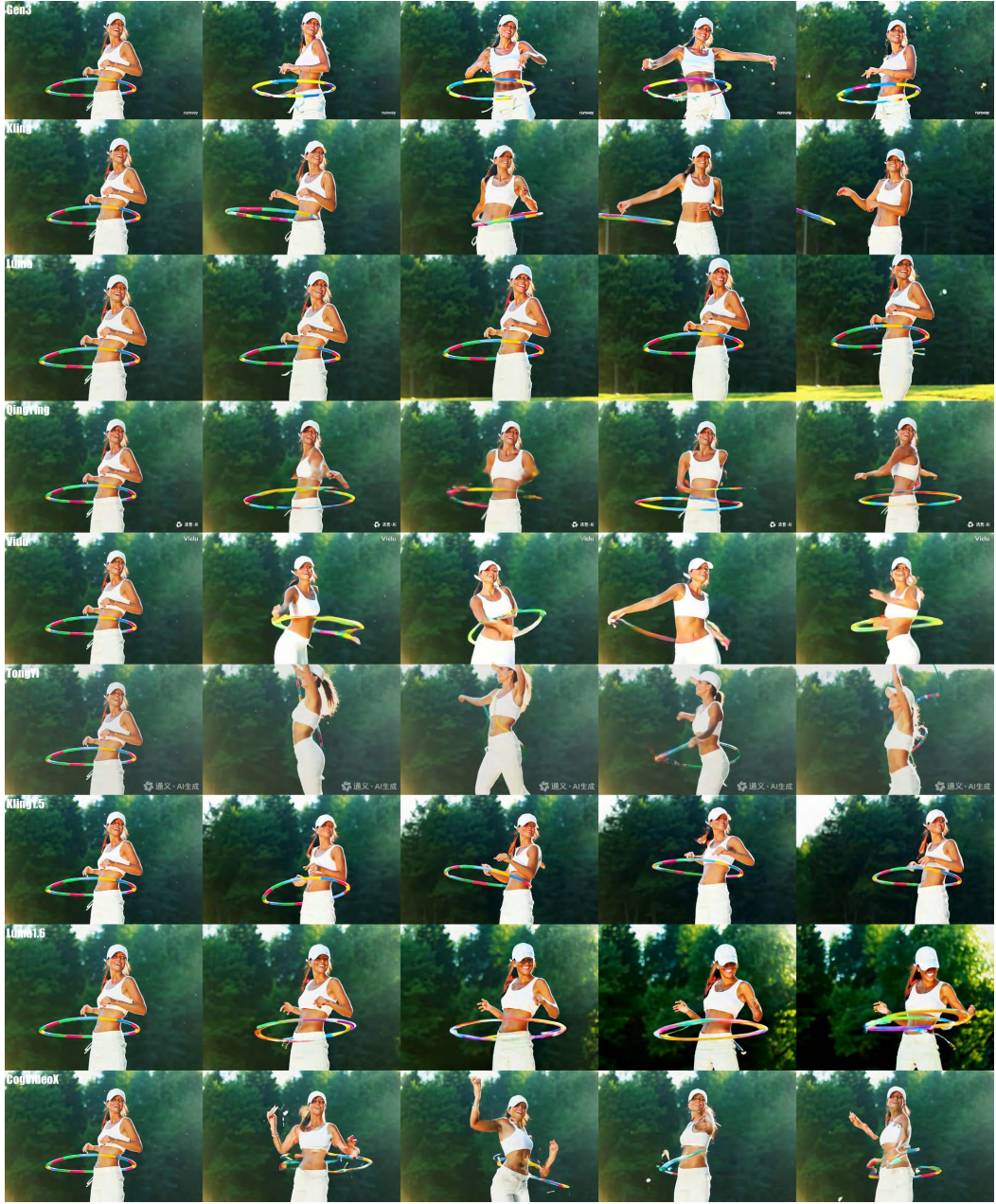

	\centering
	\small
        \begin{overpic}[width=1.\linewidth]{Figs/Sec7/00108.pdf}

	\end{overpic}
        \caption{\emph{Synergistic motion.} Prompt: (I2V-108) "Static camera, the woman is hula hooping outdoors." Dynamically handling the physical interactions between human body and the object is quite challenging under synergistic motion, requiring models handling complex coordination, ensuring temporal and spatial consistency, maintaining physical contact realism.}
        \label{Fig:Synergistic motion}
\end{figure}

\begin{figure}[!ht]
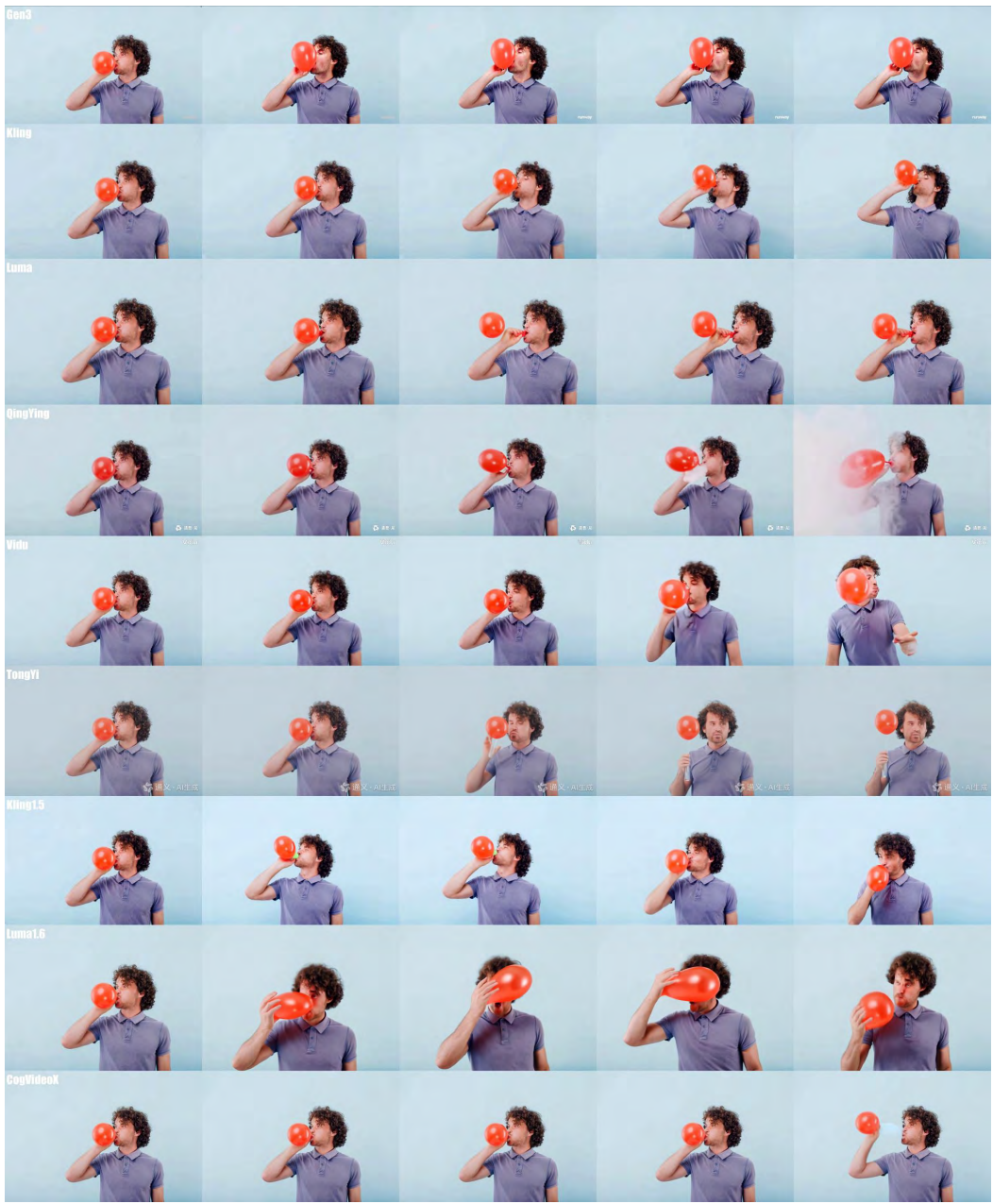

	\centering
	\small
        \begin{overpic}[width=1.\linewidth]{Figs/Sec7/00004.pdf}

	\end{overpic}
        \caption{\emph{Causality.} Prompt: (I2V-4) "The camera remains still, this man blew air into the balloon." This prompt expects that the man blowing air into the balloon causes the balloon to expand as the internal pressure increases. However, it is still challenging for all models.}
        \label{Fig:Causality_1}
\end{figure}

\begin{figure}[!ht]
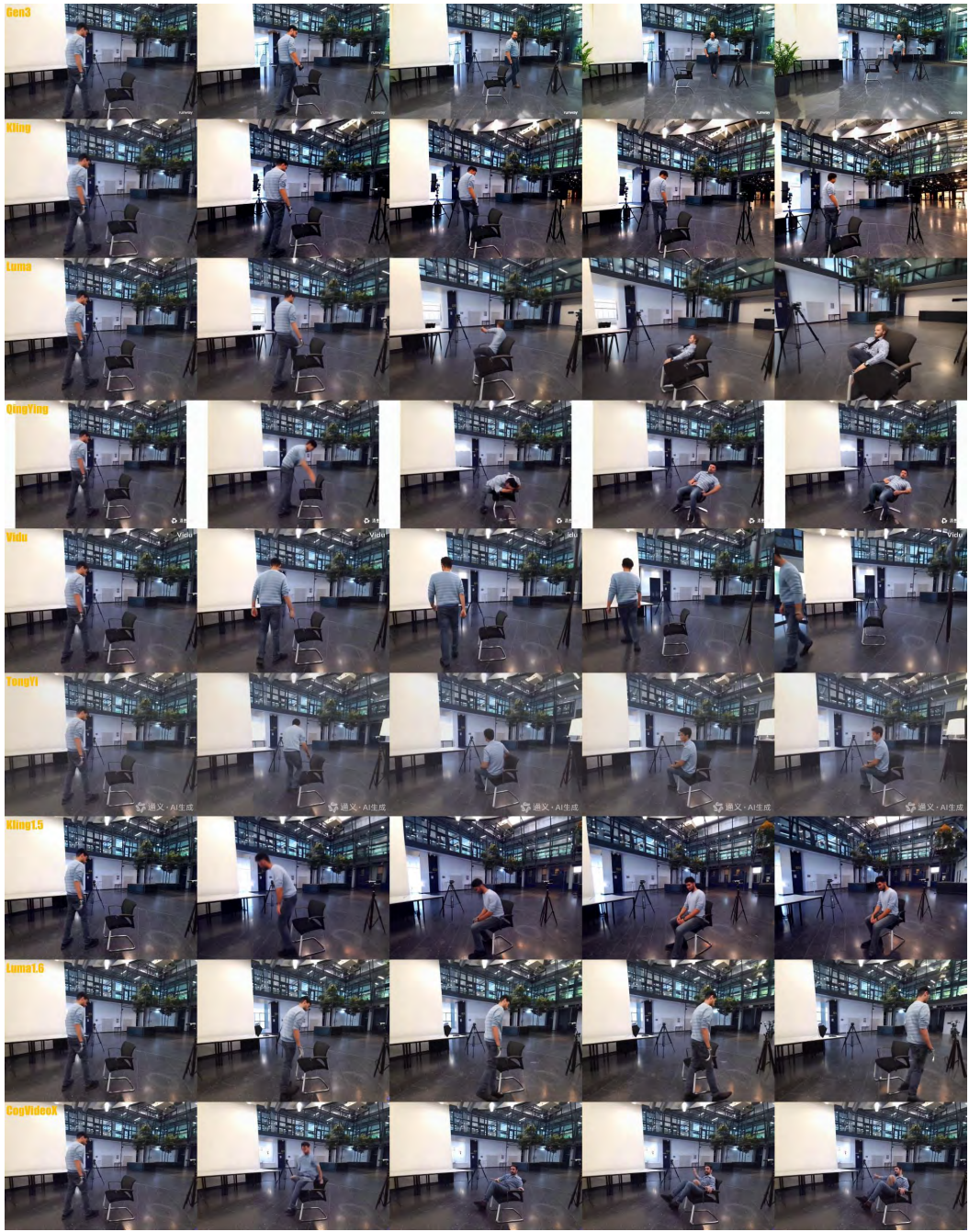

	\centering
	\small
        \begin{overpic}[width=1.\linewidth]{Figs/Sec7/00010.pdf}

	\end{overpic}
        \caption{\emph{Intentional Interaction.} Prompt: (I2V-10) "The camera remains still, the man walks up to the front of the chair and sits on it." Most models fail to see and understand the given image, leading to the simple "sit on the chair" action fails. Here, Kling1.5 successes to complete the action.}
        \label{Fig:Intentional}
\end{figure}

\textbf{Challenges.} Interactions, particularly human-centric interactions, are reflected in various application scenarios and are also a necessary pathway to generating complex content. However, according to our testing, existing methods still demonstrate limitations in handling such scenarios. Human-centric interactions are diverse and complex, \eg, interacting with numerous objects and scenes, which inherently involves numerous physical world principles, making them an ideal scenario to evaluate the capabilities of foundation models. Despite the consistency, we further outline several key challenges that warrant focused attention to inspire future work:

\begin{itemize}[leftmargin=*]

\item \textbf{Diverse Objects}: there are many types of objects involved in interactions, including rigid, articulated (Figure \ref{Fig:articulated}), deformable (Figure \ref{Fig:deformable}), etc. These objects may have distinct motion patterns and part poses during the interaction. At present, generating the motion of specific interacting objects is still very challenging, particularly the deformable objects.

\item \textbf{Synergistic Motion}: humans usually operate objects, causing them to undergo translation, rotation, or even deformation. This requires models to move beyond grasping isolated motion patterns to capture the correlated motion between humans and objects, simultaneously generating synergistic motion of multiple instances (Figure \ref{Fig:Synergistic motion}).

\item \textbf{Causality}: based on the objects' functionalities, the interactions driven by humans may alter the objects or environment. For instance, blowing up a balloon (Figure \ref{Fig:Causality_1}) or hitting with a hammer. In such situations, the model needs to reason and generate the post-interaction state. However, existing DiT-based methods tend to learn the relation of frames across spatial and temporal, rather than logical associations of contextual contents, which may limit their ability to generate such causal scenarios.

\item \textbf{Intentional Interaction}: this type of interaction involves the process from non-contact to final interaction, \eg, walking towards the chair and sitting down (Figure \ref{Fig:Intentional}). It requires the model to understand which objects to interact with, the object affordances, and the poses humans should make to form the interaction. Especially, the model needs to perceive where the interacting object within the instruction is located for I2V generation. The testing results show that the model’s capability still needs improvement in this scenario.
\end{itemize}

\textbf{Future Work.} To generate realistic interactions, it is crucial to clearly analyze the elements related to the interaction, such as the object pose during the interaction (part and global), interaction attributes, motion patterns, human actions, contact between humans and objects, and spatial relation of the interacting counterparts. Defining these interaction elements clearly and crafting corresponding data, including raw videos and fine-grained annotations (either textual or pixel-space signals), is essential.

\clearpage

\subsection{Personalization}

Personalized or customized video generation and editing is the key step towards practical and user-friendly video content creation and applying to various commercial applications. It usually needs to add some specific components to the generated videos. For instance, input an image or a few images of a person or a product and combine them with a text prompt to generate a video that contains the reference person or product with rich visual details controlled by the text prompt~\cite{bao2024vidu,ruiz2023dreambooth,xing2024make,chen2023videodreamer,ma2024magic,he2024id}. The performance of personalized video generation highly depends on the performance of foundation video generation models. Thus, the specific identity-preserving human video generation models (ID-Animator~\cite{he2024id}) are hard to keep either the facial characteristics (\emph{e.g.}, eyeglasses and other details) or follow the given instructions. In contrast, the general-purpose text-to-video generation models with identity-preserving adaptation strategies can follow instructions well (\emph{e.g.}, the \emph{character-to-video} function of Vidu). Due to the limitations in Vidu, such as unstable motion and low visual quality, its personalized video generation will be challenging to adapt the identity to the generated videos.

\begin{figure}[!ht]
\vspace{-0.4cm}
	\centering
	\small
        \begin{overpic}[width=0.9\linewidth]{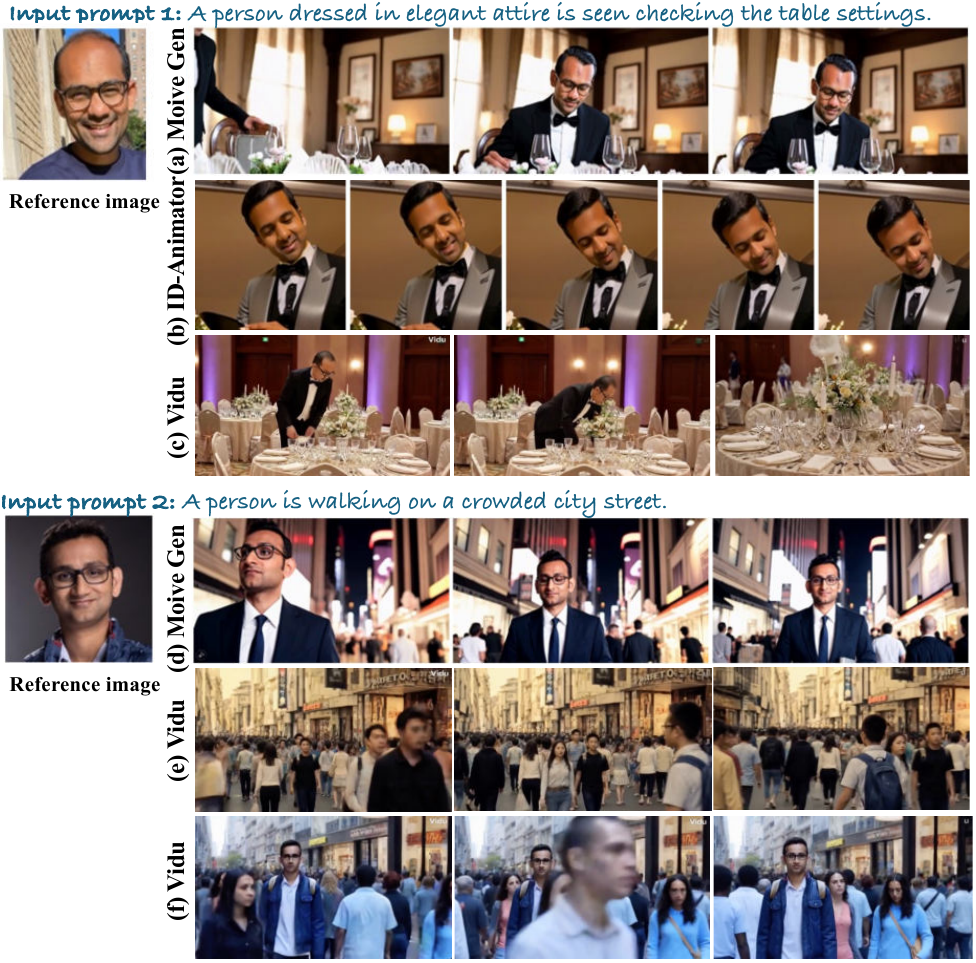}
	\end{overpic}
    \vspace{-0.2cm}
	\caption{\emph{Comparisons on personalized video generation}. Given a portrait image and text prompt, the model should generate a text-aligned video with a consistent ID. For (e-f) results from Vidu, we run two times with the same inputs but different random seeds.}
 \label{Fig:person}
 \vspace{-0.4cm}
\end{figure}

Recently, Movie Gen~\cite{meta2024moviegen} has expanded the proposed foundation media model to support personalized video generation using 4096 GPUs (\emph{e.g.}, H100) as the training resources, which is quite high-cost. We use the cases in the report and generate videos via Vidu for a fair comparison. From Figure~\ref{Fig:person}, Movie Gen, with better text-to-video performance, achieves better results when creating personalized videos that preserve human identity and motion. We run two times with the same inputs but different random seeds. We can observe the unstable results from Vidu, which means some cases will generate bad results (Figure~\ref{Fig:person}(e)), but others will be generated reasonably (Figure~\ref{Fig:person}(f)).
The efficient and high-quality personalization ways with arbitrary input images should be explored. Besides, the personalization could expand from the face to the whole-body identity and any objects, and stylized motions (\emph{e.g.}, transferring the motion of characters from Pixar animations to generated  motions). 

\clearpage

\subsection{Text Generation}
\begin{figure}[!ht]
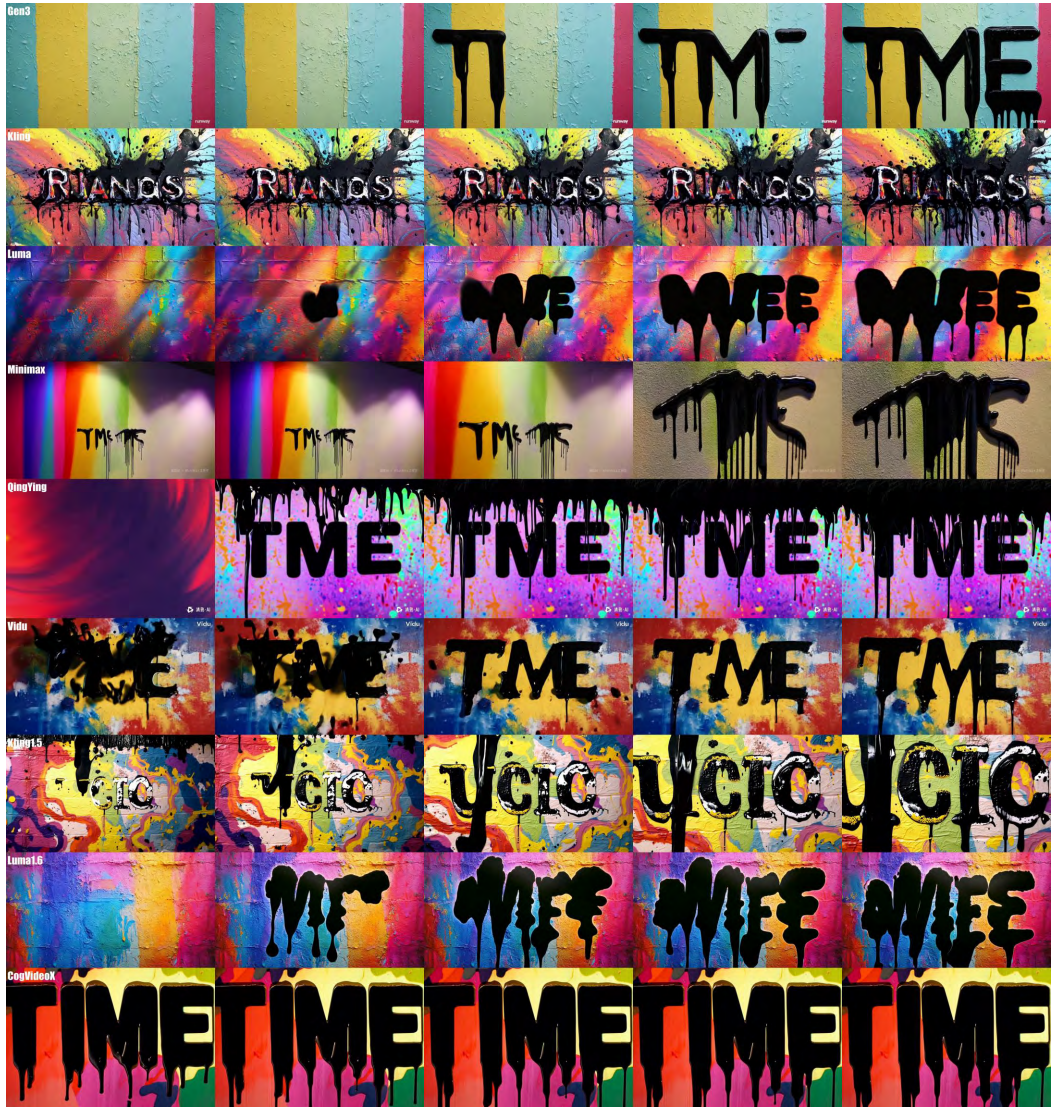

	\centering
	\small
        \begin{overpic}[width=1.\linewidth]{Figs/Sec7/00411.pdf}

	\end{overpic}
        \caption{\emph{Text generation.} Prompt: (T2V-411) "A title screen with dynamic movement. The scene starts at a colorful paint-covered wall. Suddenly, black paint pours on the wall to form the word ``TME''. The dripping paint is detailed and textured, centered, superb cinematic lighting." Gen-3, QingYing and Vidu correctly generate the text, while the remaining models exhibite issues such as failure to generate text, generating incorrect text, or producing other letters.}
        \label{Fig:Text_gen1}
\end{figure}

Existing models demonstrate a certain degree of alignment and artistic creativity when generating short text in English. However, they still struggle to handle \textbf{i)} long texts and \textbf{ii)} process multilingual texts (Figure \ref{Fig:Text_gen1}, \ref{Fig:Text_gen2}, \ref{Fig:Text_Gen-3}). This may be due to the lack of such training data with precise text recognition and dense captions from videos, and specific languages also need to consider the tokenizer and text encoder. Building high-quality datasets for text generation is also worthwhile.

\begin{figure}[!ht]
	\centering
	\small
        \begin{overpic}[width=0.8\linewidth]{Figs/Sec7/00622.pdf}

	\end{overpic}
        \caption{\emph{Text generation.} Prompt: (T2V-622) "Generate the text ``\begin{CJK}{UTF8}{min}  
            人工知能の未来へようこ！
        \end{CJK}'' with Japanese, in the city view." All models generate the background well, but they fail to correctly generate Japanese text.}
        \label{Fig:Text_gen2}
\end{figure}

\begin{figure}[!ht]
	\centering
	\small
        \begin{overpic}[width=0.8\linewidth]{Figs/Sec7/00624.pdf}

	\end{overpic}
        \caption{\emph{Text generation.} Prompt: (T2V-624) "Generate the text ``
        \begin{CJK}{UTF8}{gbsn}  
        人工智能的未来
        \end{CJK}'' with Chinese, in the city view." All models generate the background well, but they fail to correctly generate Chinese text.}
        \label{Fig:Text_Gen-3}
\end{figure}
\clearpage

\subsection{Fine-grained Controllable Generation}
\begin{figure}[!ht]
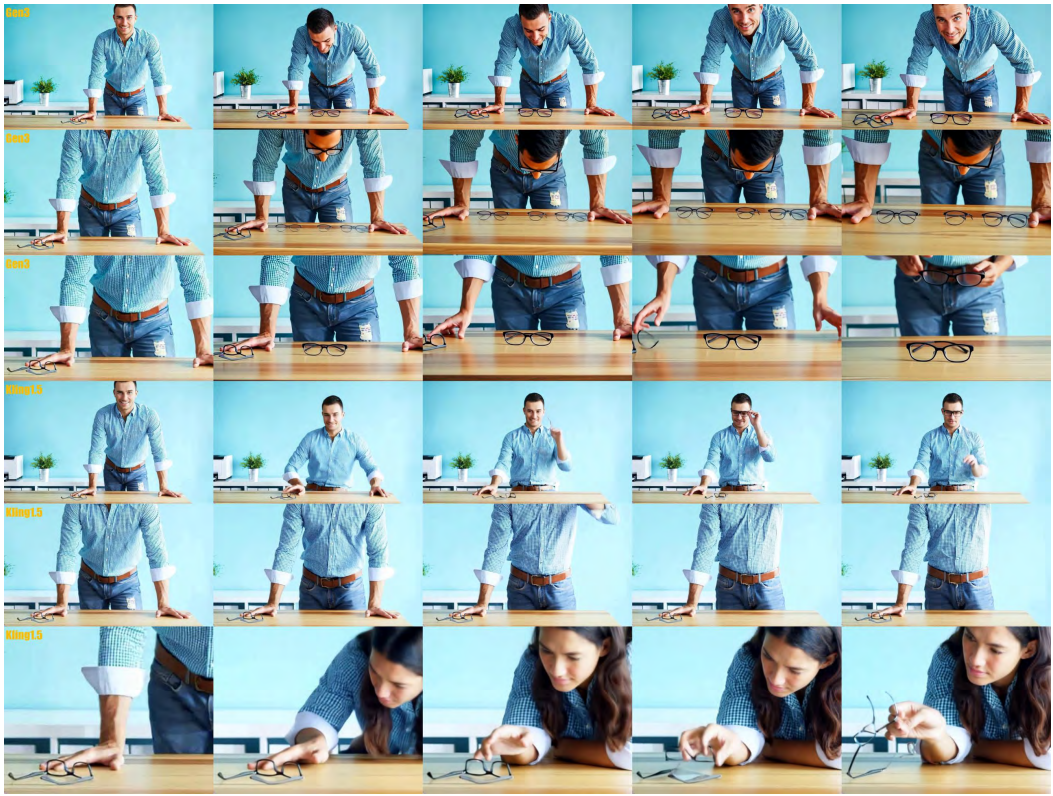

	\centering
	\small
        \begin{overpic}[width=1.\linewidth]{Figs/Sec7/scale.pdf}

	\end{overpic}
        \caption{\emph{Fine-grained control.} Prompt: (I2V) first row: "The camera remains still, the man picks up the glasses on the table with his right hand." We try to explore why existing I2V could not generate the fine-grained interaction or interact with objects in the given image. We take two models as examples (\emph{e.g.}, Kling 1.5 and Gen-3), the proportion of target instances \eg, the glasses in the input image gradually increases. }
        \label{Fig:fine-control}
\end{figure}

Fine-grained control over specific regions is quite crucial for T2V, I2V, and V2V, especially for downstream applications and practical content creation products. For instance, making a particular instance move or adding, removing, and modifying instances in the input video or image. To achieve this, an encoder that can understand the input image or video at a fine-grained level is a crucial prerequisite, as it ensures precise control over specific regions. At present, fine-grained control over local regions of I2V solely through text still struggles to produce satisfactory results. While video-to-video generation shows promising style transfer effects, it remains limited in local editing capabilities (please refer to Figure \ref{Fig:Exploration2}). We explore whether the proportion of the target instance in the image would have an impact (Figure \ref{Fig:fine-control}). Still, there is no significant effect as the proportion or scale of the target instance in the image gradually increases. 

For future work, achieving fine-grained control in the I2V and V2V generation requires not only considering what conditions to use, and how to inject those conditions, but also exploring whether there are gaps in the encoding of input images or videos and condition features. For example, whether the encoded features can align with the fine-grained control conditions, and whether it identifies the specific fine-grained regions being controlled.

\subsection{Long video, Multi shots, ID Preservation}
Generating long videos poses challenges for both training and inference computational resources. It is also related to the efficient designs of model architectures. Some methods manage to generate long videos for specific tasks, \eg, pose-driven human dance through the sliding window mechanism, or some perform continuation by taking the last generated frame as the next input. However, these methods of obtaining long videos are unrelated to the model’s native capabilities and are challenging to ensure consistency and coherence over long temporal contexts. To generate longer videos, several aspects are worth considering: \textbf{i)} efficient architectures or mechanisms capable of capturing long-range context; \textbf{ii)} well-trained 3D-VAE, the current approaches mainly take latent denoising; thus, VAEs capable of maximally compressing both the temporal and spatial dimensions allows for the generation of more frames within a limit computational resource.

Furthermore, generating long videos typically involves multiple camera shots, which presents significant challenges in maintaining ID and style consistency across different shots, especially in preserving the appearance of characters across various viewpoints. Each shot change increases the difficulty of ensuring that the appearance, motion, and attributes of characters remain coherent throughout the generated video.

\subsection{Efficiency}
Currently, commercial models take several minutes to generate a video of 5-10 seconds (excluding the queuing time). The inference efficiency urgently needs improvement. Here, we list some potential aspects for breakthroughs: \textbf{i)} mechanism for relation modeling. Recently, DiT-based methods \cite{xu2024easyanimate, yang2024cogvideox} concatenate the text and video sequences to perform full-attention operations. Although it brings an improvement, it incurs high computational overhead and is likely to have redundant calculations. Exploring the efficient mechanism for capturing correlation across different modalities and spatio-temporal dimensions is necessary; \textbf{ii)} target distribution for modeling. Some methods \cite{li2024autoregressive, li2024denoising} explore combining autoregressive models with diffusion, where the denoising focuses on pixels rather than the full image sequence. The simpler target distribution enables the denoising network to be more lightweight. Additionally, mask mechanism \cite{chang2022maskgit} enables inferring multiple tokens at a step for autoregressive models, offering the potential for improving the efficiency; \textbf{iii)} fewer sampling steps. Reducing sampling steps is the most direct way to improve the inference efficiency of video diffusion models. To achieve this, the modeling approach \cite{liu2022flow, song2023consistency} and solvers \cite{lu2022dpm} are worth considering.

\subsection{Multimodal Video Generation}
So far, text is the primary modality used to control video generation, and certain methods explore other modalities, like audio, pose sequences, and trajectories. For DiT-based models, the text condition gradually shifts from the control branch to the input sequence. Either branch relies on isolated encoders to encode the text or image, but there are still certain gaps between the embedding obtained by these encoders. For instance, in testing prompts where prepositions indicate temporal order, the model struggles to generate content aligned with the order in prompts. Recently, some methods \cite{xie2024showo, zhou2024transfusion,ao2024body,meta2024moviegen} explore unifying multiple modalities in one framework, enabling the simultaneous generation of various modalities. These approaches are conducive to learning multimodal contexts. In this paradigm, how to optimize distinct modalities during end-to-end training and whether to quantify visual representations remain open insights for further exploration.

\subsection{Continuous Improvement in Video Generation}
\begin{figure}[!ht]
	\centering
	\small
        \begin{overpic}[width=0.95\linewidth]{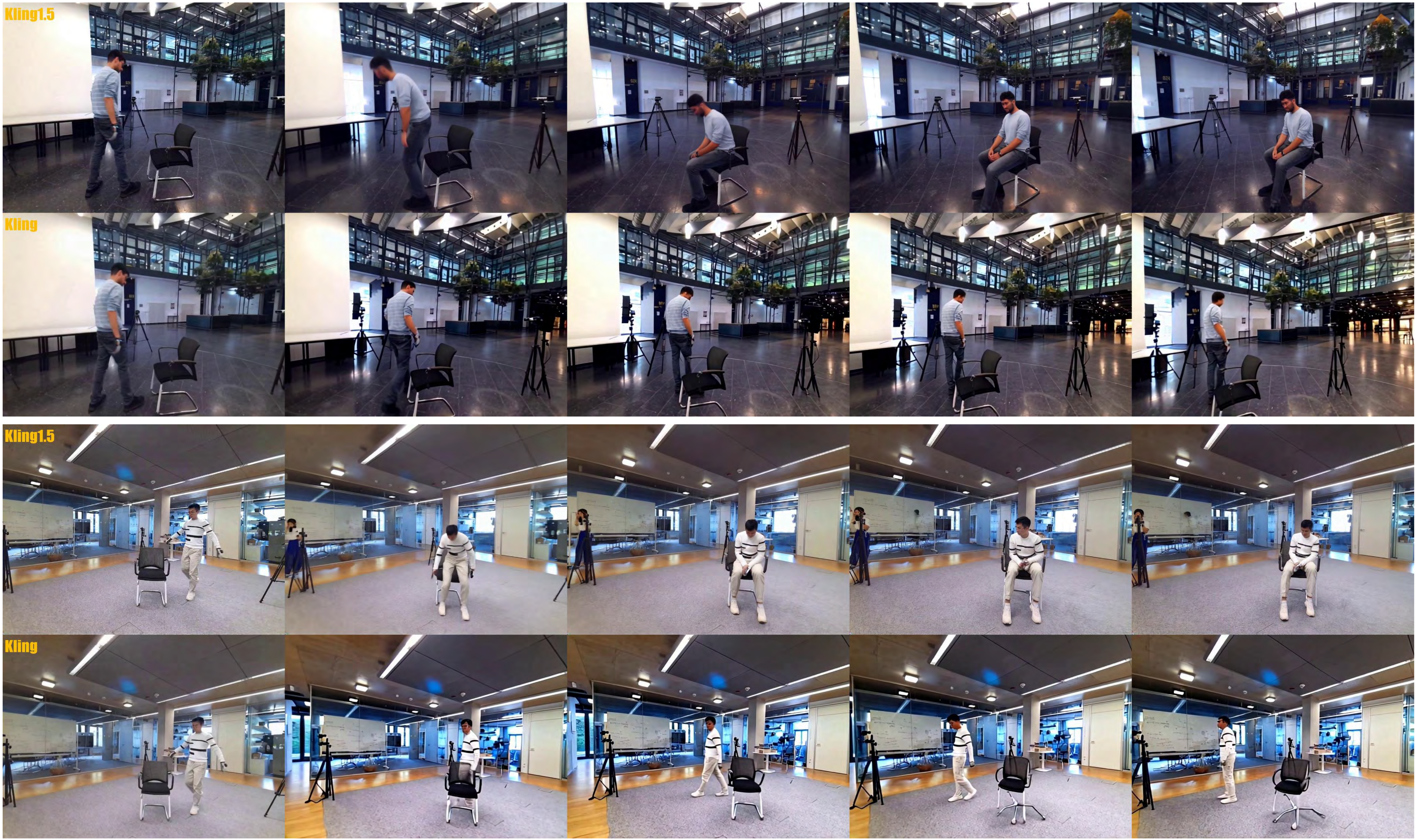}

	\end{overpic}
        \caption{\emph{Continue improvement from the foundation models.} Comparison of results from different versions of Kling. Prompt: (I2V) first row: "The camera remains still, the man walks up to the front of the chair and sits on it." Second row: "The camera remains still, the man walks up to the front of the chair and sits on it." After sclaing up the model parameters and training data, Kling 1.5 has significantly achieved interaction generation that Kling 1.0 is unable to accomplish.}
        \label{Fig:Countinue_imporve}
\end{figure}

Analogous to large language models (LLMs), video generation training is still in the pre-training and SFT stages. However, both the quantity of training data and the model parameters are far behind those of LLMs. Further scaling up high-quality data and model parameters may continue to enhance model capabilities (Figure \ref{Fig:Countinue_imporve}). Another candidate solution to improve the model ability lies in providing feedback on the generated results \cite{prabhudesai2024video, yuan2024instructvideo}. The challenge involves determining what signals or interactive inputs can be used for effective video feedback that may differ from image and text, and how to further optimize the trained model. As cases demonstrated in this report, the models' ability varies across scenarios, and distinct scenarios lead to varying key points for human evaluation. That means existing models will generate imperfect videos with various artifacts. Continuously improving the generated videos would be a quite important problem in the future. Compared to designing general feedback signals or reward models, scenario-specific feedback may be more beneficial in addressing the weaknesses of existing models.

Therefore, we release our generated videos from various closed-source and open-source models, some man-made professional videos as potential ground-truth videos, to pave the way to train and evaluate this task. For instance, we can take the imperfect generated videos as input and introduce advanced techniques to improve them to perfect or make videos available further. This is a great step in making the generated videos controllable and usable. Moreover, we provide some results from the same origin models with different sizes or versions, such as Kling 1.0 and Kling 1.5, Luma 1.0 and Luma 1.6, CogVideoX-5B, and closed-source Qingying models. The paired models could be explored for model knowledge distillation, as well as how and what has been optimized in the different models. Could we find a reliable scaling-up path?

\vspace{-0.3cm}
\section{Conclusion}
In conclusion, this report comprehensively explores SORA-like models in high-quality video generation, encompassing T2V, I2V, and V2V tasks. By devising a series of detailed prompts and case studies, we have systematically assessed the advancements, challenges, and potential applications of these models across various scenarios. Our analysis highlights the significant strides made in video generation, particularly regarding visual quality, motion naturalness and dynamics, and vision-language alignment. 

Despite these achievements, we identify the potential gaps between academic research and industry practice, suggesting areas for future directions in fine-grained and hierarchical video caption and understanding for controllable video generation, efficiency, scalability, stability, human feedback for constant result improvement and modification, complex interaction modeling, long video generation with multi-shot and ID-consistency, automatically human-aligned evaluation strategies, and multi-modal interactive generation systems.

We hope this study and our generated videos serve as a new video generation benchmark and inspire further research. We encourage the community to use our public prompts and videos for further research, such as continuing learning in generated videos. By directly observing generated videos, we expect to contribute to a more nuanced understanding of video generation capabilities. As the field evolves rapidly, we are optimistic about future breakthroughs and look forward to further enhancing the quality and versatility of video generation. Meanwhile, we commit to updating our findings to reflect ongoing advancements with more comprehensive and detailed prompts to cover all aspects in videos, exploring reliable quantitative evaluation, and fostering deeper insights into this rapidly progressing domain.

{
\bibliography{main}
\bibliographystyle{plain}
}

\end{document}